\documentclass[preprint,12pt]{elsarticle}
\usepackage[utf8]{inputenc} 
\usepackage[T1]{fontenc}    
\usepackage{hyperref}       
\usepackage{url}            
\usepackage{booktabs}       
\usepackage{microtype}      
\usepackage{lipsum}	    	
\usepackage{multirow}
\usepackage{graphicx}
\usepackage{doi}
\usepackage[section]{placeins}
\usepackage{mathtools}
\usepackage{amsmath}
\usepackage{amssymb}
\usepackage[ruled,vlined]{algorithm2e}
\usepackage{lineno}
\usepackage[dvipsnames]{xcolor}
\usepackage{enumitem}
\usepackage{tabularx} 
\usepackage{array}
\usepackage{subcaption}

\usepackage{siunitx}

\usepackage{{makecell}}

\setcellgapes{4pt} 

\newcolumntype{Y}{>{\centering\arraybackslash}X}
\newcolumntype{M}[1]{>{\centering\arraybackslash}m{#1}}

\setlist[itemize]{noitemsep, topsep=0pt}
\usepackage{amssymb}
\usepackage[a4paper, total={6in, 8in}, margin=1in]{geometry}

\journal{Elsevier}
\usepackage{color}
\usepackage{soul}

\begin{document}

\begin{frontmatter}

\title{Physics-informed DeepONet with stiffness-based loss functions for structural response prediction}

\author[inst1]{Bilal Ahmed\corref{cor1}}\ead{ba2702@nyu.edu}
\author[inst1,inst2]{Yuqing Qiu}
\author[inst1,inst3]{Diab W. Abueidda\corref{cor1}}\ead{da3205@nyu.edu}
\author[inst1,inst4]{Waleed El-Sekelly}
\author[inst1]{Borja García de Soto}
\author[inst1]{Tarek Abdoun}
\author[inst1]{Mostafa E. Mobasher\corref{cor1}}\ead{mostafa.mobasher@nyu.edu}

\affiliation[inst1]{
    organization={Civil and Urban Engineering Department},
    addressline={New York University Abu Dhabi}, 
    country={United Arab Emirates}
}

\affiliation[inst2]{
    organization={The State Key Laboratory of Mechanics and Control of Mechanical Structures},
    addressline={Nanjing University of Aeronautics and Astronautics}, 
    country={China}
}

\affiliation[inst3]{
    organization={National Center for Supercomputing Applications},
    addressline={University of Illinois at Urbana-Champaign}, 
    country={United States of America}
}

\affiliation[inst4]{
    organization={Department of Structural Engineering},
    addressline={Mansoura University}, 
    country={Mansoura, Egypt}
}

\cortext[cor1]{Corresponding author}

\begin{abstract}

Finite element modeling is a well-established tool for structural analysis, yet modeling complex structures often requires extensive pre-processing, significant analysis effort, and considerable time. This study addresses this challenge by introducing an innovative method for real-time prediction of structural static responses using DeepOnet which relies on a novel approach to physics-informed networks driven by structural balance laws. This approach offers the flexibility to accurately predict responses under various load classes and magnitudes. The trained DeepONet can generate solutions for the entire domain, within a fraction of a second. This capability effectively eliminates the need for extensive remodeling and analysis typically required for each new case in FE modeling. We apply the proposed method to two structures: a simple 2D beam structure and a comprehensive 3D model of a real bridge. To predict multiple variables with DeepONet, we utilize two strategies: a split branch/trunk and multiple DeepONets combined into a single DeepONet. In addition to data-driven training, we introduce a novel physics-informed training approaches. This method leverages structural stiffness matrices to enforce fundamental equilibrium and energy conservation principles, resulting in two novel physics-informed loss functions: energy conservation and static equilibrium using the Schur complement. We use various combinations of loss functions to achieve an error rate of less than 5\% with significantly reduced training time. This study shows that DeepONet, enhanced with hybrid loss functions, can accurately and efficiently predict displacements and rotations at each mesh point, with reduced training time.
\end{abstract}

\begin{highlights}
\item Advanced ML technique is developed based on DeepONet to replicate FEM responses for civil structures.
\item Various DeepONets are investigated to predict multiple degrees of freedom response of the structure.
\item Novel physics incorporation methods based on structural stiffness are introduced.
\item Achieved over 95\% accuracy in predictions with extreme efficiency for various loading combinations.

\end{highlights}


\end{frontmatter}


\section{Introduction} \label{intro}

\subsection{Overview}\label{Intro_Overview}

Over the past few decades, the finite element method (FEM) \cite{brenner2002mathematical,hughes2012finite} has become the practical standard tool for simulating and understanding the complex behaviors of many engineering structures and mechanical systems. FEM is particularly effective for solving partial differential equations (PDEs) in scenarios where analytical solutions are unattainable, providing accurate approximations of structural deformation grounded in strong mathematical principles. However, as problem complexity increases, developing the FEM model requires significant time, and even slight changes in loads, material properties, and boundary conditions necessitate reprocessing and reanalysis of the model. To address this challenge, our work introduces DeepONet enhanced with physics information to efficiently predict the elastic deformation of structures. This method accurately predicts structural behavior across the entire domain within a fraction of a second. We focus on predicting multiple variables (displacement and rotations) at each mesh point of real-life structures under variable static load classes and magnitudes, ensuring that the predictions adhere to fundamental equilibrium and energy conservation principles.

\subsection{Literature review and research gaps}\label{Intro_Lit}

With the increase in computational power, machine learning (ML) methods such as artificial neural networks (ANNs) have been extensively used to forecast the behavior of various systems. Among ANNs, multilayer perceptrons (MLPs) were commonly applied for damage detection in civil structures \cite{xu2022real, ye2022real}. The general applications of ANNs in predicting the dynamic behavior of diverse structures, such as buildings, marine constructions, and bridges, can be found in \cite{guarize2007neural, lagaros2012neural, liao2021automated, wang2022end}. Significant research was conducted on damage detection, fatigue life prediction, fault diagnosis, and seismic damage prediction \cite{xu2020prediction, lei2019fault, mas2017initial, qiu2019modified, saleem2024machine}. Additionally, substantial work on spalling, cracking, rebar exposure, and buckling in reinforced concrete members can be found in \cite{xu2023computer, xu2023vision, xu2019automatic}. Xu et al. \cite{xu2022typical} presented a review of recent developments in AI applications in civil structures, focusing on structural health monitoring, automation, and reliability analysis. Moreover, ML models addressed various challenges in additive manufacturing \cite{WANG2020101538}, bio-inspired structures \cite{he2024sequential,zhang2022dynamic}, damage detection in civil structures \cite{ahmed2022seismic,ahmed2023generalized,ahmed2023unveiling,cha2017deep}, nonlinear material responses \cite{mozaffar2019deep,egli2021surrogate,abueidda2021deep}, and other applications. Despite these advancements, to the best of the authors' knowledge, no work in the literature focuses on replicating FEM responses using ML techniques with physics-based constraints to efficiently predict deformations in civil structures.

Most of the specified studies solely relied on a data-driven approach, training the ML network based only on the available data, which may lead to incorrect predictions when the testing data lies beyond the training domain. To address this issue, researchers integrated physics into neural networks (NNs), resulting in the development of physics-informed neural networks (PINNs) \citep{raissi2019physics, abueidda2021meshless}. Physics-informed models incorporate physical laws and governing equations into the loss function, ensuring that predictions are based on both the training data and the underlying physics.  Recent studies demonstrated the potential of integrating PINNs with FEM \citep{pantidis2023integrated, abueidda2024fenn}. One way to incorporate physics into the network is by formulating the loss function based on the residual of PDEs or governing equations \citep{niaki2021physics,henkes2022physics,rao2021physics}. Another method involves variational and energy-based approaches \citep{yu2018deep,liao2019deep,samaniego2020energy,nguyen2020deep,abueidda2022deep}.  In this study, we use a different approach: instead of relying on PDEs or governing equations, we leverage structural stiffness matrices and enforce the principles of static equilibrium (SE) and energy conservation (EC) to develop the physics-informed loss functions. We employ various combinations of loss function types; more details can be found in Section \ref{Proposed_Section}.

Most existing approaches (NNs and PINNs) require retraining or transfer learning for any change in input parameters, such as loads or discretization, similar to traditional numerical methods that require new simulations for each new set of parameters. To address this, Lu et al. \citep{lu2021learning} developed DeepONet. DeepONet contains a branch network that encodes the input functions and a trunk network that encodes the input domain geometry, allowing it to solve entire families of PDEs for specific problems and learn operators like integrals or differentials. DeepONet was extended to incorporate traditional physical laws (PDE-based) in the learning process, improving prediction accuracy and data handling efficiency at the expense of higher computing costs \citep{wang2021learning}. Significant accuracy was achieved in various applications of continuum mechanics \cite{koric2023data,yin2022interfacing,goswami2022physics,he2023novel}. So far, DeepONet was primarily used for continuum 2D problems, solving PDEs such as advection, Burgers, diffusion, wave propagation, Darcy flow, and flow in porous media \cite{wang2021learning, wang2023long, goswami2023physics}. However, it had not been tested for practical structural modeling involving assemblies of discrete structural elements (beams, shells) and multiple output variables at each mesh point.  In this work, we develop DeepONet to simultaneously predict the displacements and rotations of discrete real-life structures under static loading, ensuring adherence to physical laws.

DeepONet was designed to handle both single and multiple output functions simultaneously \citep{lu2021learning}. Most applications found in the literature focus on a single output function from DeepONet \citep{wang2021learning, he2024sequential}. In our work, we need to predict multiple output functions across the entire domain. To address this, we adopt the strategies specified by Lu et al. \citep{lu2022comprehensive} to handle multiple output functions. The first strategy involves using $N$-independent DeepONets, each outputting one function. The second strategy is the split branch and trunk network approach, which splits the outputs of both the branch/trunk networks into $N$ groups, with each group outputting a single function.

\subsection{Contributions and Paper Structure}\label{Intro_Contri}
In the field of structural engineering, most ML applications focus on damage detection, concrete spalling, cracking, seismic response, and fatigue life predictions \cite{xu2020prediction, lei2019fault, mas2017initial, qiu2019modified, xu2023computer, xu2023vision, xu2019automatic}. However, to the author's knowledge, no previous work has attempted to replicate the FEM resolution of structural response, i.e. at every material point in each structural member. The purpose of this work is to avoid the extensive remodeling effort associated with FEM for every new case and to predict the governing variables (displacements and rotations) across the entire domain of the structure. To achieve this goal, we present an operator-learning solution using DeepONet. To ensure maximum accuracy and avoid underfitting or overfitting, we perform a detailed parametric study by varying the number of layers in the network, the number of neurons in each layer, batch size, and aspect ratio of the network. Additionally, we test various architectures of DeepONet to manage multiple outputs from a single main network, including the split branch/trunk strategy and the $N$-independent DeepONet strategy. To ensure that predictions adhere to the underlying physics of the structure, we leverage a pre-calculated stiffness matrix to develop novel physics-informed loss functions based on energy conservation and static equilibrium principles, avoiding the cumbersome and expensive use of governing PDEs. To further enhance the efficiency of DeepONet and reduce training time, we use the Schur complement to reduce the problem size. Finally, to verify the applicability of the proposed model, we test it on two structures: a 2D beam structure and a 3D model of a real bridge, KW-51, in Leuven, Belgium.

The structure of the paper is as follows: Section \ref{Proposed_Section} provides a detailed methodology, including the development of DeepONet and the formulation of loss functions. Section \ref{toybridge} discusses the application of DeepONet to a 2D beam structure, covering FEM modeling, data generation, a parametric study, results, and discussions. Section \ref{KW51_Sec} extends the application to the real-life KW-51 structure, detailing FEM modeling, validation, data generation, and training with different combinations of loss functions. Finally, Section \ref{conclu} presents the conclusions of the study.

\section{Proposed Method}\label{Proposed_Section}

This section details the methodology used for the proposed work, which includes the overall methodology of the work, the proposed DeepONet, and the development of novel loss functions. 

\subsection{Overall Methodology}\label{Overall_Method}

\begin{figure}[!htb]
    \centering
    \includegraphics[width=0.85\textwidth]{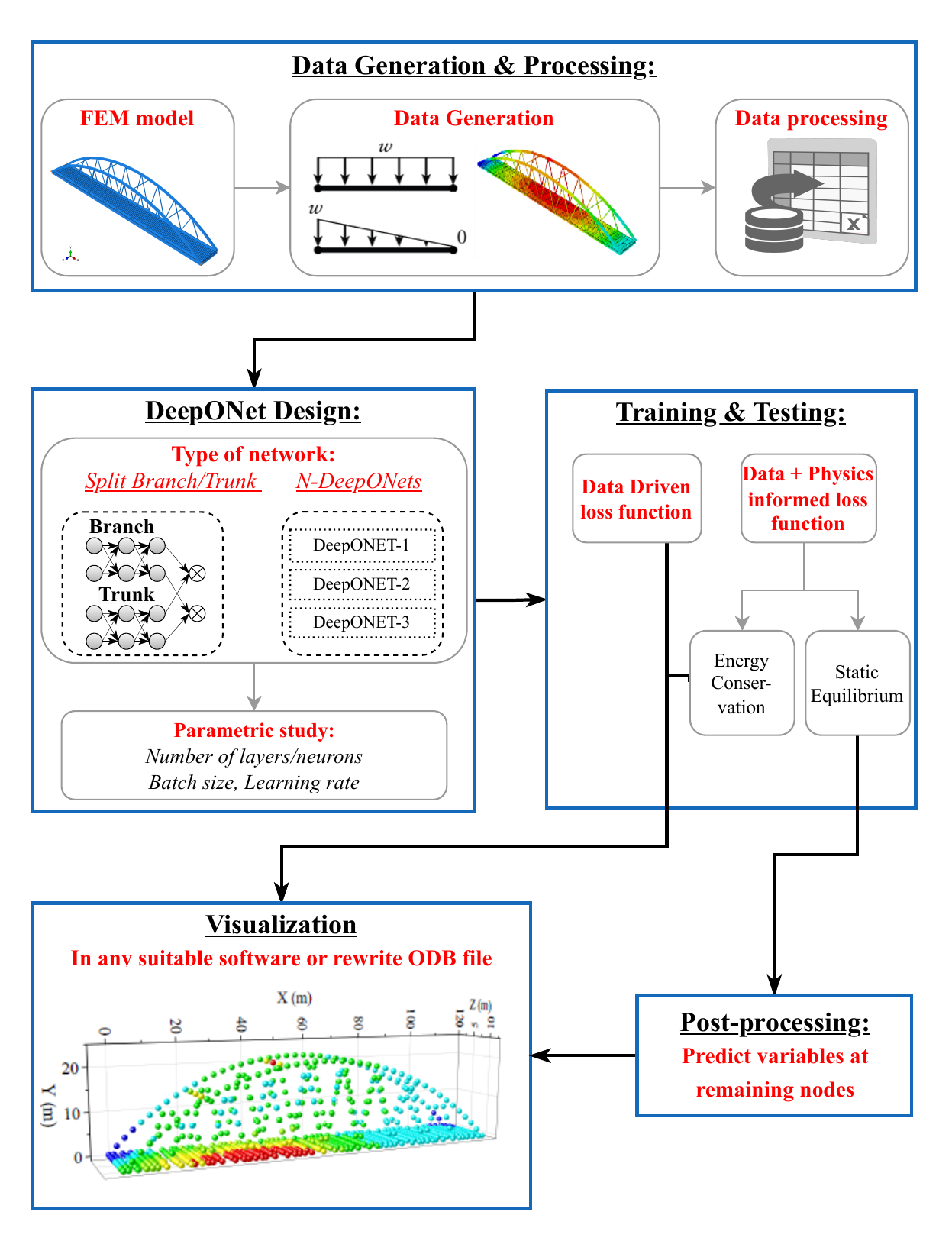}
    \caption{Flowchart of the proposed methodology} 
    \label{flowchart}
\end{figure}

The steps involved in the proposed framework (Figure \ref{flowchart}) are outlined as follows:

\begin{enumerate}
    \item \textbf{Data Generation and Processing:}
    \begin{itemize}
        \item \textbf{FEM model development:} Build and validate a FEM model to simulate structural responses.
        \item \textbf{Dataset generation:} Produce datasets under specified loading and boundary conditions.
        \item \textbf{Data formatting:} Extract and format data for network training.
    \end{itemize}

    \item \textbf{DeepONet Design:}
    \begin{itemize}
        \item \textbf{Output strategy:} Choose between split branch/trunk network or multiple DeepONets ($N$-independent DeepONets) to manage multiple outputs.
        \item \textbf{Network configuration:} Perform a parametric study to optimize the network design.
    \end{itemize}

    \item \textbf{Training and Testing:}
    \begin{itemize}
        \item \textbf{Physics-informed training:} Select and apply a suitable physics principle—energy conservation (EC) or static equilibrium with Schur complement (SE-S)—based on the problem type.
        \item \textbf{Model evaluation:} Train the network on known cases and test on unseen scenarios to assess generalization.
    \end{itemize}

    \item \textbf{Post-Processing:}
    \begin{itemize}
        \item \textbf{Domain computation:} Perform post-processing based on the physical principle (SE-S) and/or the FEM model constraints (e.g., master-to-slave node relationships).
    \end{itemize}

    \item \textbf{Visualization:}
    \begin{itemize}
        \item \textbf{Visual output:} Directly visualize DeepONet results or prepare them for software-based visualization (e.g., 3D scatter plots, ODB file rewriting).
    \end{itemize}
\end{enumerate}

\subsection{Deep Operator Neural Network (DeepONet)}\label{DEEPONET}

Recently, Lu et al. \citep{lu2021learning} introduced a groundbreaking operator learning architecture called DeepONet, inspired by the universal approximation theorem for operators \citep{chen1995universal, back2002universal}. DeepONet offers a straightforward and intuitive model architecture that trains quickly and provides a continuous representation of target output functions independent of resolution. A DeepONet comprises two sub-networks: a branch network that encodes the input function $u(x_i)$ at fixed points $x_i, i = 1, \ldots, m$, and a trunk network that encodes the locations $a$ for the real value output function $G(u)(a)$, where $G$ is an operator acting on the input function $u$, producing the output function $G(u)$. In our case, the branch network takes the loading function as input and outputs a feature embedding $[b_1, b_2, \ldots, b_n]$. The trunk network takes the coordinates $a$ as input and outputs a feature embedding $[t_1, t_2, \ldots, t_n]$. The final output is obtained by the dot product of the outputs of the branch and trunk networks, followed by the addition of a bias term (Figure \ref{deeponet_fig}\textbf{C.}). Multiple architectures of the branch and trunk networks can be used such as MLP (Multi-Layer Perceptron), Convolutional Neural Networks (CNN), or Recurrent Neural Networks (RNN)\citep{lu2021learning}. As in this work, our input (coordinates and load values) and output data (displacements and rotations at each coordinate) are in the Cartesian coordinate system, we choose MLP as the base structure for both the branch and trunk networks (Figure \ref{deeponet_fig}\textbf{C.}).

There are a few variations of DeepONet architectures. One variation features multiple branch networks and a single trunk network to predict the output function, known as stacked DeepONet (Figure \ref{deeponet_fig}\textbf{A.}), typically used when multiple functions combine to form one output. The other type uses a single branch network and a single trunk network, known as unstacked DeepONet (Figure \ref{deeponet_fig}\textbf{B.}). In our work, we focus on mapping a single function (load) to output functions (displacements and rotations). Therefore, we adopt the unstacked DeepONet architecture. Additionally, we aim to predict multiple variables across the entire domain. To achieve this, we utilize strategies outlined by Lu et al. \citep{lu2022comprehensive} for handling multiple output functions:

\begin{figure}[!htb]
    \centering
    \includegraphics[width=0.85\textwidth]{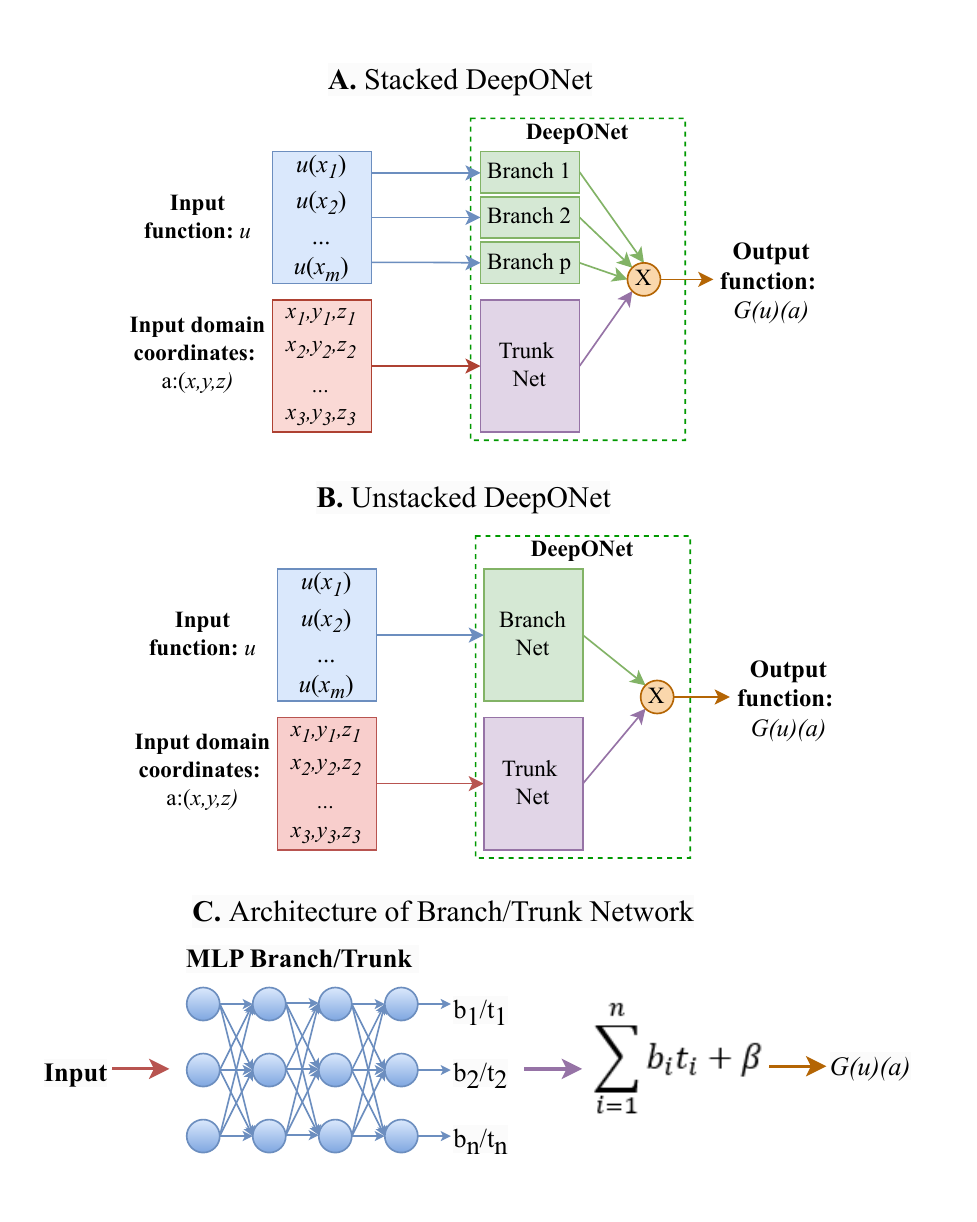}
    \caption{Illustration of DeepONet structure and types. \textbf{A.} Stacked DeepONet, \textbf{B.} Unstacked DeepONet, \textbf{C.} Multilayer Perceptrons(MLP) architecture of Branch/Trunk Network} 
    \label{deeponet_fig}
\end{figure}

\begin{figure}[!htb]
    \centering
    \includegraphics[width=0.8\textwidth]{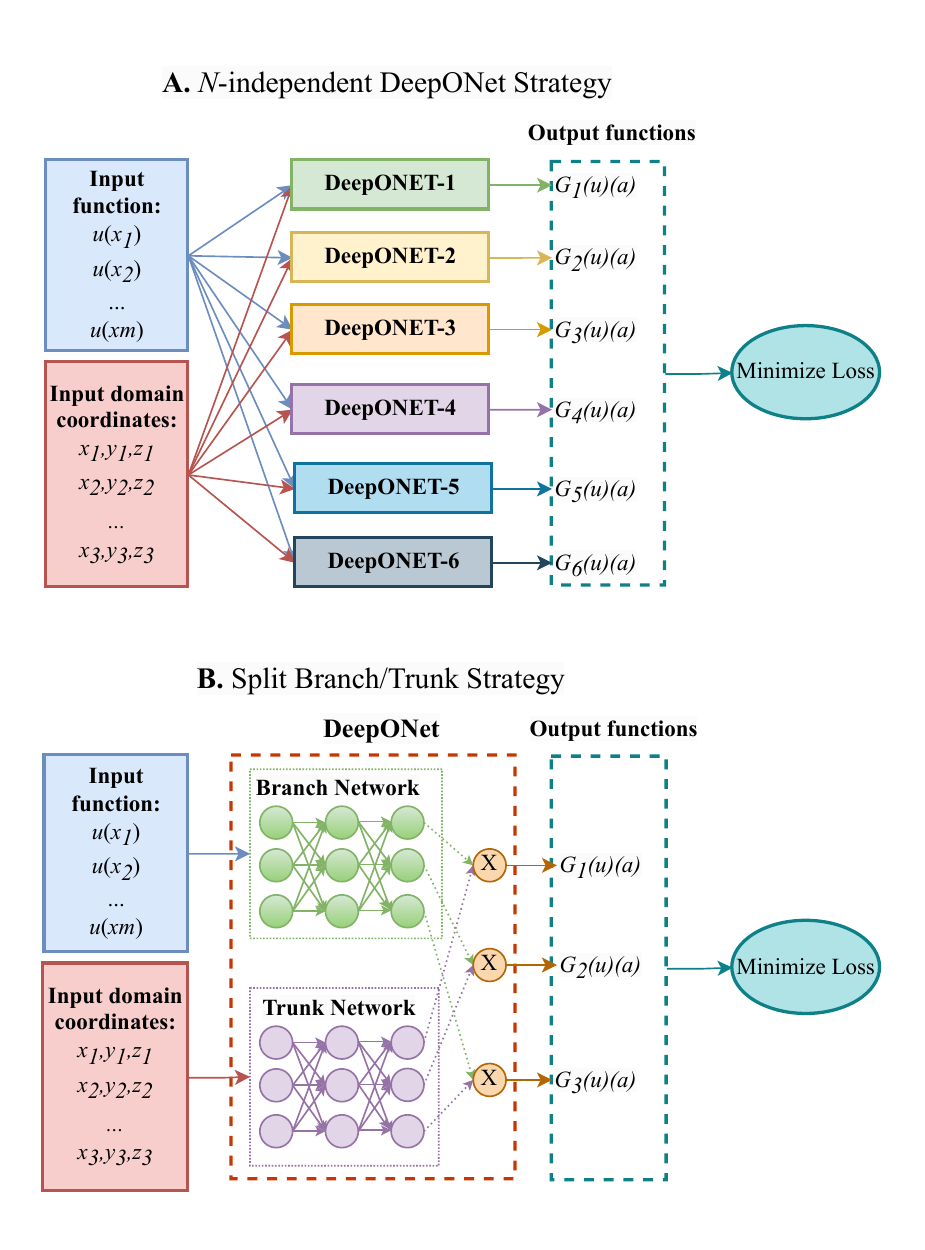}
    \caption{Illustration to handle multiple outputs using DeepONet: \textbf{A.} 6 Independent DeepONets to handle 6 outputs, \textbf{B.} 1 DeepONet to handle 3 outputs.} 
    \label{Multiple_Statergy}
\end{figure}

\begin{itemize}
    \item \textbf{$N$-independent DeepONets:} This method involves using $N$-independent DeepONets, each responsible for outputting one function (Figure \ref{Multiple_Statergy}\textbf{A.}). For example, to predict six functions, we use six separate DeepONets. The weight updates during training for one output function do not impact the weights of the other outputs, as each output is managed by a separate DeepONet.

    \item \textbf{Split branch/trunk network:} This method divides the outputs of both the branch and trunk networks into $N$ groups, with each group predicting one function (Figure \ref{Multiple_Statergy}\textbf{B.}). For instance, to predict three variables ($N = 3$), if both networks have 48 neurons in the output layer, the dot product of the first 16 neurons of both networks produces the first function, the next 16 neurons produce the second function, and the remaining 16 neurons produce the third function. In this approach, the neurons' weights are shared, so changes to one neuron's weight can affect the outputs of other groups.
\end{itemize}

The choice of the optimal strategy depends on the specific problem \citep{lu2022comprehensive}. In our work, we employ both strategies to manage multiple outputs.

\subsection{Loss functions}\label{New_loss_Section}

The choice of loss function significantly impacts the performance of the ML network, as it is a critical component that influences training. The calculated loss value propagates back through the network to update the weights of the neurons, with the specific goal of minimizing the loss. In this work, we use a combination of data-driven and physics-informed losses, expressed as:

\begin{equation}
\begin{split} \label{Loss}
\mathcal{L}_{\text{Total}} = \mathcal{L}_{\text{data}} + \mathcal{L}_{\text{physics}}\\
\end{split}
\end{equation}

In this work, two novel physics-informed loss functions are developed based on a pre-calculated stiffness matrix, thus avoiding the use of PDE-based loss functions.

\subsubsection{Data-Driven loss function}\label{DD_Loss}

Data-Driven (DD) loss functions are commonly used to ensure that the ML network learns based on the provided data \cite{ahmed2022seismic,xu2020prediction,lei2019fault,xu2019automatic}. These functions ensure that the ML network's training adheres strictly to the available data, with the quality of predictions being dependent on the data used during training. In this work, the DD loss function chosen is the mean relative \( L_2 \) loss, defined as:

\begin{equation}
\begin{split} \label{Loss_data}
\mathcal{L}_{\text{data}} = \frac{1}{N} \sum_{i=1}^N \frac{\|G_{\text{True}}(u_i)(a) - G(u_i)(a)\|}{\|G_{\text{True}}(u_i)(a)\|}
\end{split}
\end{equation}

Here, $G(u_i)(a)$ represents the predicted output function value at point $a$ (coordinate), and $G_{\text{True}}(u_i)(a)$ is the true function value (from FEM) at the same point $a$. The parameter $N$ denotes the total number of functions processed by the branch network. This loss function aims to minimize the error between predicted values and actual values.

While this DD approach ensures the network learns from the data, it is also crucial to incorporate the underlying physical principles of the system to produce realistic predictions. Purely DD loss functions may sometimes fail to ensure that predictions adhere to the governing physics, especially when the training data is of insufficient quality or quantity. To address this, we propose two novel physics-informed loss functions. 

\subsubsection{Physics-Informed loss functions}\label{PI_Loss}

Researchers have integrated physics into DD approaches, leading to the development of PINNs (\cite{raissi2019physics, abueidda2021meshless}). Physics-informed models ensure that predictions from machine learning do not rely solely on available training data but also adhere to the underlying physics of the problem, resulting in more realistic and accurate predictions. Generally, the incorporation of physics in training is achieved by formulating the loss function as the residual of PDEs at designated collocation points within the physical domain and its boundary and initial conditions \cite{niaki2021physics, henkes2022physics, rao2021physics}. An alternative approach is based on variational and energy methods, where the loss function is constructed using a variational form \cite{yu2018deep, liao2019deep, samaniego2020energy}. However, for real-life complex structures consisting of multiple types of members (beams and shells) in a discrete domain, the development and incorporation of PDE-based or energy method-based approaches for the entire problem is computationally expensive and challenging. Therefore, we introduce a new approach that leverages the structural stiffness matrix for the development of a physics-informed loss function. Physics can be enforced using two approaches: 1) the energy conservation principle and 2) the static equilibrium equation.

The energy conservation principle is a fundamental concept in physics, ensuring the system's equilibrium by balancing external work with internal work. In the case of elasticity, the principle ensures energy conservation under isothermal conditions where no dissipative processes occur (e.g., plasticity, damage, fracture). The balance of internal work to external work can be expressed as follows:

\begin{equation}
\begin{split} \label{Wext_Lab}
{W}_{\text{internal}} - {W}_{\text{external}} = 0
\end{split}
\end{equation}

In the context of FEM, energy conservation in quasi-static loading can be expressed as:

\begin{equation}
\begin{split} \label{WD_Eq}
\mathit{U}^{\textit{T}} \mathit{K} \mathit{U} - \mathit{U}^{\textit{T}} \mathit{F} = 0
\end{split}
\end{equation}

where \(\mathit{U}\) represents the degrees of freedom (DOFs: displacements and rotations), which are also the predicted output function values \(G(u_i)(a)\) from DeepONet at given coordinates; \(\mathit{K}\) is the global stiffness matrix; and \(\mathit{F}\) is the global force acting on the system. The rationale for using the energy conservation principle as a loss function lies in the straightforward accessibility of the global stiffness matrix from the FEM model of the system. With the global stiffness matrix, we can easily compute the external and internal work done during network training. The objective is to minimize the difference between the external work done and the internal work done, represented by the EC loss function (\(\mathcal{L}_{\text{EC}}\)):

\begin{equation}
\begin{split} \label{WDLoss}
\mathcal{L}_{\text{EC}} = \frac{1}{N} \sum_{i=1}^N \|G(u_i)(a)^{\textit{T}} \cdot K_i \cdot G(u_i)(a) - G(u_i)(a)^{\textit{T}} \cdot F_i\|
\end{split}
\end{equation}

The developed \(\mathcal{L}_{\text{EC}}\) ensures energy conservation across the entire system, but this adds computational complexity, as matrix multiplication involving thousands of DOFs can be challenging and lead to prolonged training times. To overcome this, we introduce the Schur complement in the static equilibrium equation, reducing the problem domain for DeepONet training. This aims to reduce training time and enhance accuracy while preserving the characteristics of the stiffness relationships.

The Schur complement \cite{CARLSON1986257} is a mathematical approach in linear algebra used to reduce the dimensionality of systems of equations, making them easier to solve. It is particularly valuable in the Finite Element Tearing and Interconnecting (FETI) method, where it reduces the problem size by dividing the computational domain into subdomains, solving them independently, and managing interface conditions \cite{langer2005coupled,pechstein2012finite}. In the Boundary Element Method (BEM), the Schur complement simplifies boundary integral equations by expressing them in block matrix form \cite{mobasher2016adaptive,hackbusch2005direct}. In this work, we utilize the Schur complement to reduce the dimensionality of the equilibrium equation and use it as a novel loss function. Typically, the static equilibrium equation for a structural system is expressed as:
\begin{equation}
\begin{split} \label{STiffness}
\mathit{K}\mathit{U} = \mathit{F} \\
\end{split}
\end{equation}

Our goal is to reduce the size of the system using the Schur complement. First, we partition the matrix into blocks as follows:

\begin{equation}
\begin{split} \label{Matrix}
\mathit{K} = \begin{bmatrix}
\mathit{K}_{II} & \mathit{K}_{IN} \\
\mathit{K}_{NI} & \mathit{K}_{NN}
\end{bmatrix}, \quad
\mathit{U} = \begin{bmatrix}
\mathit{U}_I \\
\mathit{U}_N
\end{bmatrix}, \quad
\mathit{F} = \begin{bmatrix}
\mathit{F}_I \\
\mathit{F}_N
\end{bmatrix}\\
\end{split}
\end{equation}

\(\mathit{U}_I\) are selected DOFs, for which we are interested in obtaining results using DeepONet, \(\mathit{U}_N\) are remaining DOFs, which can be obtained from post-processing analysis. By expanding and reorganizing the above system of equations, we get:

\begin{equation}
\begin{split} \label{Matrix_1}
\mathit{K}_{II} \mathit{U}_I + \mathit{K}_{IN} \mathit{U}_N = \mathit{F}_I 
\end{split}
\end{equation}
\begin{equation}
\begin{split} \label{Matrix_2}
\mathit{U}_N = \mathit{K}_{NN}^{-1} (\mathit{F}_N - \mathit{K}_{NI} \mathit{U}_I)
\end{split}
\end{equation}

Substituting the Eq. \ref{Matrix_2} in Eq. \ref{Matrix_1}, we obtain the reduced order equation:

\begin{equation}
\begin{split} \label{Processing} 
\mathit{K}_{II} \mathit{U}_I + \mathit{K}_{IN} \mathit{K}_{NN}^{-1} \mathit{K}_{NI} \mathit{U}_I = \mathit{F}_I - \mathit{K}_{IN} \mathit{K}_{NN}^{-1} \mathit{F}_N 
\end{split}
\end{equation}
This can be simplified as $SU_{I}=F_c$. Where, $S$ is the Schur complement matrix defined as $S=\mathit{K}_{II}  + \mathit{K}_{IN} \mathit{K}_{NN}^{-1} \mathit{K}_{NI}$, a combination of sub-matrices ($K_{II}, K_{IN}, K_{NI},$ $K_{NN}$) from the original stiffness matrix ($K$). $F_c$ is the reduced order global force defined as $F_c=\mathit{F}_I - \mathit{K}_{IN} \mathit{K}_{NN}^{-1} \mathit{F}_N $, $F_I$ represents the global forces associated with picked DOFs, and $F_N$ represents the global forces associated with picked DOFs. Here, the objective is to minimize the difference between the reduced order global force ($F_c$) and the product of the Schur complement matrix ($S$) with selected DOFs \(\mathit{U}_I\), which is represented by the SE-S loss function ($\mathcal{L}_{\text{SE-S}}$):

\begin{equation}
\begin{split} \label{SchurLoss}
\mathcal{L}_{\text{SE-S}} = \frac{1}{N} \sum_{i=1}^N \|S_i \cdot G(u_i)(a) - F_{c,i}\|
\end{split}
\end{equation}

\begin{figure}[!htb]
    \centering
    \includegraphics[width=1\textwidth]{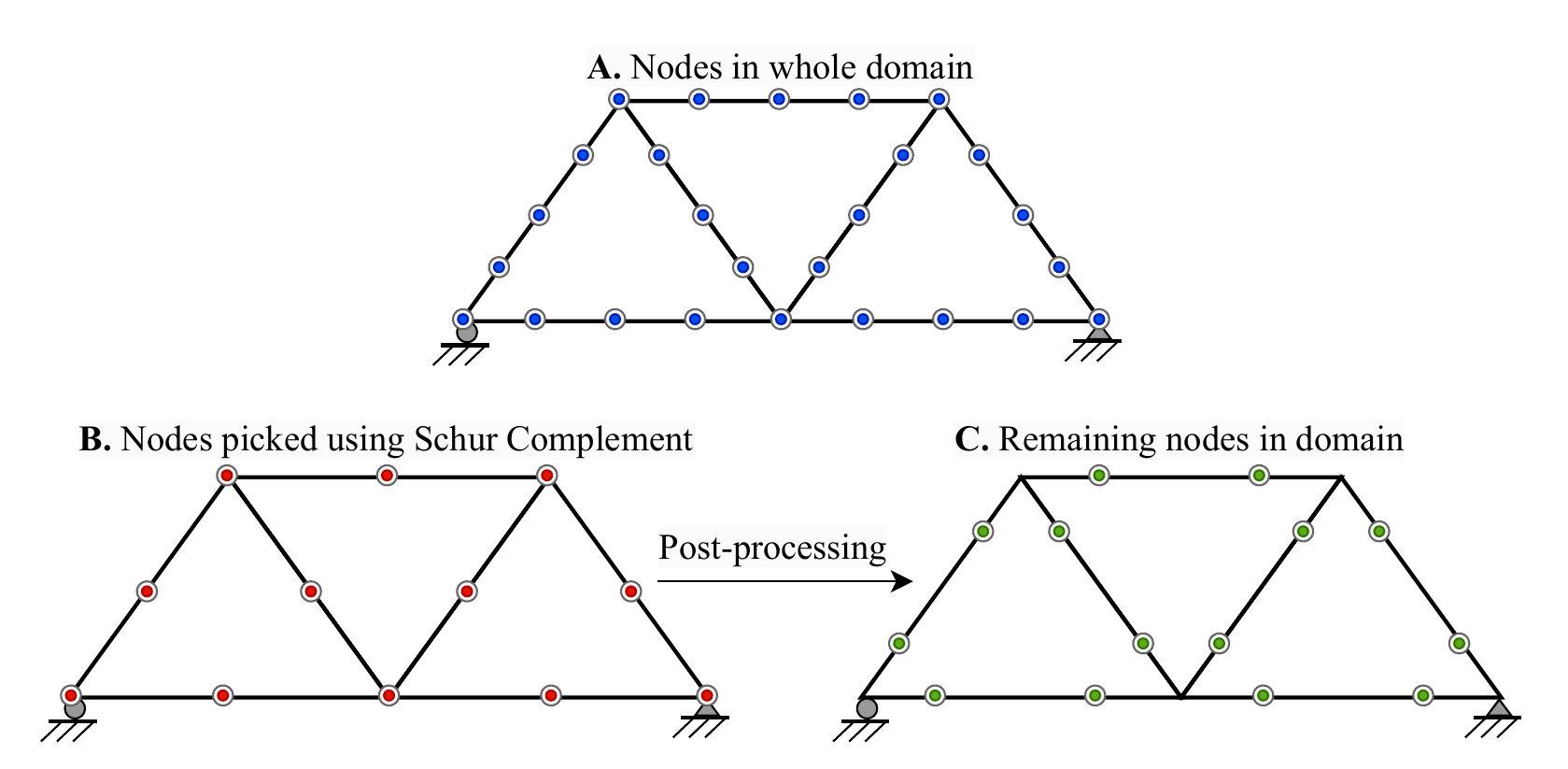}
    \caption{Illustration of the Schur Complement to reduce the size of the system: \textbf{A.} Total nodes in the structural domain, \textbf{B.} Picked nodes by applying Schur complement, \textbf{C.} The remaining nodes solution obtained by using post-processing (Eq. \ref{Matrix_2})} 
    \label{Schur_Decomp}
\end{figure}

Adopting the $\mathcal{L}_{\text{SE-S}}$ offers significant advantages. By reducing a large complex system of equations into a much smaller system, we can incorporate physical constraints into the DeepONet training process more efficiently. This reduction significantly decreases the number of domain points (mesh), allowing the network to learn faster and more effectively. The illustration of the reduced structural domain points using the Schur complement can be found in Figure \ref{Schur_Decomp}. Using the Schur complement, we can train the network on a smaller domain and obtain the solution for the entire domain through post-processing (Eq. \ref{Matrix_2}). Instead of applying the \( \mathcal{L}_{\text{EC}} \) to the entire system, we can achieve the same results with a smaller system, ensuring that the network learns the underlying physics even with fewer domain points.

\subsubsection{Combinations of loss functions}\label{Combi_Loss}
This section outlines the loss function combinations used in Sections \ref{toybridge} and \ref{KW51_Sec}. Eq. \ref{eq:DDEC} integrates the mean \( L_2 \) relative error with the EC loss function, ensuring energy conservation throughout the entire domain. Eq. \ref{eq:SESchur} combines the mean \( L_2 \) relative error with the SE-S loss function, ensuring the network satisfies the equilibrium equation even with a reduced number of domain points. The full-domain solution is then obtained through post-processing using Eq. \ref{Matrix_2}.

\begin{equation}
\begin{split} 
\label{eq:DDEC}
\mathcal{L}_{\text{DD+EC}} = \frac{1}{N} \sum_{i=1}^N \biggl( 
\frac{\|G_{\text{True}}(u_i)(a) - G(u_i)(a)\|}{\|G_{\text{True}}(u_i)(a)\|} +  \\
 \|G(u_i)(a)^{\textit{T}} \cdot K_i \cdot G(u_i)(a) - 
G(u_i)(a)^{\textit{T}} \cdot F_i\| \biggr)
\end{split}
\end{equation}

\begin{equation}
\begin{split} \label{eq:SESchur}
\mathcal{L}_{\text{DD+SE-S}} = \frac{1}{N} \sum_{i=1}^N \left( 
\frac{\|G_{\text{True}}(u_i)(a) - G(u_i)(a)\|}{\|G_{\text{True}}(u_i)(a)\|} + 
\|S_i \cdot G(u_i)(a) - F_{c,i}\| \right)
\end{split}
\end{equation}

\section{2d beam structure} \label{toybridge}
\subsection{FEM Model \& Data Generation}
\begin{figure}[h]
    \centering
    \includegraphics[width=1.0\textwidth]{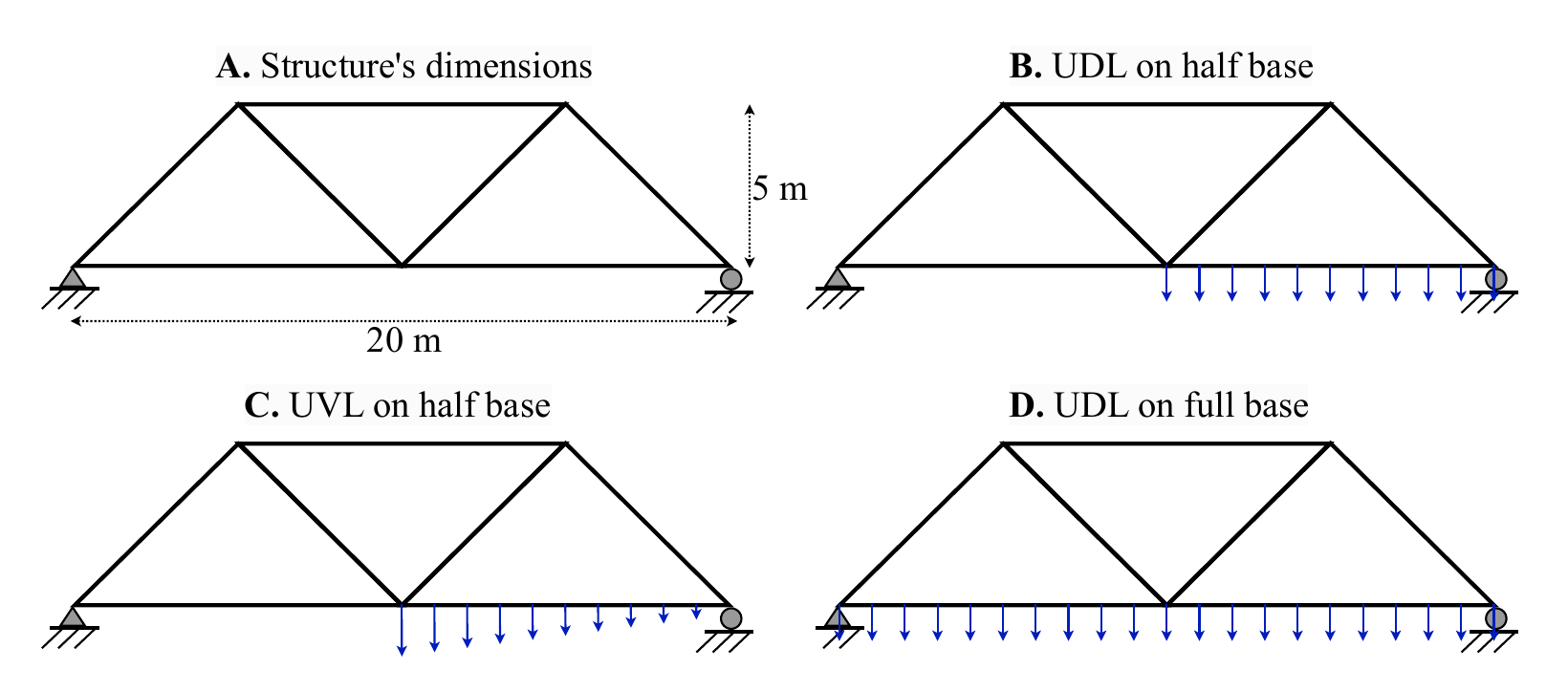}
    \caption{Illustration of the 2D beam structure and loading scenarios: \textbf{A.} FEM nodes, \textbf{B.} 1st loading scenario: UDL on half base, \textbf{C.} 2nd loading scenario: UVL on half base, \textbf{D.} 3rd loading scenario: UDL on full base}
    \label{ToyBridge}
\end{figure}

To evaluate the proposed method for response prediction, a 2D beam structure is modeled using Abaqus with 2D Timoshenko beam elements configured to resemble a truss structure. The structure's dimensions are 20 meters horizontally and 5 meters vertically, with boundary conditions applied as hinge (bottom left) and roller supports (bottom right) (Figure \ref{ToyBridge}\textbf{A.}). The model consists of 56 nodes, each providing outputs for displacement in the $x$ and $y$ directions and rotation about the $z$ direction ($U_x$, $U_y$, and $R_z$). The steel material properties used in the model are as follows: Young's modulus of 210 GPa, a Poisson's ratio of 0.3, and a density of 7850 kg/m³. The beam section is rectangular, measuring 400 mm by 250 mm.

Three loading scenarios are applied vertically downward as depicted in Figure \ref{ToyBridge}. The first scenario involves a uniformly distributed load (UDL) applied to half of the base, the second scenario features a uniformly varying load (UVL) over half of the base, and the third scenario applies a UDL across the entire base. Each scenario includes 1016 random loadings ranging from 0.1 kN/m to 15 kN/m, resulting in a total of 3048 cases. The resulting displacements and rotations at each node were extracted from the ODB files and formatted for compatibility with DeepONet. The DeepONet dataset consists of three components: branch input ($3048\times21$), trunk input ($56\times2$), and output ($3048\times168\times3$). Here, 3048 represents the total number of samples, 21 denotes the load values at the nodes, $56\times2$ corresponds to the $x$ and $y$ coordinates for each node, and the output contains three variables ($Ux$, $Uy$, $Rz$) for each node within each sample.

\subsection{DeepONet Training}

\begin{figure}[!htb]
    \centering
    \includegraphics[width=0.9\textwidth]{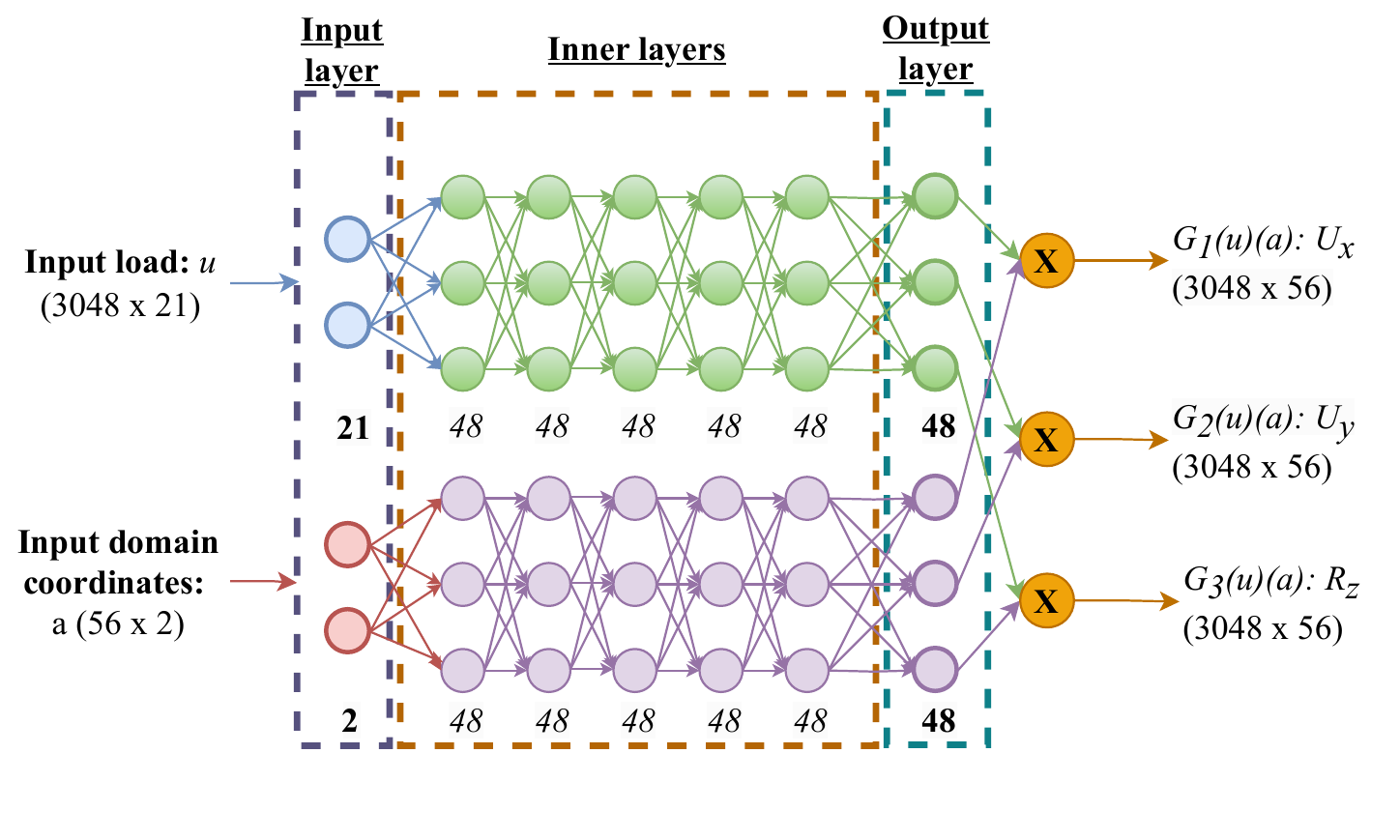}
    \caption{DeepONet architecture for the 2D beam structure}
    \label{Toy_DeepONet}
\end{figure}

We train the unstacked DeepONet (Figure \ref{Toy_DeepONet}) using Eq. \ref{Loss_data} (DD) and Eq. \ref{eq:DDEC} (DD+EC). The EC loss function, which incorporates the principle of energy conservation, requires the use of the structure's stiffness matrix. For the 2D beam structure with 56 nodes and 3 degrees of freedom per node, this results in a $168\times168$ global stiffness matrix. Utilizing the EC loss function, which aims to minimize the discrepancy between external and internal work, introduces additional complexity and necessitates more epochs for training compared to using the DD loss function alone. However, to facilitate a fair comparison between the DD and DD + EC approaches, we maintained a consistent number of epochs. The network is trained using the ADAM optimizer with a learning rate of 0.001, a default initializer, and an activation function of DeepXDE \citep{lu2021deepxde}. Prior to training, a comprehensive parametric study was conducted to ensure optimal accuracy and mitigate issues of underfitting or overfitting.

\subsection{Parametric Study of Network Design}
To handle multiple output variables from DeepONet, we employ the split branch/trunk network strategy (Figure \ref{Multiple_Statergy}\textbf{B.}). Since there are three output functions ($U_x$, $U_y$, and $R_z$), the output layers in both networks must contain a multiple of three neurons. A parametric study is performed to identify the optimal configuration of layers, neurons, and batch size while keeping the dataset and training/testing ratio constant. This study utilizes the DD loss function. The summarized findings of the parametric study are given in Table \ref{tab:parametric_study}.

\subsubsection{Selection of Number of Neurons}
Initially, both the branch and trunk networks are set to six layers. The network's performance is then assessed by varying the number of neurons in the output layer [6, 12, 18, 24, 30, 36, 42, 48] and adjusting the neurons in the inner layers [6, 12, 18, 24, 30, 36, 42, 48] correspondingly. For each fixed number of output layer neurons, the inner layer neurons are varied from 6 to 48 to evaluate their impact on performance. Figures \ref{Parametric}\textbf{A.} and \ref{Parametric}\textbf{B.} display results for output layer neurons ranging from 36 to 48. Based on these findings, we choose 48 neurons for both the inner and output layers (Figure \ref{Toy_DeepONet}).

\subsubsection{Number of Layers}
Next, the total number of layers in both the branch and trunk networks is varied from 2 to 8, while maintaining 48 neurons per layer. As shown in Figure \ref{Parametric}\textbf{C.}, the best performance is observed with 6 or 7 layers. Based on these findings, we select a network configuration with 6 layers for both the branch and trunk networks.

\subsubsection{Final Selection Based on Aspect Ratio}
The aspect ratio, defined as the ratio of a network's width (number of neurons per layer) to its depth (number of layers), plays a critical role in model performance. A wider network can capture more complex features, while a deeper network excels at learning hierarchical features. Striking a balance is essential to avoid underfitting (too narrow/shallow) or overfitting (too wide/deep) \cite{pantidis2023error}. In this study, we fixed the number of neurons in the inner layers at 240 and varied the number of layers [24, 20, 16, 12, 8, 6, 5, 4, 3] and neurons per layer [10, 12, 15, 20, 30, 40, 48, 60, 80] to achieve aspect ratios ranging from [0.416, 0.6, 0.937, 1.667, 3.75, 6.667, 9.60, 15.0, 26.67]. Based on the results (Figure \ref{Parametric}\textbf{D.}), we selected an aspect ratio of 9.6, leading to the final network configuration (Figure \ref{Toy_DeepONet}) with a branch network [\textbf{21}\textit{, 48, 48, 48, 48, 48,} \textbf{48}] and a trunk network [\textbf{2}\textit{, 48, 48, 48, 48, 48,} \textbf{48}], where bold indicates the input and output layers and italics denote the 5 inner layers.

\subsubsection{Batch Size Selection}
\begin{figure}[!htb]
    \centering
    \includegraphics[width=1\textwidth]{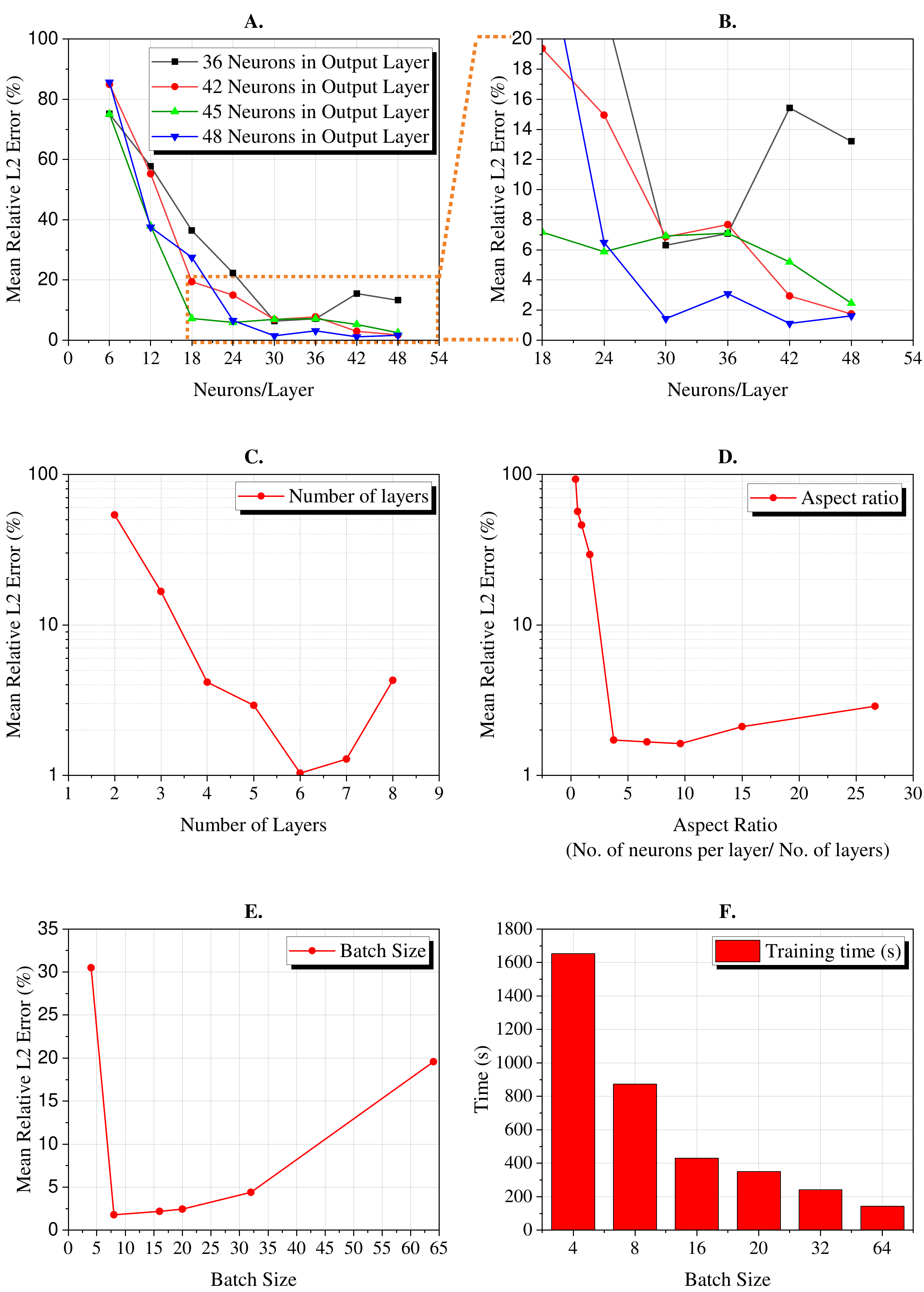}
    \caption{Parametric study: \textbf{A.} \& \textbf{B.} Relative error for neurons per layer, \textbf{C.} Relative error vs. number of layers in the network, \textbf{D.} Relative error based on the network's aspect ratio, \textbf{E.} Relative error based on batch sizes, \textbf{F.} Training time for 1000 epochs with different batch sizes}
    \label{Parametric}
\end{figure}

With the final network configuration, we tested batch sizes of [4, 8, 16, 20, 32, 64], while keeping the number of epochs fixed at 2,500. As shown in Figures \ref{Parametric}\textbf{E.}, batch sizes of 8, 16, and 20 yielded the best accuracy, whereas batch sizes of 16, 20, 32, and 64 resulted in less training times (Figure \ref{Parametric}\textbf{F.}). To maintain a balance between accuracy and training efficiency, we selected a batch size of 20 for this study.

\begin{table}[h!]
\centering
\caption{Summary of parametric study on network design}
\begin{tabular}{|>{\centering\arraybackslash}m{2.5cm}|>{\centering\arraybackslash}m{3cm}|>{\centering\arraybackslash}m{3cm}|>{\centering\arraybackslash}m{4cm}|}
\hline
\textbf{Parameters} & \textbf{Tested values} & \textbf{Optimal selection} & \textbf{Key findings} \\ \hline
\textbf{Neurons per layer} & [6, 12, 18, 24, 30, 36, 42, 48] & 48 neurons in each output and inner layers & 48 neurons for both inner and output layers provide the best performance. \\ \hline
\textbf{Number of layers} & [2, 3, 4, 5, 6, 7, 8] & 6 layers in both branch and trunk networks & Optimal performance with 6 layers, balancing model complexity and accuracy. \\ \hline
\textbf{Aspect ratio} & [0.416, 0.6, 0.937, 1.667, 3.75, 6.667, 9.60, 15.0, 26.67] & 9.6 & Aspect ratio of 9.6 yields the best results, providing a balanced network width and depth. \\ \hline
\textbf{Batch size} & [4, 8, 16, 20, 32, 64] & 20 & Batch size of 20 effectively balance accuracy and training time. \\ \hline
\end{tabular}
\label{tab:parametric_study}
\end{table}

\subsection{Results}
\begin{figure}[!htb]
    \centering
    \includegraphics[width=0.95\textwidth]{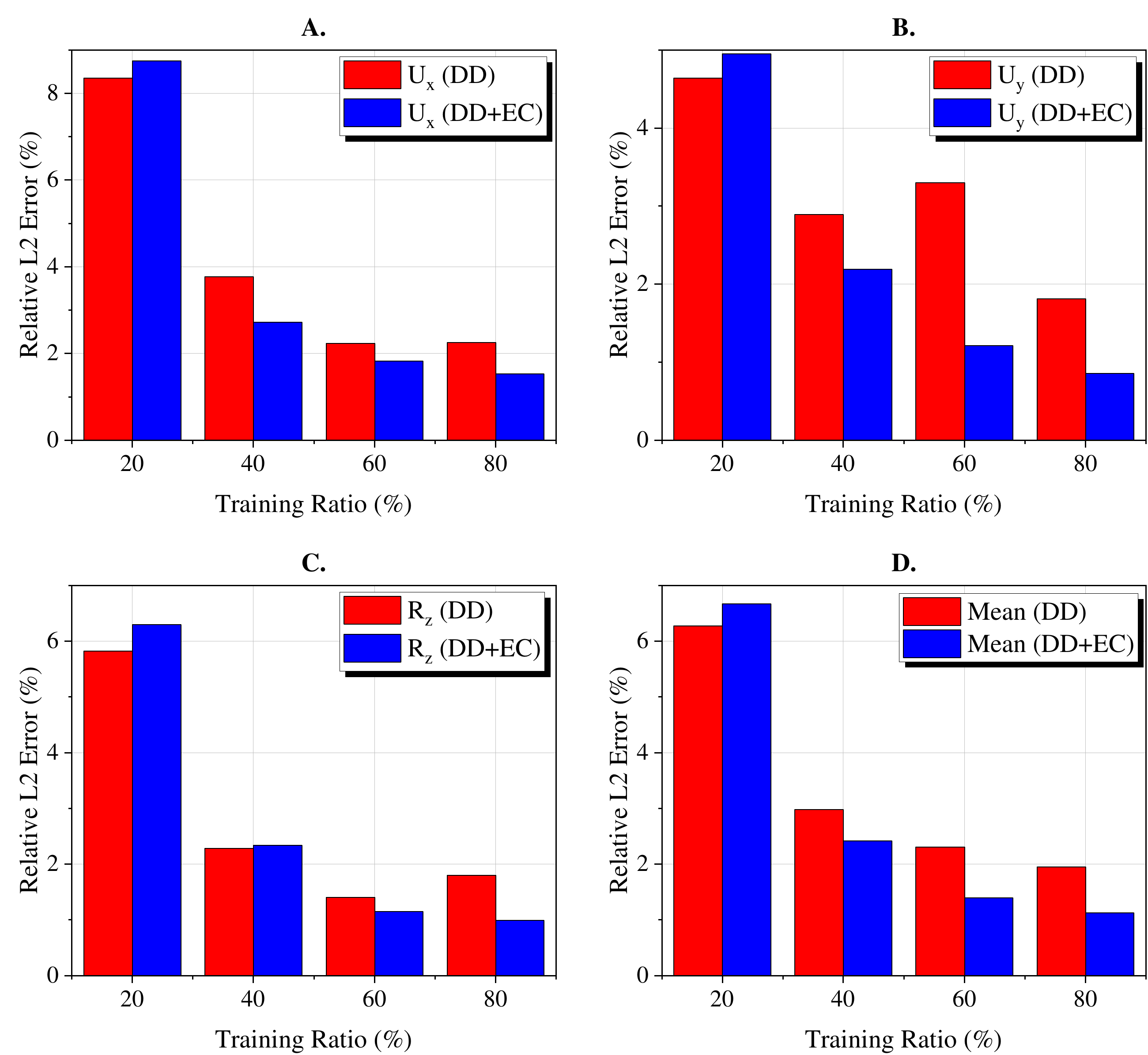}
    \caption{Error comparison of DD (Eq. \ref{Loss_data}) and DD+EC (Eq. \ref{eq:DDEC}) loss: \textbf{A.} Relative error for $U_x$, \textbf{B.} Relative error for $U_y$, \textbf{C.} Relative error for $R_z$, \textbf{D.} Mean relative error for $U_x, U_y$, and $R_z$}
    \label{Toy_Error_Comp}
\end{figure}

\begin{figure}[!htb]
    \centering
    \includegraphics[width=1.0\textwidth]{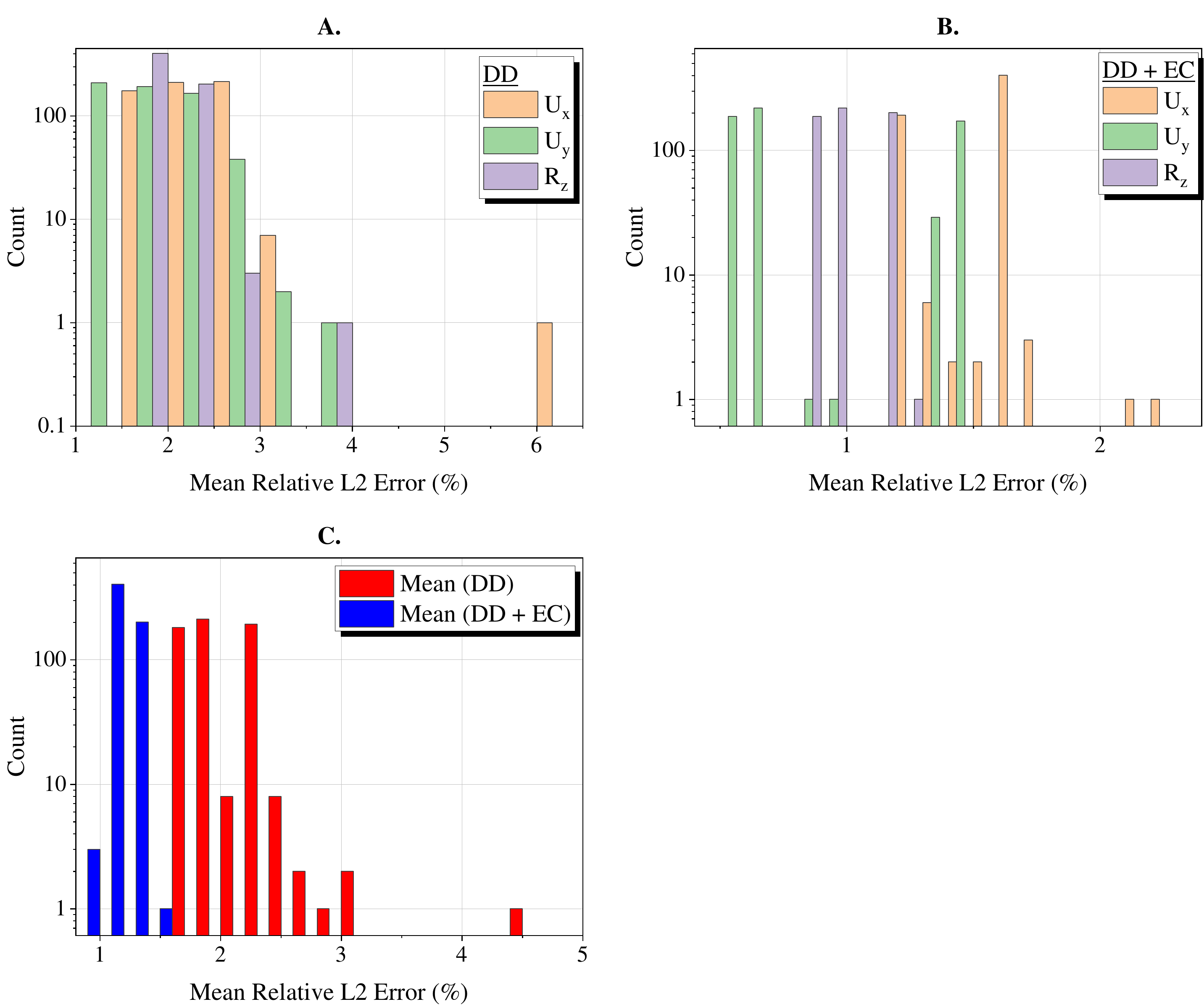}
    \caption{Histogram of relative error for 80\% training ratio: \textbf{A.} Error due to DD loss function (Eq. \ref{Loss_data}), \textbf{B.} Error due to DD + EC loss function (Eq. \ref{eq:DDEC}), \textbf{C.} Mean error for $U_x, U_y$, and $R_z$}
    \label{Toy_Histogram}
\end{figure}

\begin{figure}[!htb]
    \centering
    \includegraphics[width=1.0\textwidth]{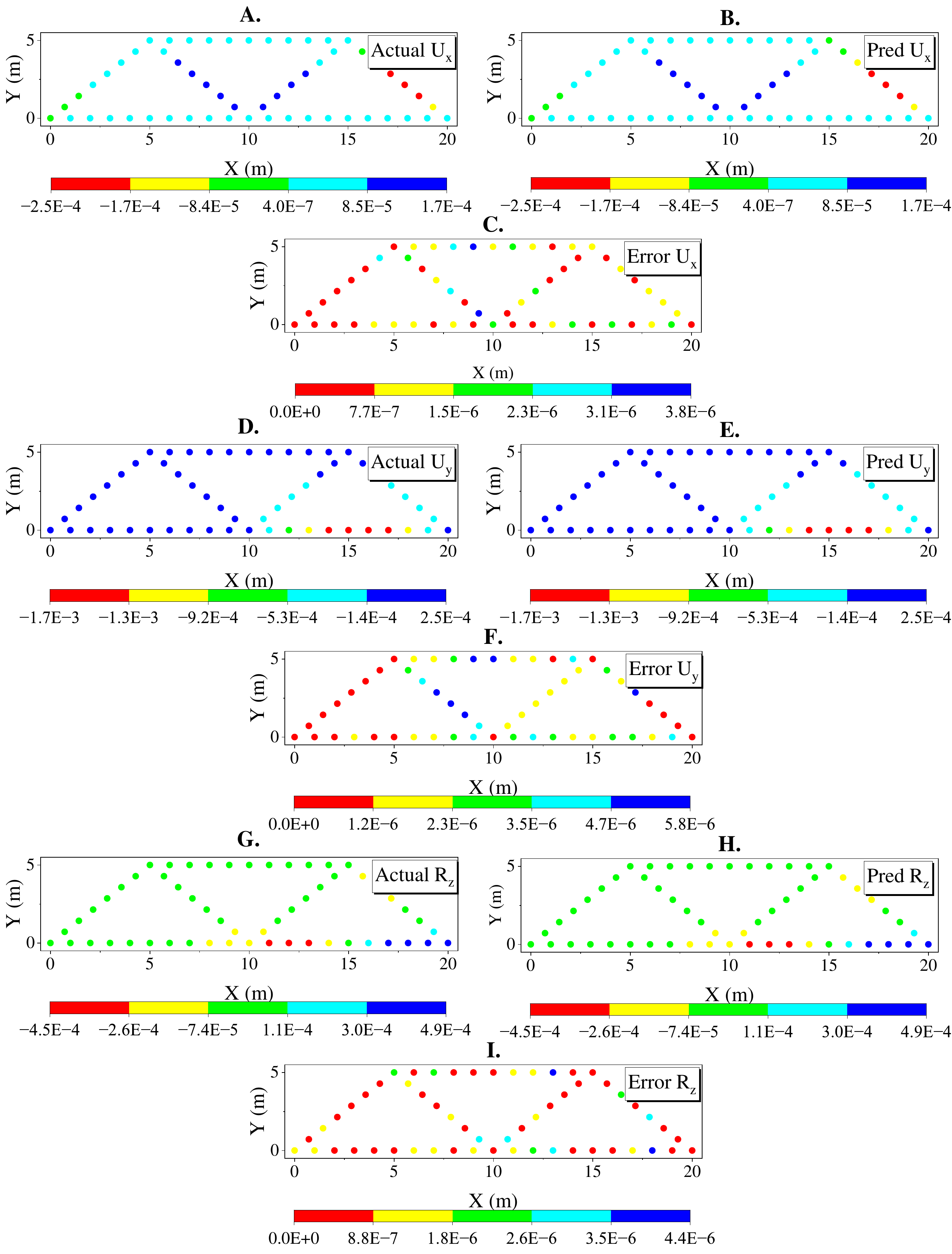}
    \caption{Actual, predicted, and error values for a random sample: \textbf{A.} Actual $U_x$, \textbf{B.} Predicted $U_x$, \textbf{C.} Error for $U_x$, \textbf{D.} Actual $U_y$, \textbf{E.} Predicted $U_y$, \textbf{F.} Error for $U_y$, \textbf{G.} Actual $R_z$, \textbf{H.} Predicted $R_z$, \textbf{I.} Error for $R_z$}
    \label{Toy_Actual_Response}
\end{figure}

To validate the proposed method, we used training ratios of 20\%, 40\%, 60\%, and 80\% of the total dataset. DeepONet performed well even with a 20\% training ratio, demonstrating its capability to learn effectively from smaller datasets. Figure \ref{Toy_Error_Comp} displays the mean error of output variables $(U_x, U_y$, and $R_z)$. Error histograms for the 80\% training data are provided in Figure \ref{Toy_Histogram}. Incorporating the EC loss improved network performance across most cases, except for the 20\% training ratio. The primary reason is that adding the EC loss makes the training process more challenging. When combined with a lower number of epochs and a reduced training ratio, this hinders the network's ability to fully learn. More epochs are needed when using a lower training ratio. Conversely, higher training ratios reduce the relative error, indicating that incorporating EC loss enhances network performance by ensuring predictions adhere to the energy conservation principle. Figures \ref{Toy_Histogram}\textbf{A.} and \textbf{B.} show that errors for $U_x$ are up to 6\% with DD loss, while DD+EC loss errors are under 2.5\%. The maximum mean error for DD+EC loss is 1.6\%, compared to 4.5\% for DD loss (Figure \ref{Toy_Histogram}\textbf{C.}). This improvement highlights the effectiveness of the novel EC loss function. The predicted values of a random sample ($U_x, U_y$, and $R_z$) plotted on the mesh points (Figure \ref{Toy_Actual_Response}) indicate minimal prediction error, demonstrating the promise of the proposed method.

\subsection{Discussion}
The results demonstrate that DeepONet is highly effective in predicting structural responses. The split branch/trunk strategy successfully managed multiple outputs while maintaining accuracy, even with a 20\% training ratio. Increasing the training ratio to 40\%, 60\%, and 80\% further enhances prediction accuracy.

\section{KW-51 bridge} \label{KW51_Sec}
\subsection{Descriptions}

\begin{figure}[h]
    \centering
    \includegraphics[width=0.5\textwidth]{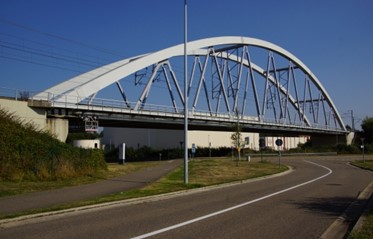}
    \caption{KW-51 railway bridge (\citep{Structurae}) in Leuven, Belgium}
    \label{KW51}
\end{figure}

To evaluate the performance of the proposed method for real-life structures, the KW-51 railway bridge in Leuven, Belgium, is considered (Figure \ref{KW51}). The KW-51 is a steel arch railway bridge of the bow-string type, with a length of 115 meters and a width of 12.4 meters. Located between Leuven and Brussels, it features two ballasted electrified tracks, both of which are curved with radii of 1125 meters and 1121 meters, respectively.

The bridge was monitored for 15 months, from October 2018 to January 2020 \citep{KW51-Monitoring-Frequency}. During this period, 12 accelerometers were installed on the arches and deck, 12 strain gauges on the deck and diagonal members (connecting the arch and deck), 4 strain gauges on the rails, and 2 displacement sensors on the roller supports. Further details on the bridge specifications can be found in \citep{KW51-Monitoring-Frequency}.

\subsection{FEM modeling and validation}

The KW-51 bridge comprises various structural components, including two arches, thirty-two diagonals, four pipe connectors, two main girders, thirty-three transverse beams, twelve stiffeners, one deck plate, one ballast layer, and two rails. In the FEM, the deck plate and ballast layer are modeled using four-node shell elements (S4R), while the remaining members are modeled with two-node Timoshenko beam elements (B31). Specifically, the arches and diagonals are modeled as box sections, the girders and transverse beams as inverted T beams, and the stiffeners as U-shaped members. The section and material properties of all members are detailed in Table \ref{Parameters of KW51}.

\begin{figure}[h]
    \centering
    \includegraphics[width=1.0\textwidth]{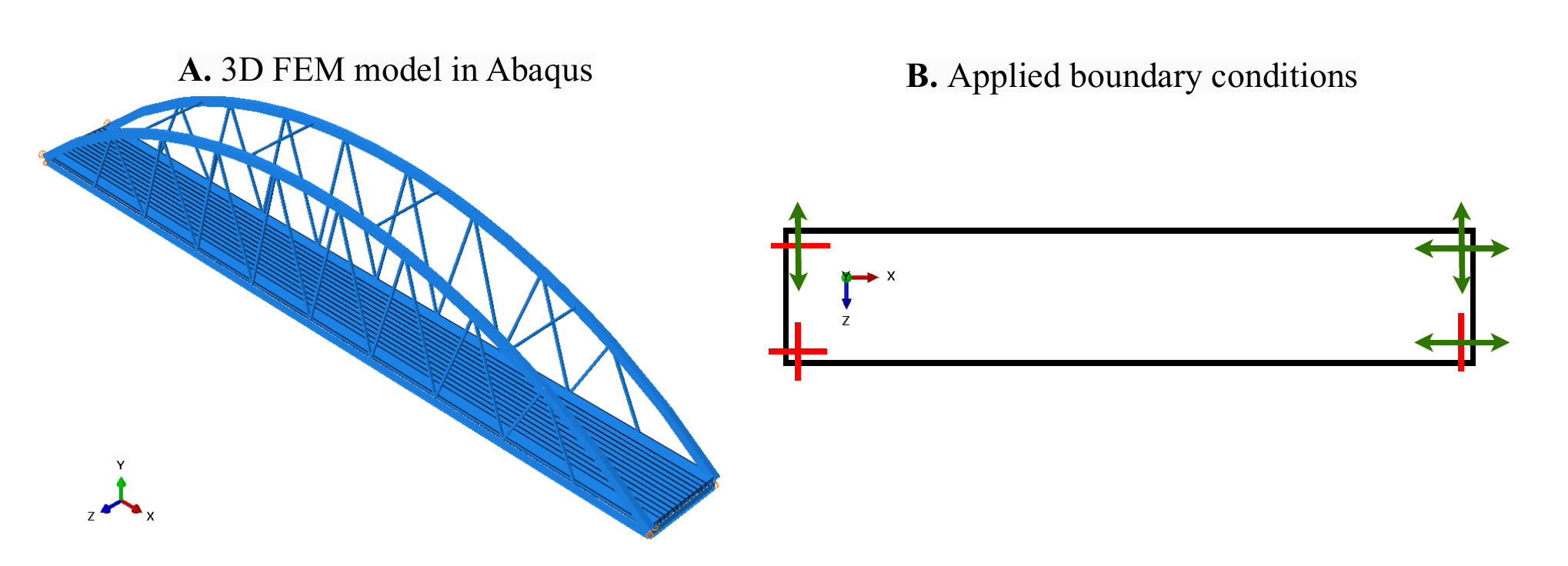}
    \caption{KW-51 modeling \textbf{A.} 3D FEM model in Abaqus, \textbf{B.} Applied boundary conditions in $xz$ plane}
    \label{KW51_modeling}
\end{figure}

All these components are interconnected with tie constraints: the deck plate is tied to the main girders, transverse beams, and U-shaped stiffeners; the ballast layer is tied to the deck plate; and the rail is tied to the ballast layer. Since the focus is on the elastic response, nonlinear effects such as friction or slip between the ballast and the deck are not considered. The actual bridge is supported by four-pot bearings; in the FEM model, boundary conditions are applied as pin and roller supports (Figure \ref{KW51_modeling}). The FEM model includes a total of 1882 nodes, each with six degrees of freedom (DOFs), leading to a total of 11,292 DOFs.

The FEM model is validated based on the natural frequency of the bridge obtained from operational modal analysis (OMA) \citep{KW51-Monitoring-Frequency, KW51-MSSP-stresses}. Six accelerometers were installed on the bridge deck and six on the arches. By collecting long-term data from these accelerometers and performing OMA analysis, the natural frequencies of the structure were measured \citep{KW51-NaturalFrequency-MSSP}. During the monitoring period, 14 natural frequencies of the structure were tracked, based on nearly 3000 OMA analyses. Table \ref{Natural_Freq} presents a comparison between the tracked natural frequencies and those obtained from the FEM model. As shown in Table \ref{Natural_Freq}, the FEM frequencies achieve over 90\% accuracy compared to the measured frequencies, except for the 4th and 5th modes. This discrepancy might be due to the lack of specific structural details, such as the thickness of the web, the thickness of the arch box section, and the width of the arch box section, which were unavailable in the open literature. These values are present in the structure's blueprints, but the authors did not have access to them.  Using design code guidelines, we assumed some values (highlighted in bold in Table \ref{Parameters of KW51}) for the simulation. Despite these assumptions, the FEM frequencies average 93\% accuracy, which is considered sufficient for further analysis.

\begin{table}[]
\centering
\caption{Sectional and material properties used in FEM (Assumed values in \textbf{bold})}
\label{Parameters of KW51}
\begin{tabular}{|c|c|}
\hline
\text{\textbf{Description}} & \text{\textbf{Dimensions}} \\ \hline
\text{Length}&\text{115m}\\
\text{Width} &\text{12.4m} \\
\text{Box arch} &\text{\textbf{0.86m}$\times$1.3m$\times$\textbf{0.045m}} \\
\text{Box diagonal} &\text{0.345m$\times$0.35m$\times$0.016m} \\
\text{Pipe connector} &\text{\textbf{0.2m}$\times$\textbf{0.002m}} \\
\text{Deck thickness} &\text{0.015m} \\
\text{Ballast thickness} &\text{0.6m} \\
\text{U-shape stiffener} &\text{\textbf{0.25m}$\times$\textbf{0.25m}$\times$\textbf{0.008m}} \\
\text{T-shape girder} &\text{\textbf{0.6m}$\times$1.235m$\times$\textbf{0.08m}} \\
\hline
\text{\textbf{Description}} & \text{\textbf{Material properties}} \\ \hline
\text{Steel}&\text{$\rho$ = 7750kg/m$^{3}$, E = 210GPa, v = 0.3}\\
\text{Ballast}&\text{$\rho$ = 1900kg/m$^{3}$ , E = 550MPa, v=0.3}\\
\hline
\end{tabular}
\end{table}

\begin{table}[]
\centering
\caption{Comparison of natural frequency between the FEM and measured \citep{KW51-Monitoring-Frequency} values}
\label{Natural_Freq}
\begin{tabular}{|c|c|c|c|}
\hline
\text{\textbf{Description}} & \text{\textbf{FEM}} & \text{\textbf{Measured}} & \text{\textbf{Accuracy}}\\ \hline
\text{1$^{st}$ lateral mode of the arches}&\text{0.55}&\text{0.51}& \text{92.1\%} \\
\text{2$^{nd}$ lateral mode of the arches} &\text{1.22} & \text{1.23}&\text{99.2\%} \\
\text{1$^{st}$ lateral mode of the bridge deck} &\text{1.73} & \text{1.87}&\text{92.5\%} \\
\text{1$^{st}$ global vertical mode} &\text{2.07} & \text{2.43}&\text{85.1\%} \\
\text{3$^{rd}$ lateral mode of the arches} &\text{2.02} & \text{2.53}&\text{79.8\%} \\
\text{2$^{nd}$ global vertical mode} &\text{2.78} & \text{2.92}&\text{95.2\%} \\
\text{4$^{th}$ lateral mode of the arches} &\text{3.21} & \text{3.55}&\text{90.4\%} \\
\text{1$^{st}$ global torsion} &\text{3.53} & \text{3.90}&\text{90.5\%} \\
\text{3$^{rd}$ global vertical mode} &\text{4.04} & \text{3.97}&\text{98.2\%} \\
\text{2$^{nd}$ global torsion} &\text{4.10} & \text{4.29}&\text{95.5\%} \\
\text{2$^{nd}$ lateral mode of the bridge deck} &\text{4.52} & \text{4.81}&\text{93.9\%} \\
\text{4$^{th}$ global vertical mode } &\text{5.28} & \text{5.31}&\text{99.4\%} \\
\text{3$^{rd}$ global torsion} &\text{6.11} & \text{6.30} &\text{96.9\%}\\
\text{5$^{th}$ global vertical mode} &\text{6.34} & \text{6.83}&\text{92.8\%} \\
 \hline
\end{tabular}
\end{table}

\subsection{Data generation}

\begin{figure}[h]
    \centering
    \includegraphics[width=1.0\textwidth]{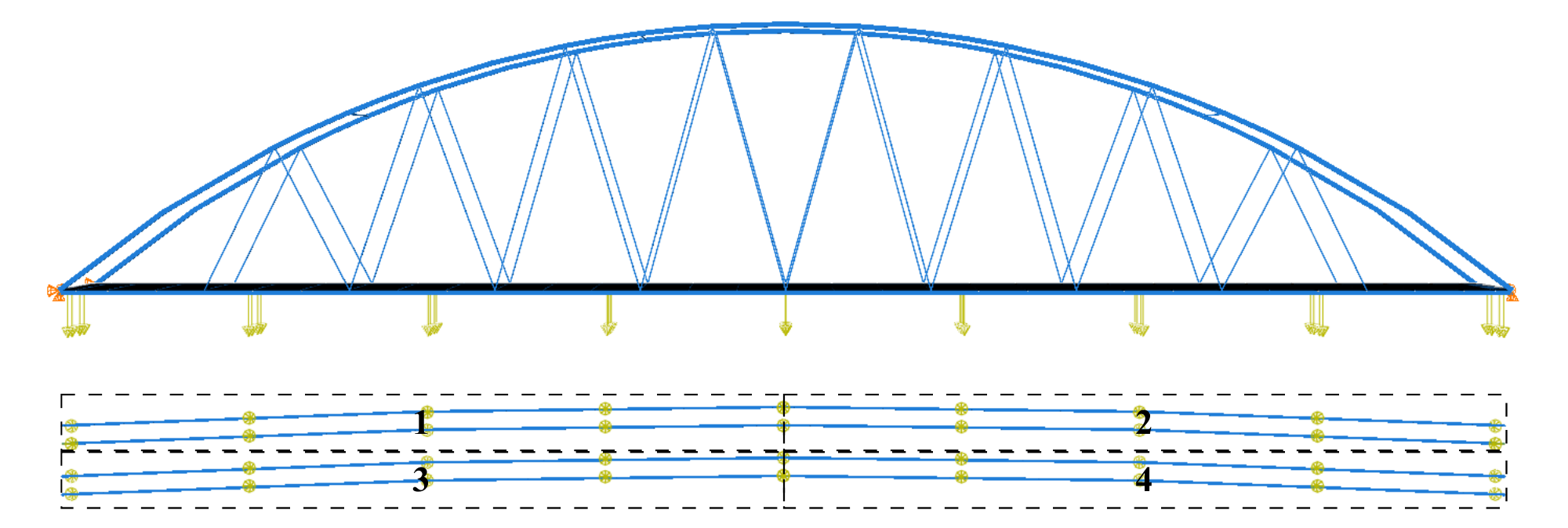}
    \caption{Illustration of the loading areas on the KW-51 railway bridge}
    \label{KW51_loading}
\end{figure}

After validating the model, 15 different loading cases are considered for data generation. These loading cases involve a uniformly distributed load (UDL) applied to the rails. The combinations of loading cases include: [1, 2, 3, 4, 12, 13, 14, 23, 24, 34, 123, 124, 134, 234, and 1234]. For example, "124" indicates that a UDL load is applied simultaneously on the rails in areas 1, 2, and 4 (Figure \ref{KW51_loading}). For each loading case, 600 random samples were generated with loadings ranging from 5 kN/m to 15 kN/m, resulting in a total of 9000 load cases for the entire structure. FEM simulations were performed for these 9000 cases. The results were extracted from the ODB files and processed to make the inputs and outputs compatible with DeepONet. The total data for DeepONet consists of three parts: branch input ($9000\times156$), trunk input ($1882\times3$), and output ($9000\times1882\times6$). Here, 9000 represents the total number of samples, 156 is the number of load values acting on the nodes of the structure, and $1882\times3$ denotes the $x$, $y$, and $z$ coordinates for each node. In the output, we have 9000 samples, and for each of the 1882 nodes, there are six output variables representing displacement and rotation in all directions ($U_x$, $U_y$, $U_z$, $R_x$, $R_y$, and $R_z$). In the next section, these inputs and outputs will be adjusted depending on the specific approach we follow.

\subsection{DeepONet}

To handle multiple outputs, two strategies are considered, as detailed in Section \ref{DEEPONET}. The first strategy uses a split branch/trunk network, similar to the method applied to the 2D beam structure. The second strategy involves a single main network with six independent DeepONet architectures, each dedicated to one of the six output variables.

\subsubsection{Split Branch/Trunk}\label{splitbranch}

Initially, we use the split branch/trunk strategy to handle multiple outputs. This network requires an output layer with neurons in multiples of six. Given the complexity and size of the dataset, which includes 9000 samples, 1882 mesh points, and six output variables per mesh point, we increase the number of neurons in the network. The branch network architecture is [\textbf{156}\textit{, 150, 150, 150, 150, 150,} \textbf{150}], and the trunk network architecture is [\textbf{3}\textit{, 150, 150, 150, 150, 150,} \textbf{150}]. Each output has 25 neurons, totaling 150 neurons. We train the network for 1000 epochs with a batch size of 60, using the ADAM optimizer with a learning rate of 0.001, based solely on DD loss (Eq. \ref{Loss_data}). We use a default initializer and the DeepONet activation function described by DeepXDE \citep{lu2021deepxde}.

The prediction errors for each variable are as follows: $U_x$=29\%, $U_y$=46\%, $U_z$=47\%, $R_x$=41\%, $R_y$=23\%, and $R_z$=34\%. The training time is approximately 15 hours. These results indicate that this technique is unsuitable for complex structures with intricate relationships between applied loading and resulting displacements/rotations.

A major issue is that training a single network, which shares information across all outputs, leads to large errors. Additionally, there is a substantial difference in the scale of each output: $R_y$ is nearly ten times smaller than $R_x$, thirty times smaller than $R_z$, and about five hundred times smaller than $U_x$. During weight updates, even slight adjustments in one neuron of any layer cause significant changes in the values of all other output variables, complicating the network’s training. It was also specified by Lu et al. \citep{lu2022comprehensive}, that dealing with multiple outputs depends on the problem type, and the strategy may vary from case to case. A comprehensive study is needed to determine the best strategy for managing multiple outputs \citep{lu2022comprehensive}. Additionally, as the structure’s complexity increases, even a relatively small number of epochs (1000) results in considerable training time (15 hours). Based on these findings, we conclude that the split branch/trunk strategy is not suitable for the presented case.

\subsubsection{N Independent DeepONet for N Outputs}

In this section, we implement one main network containing six independent DeepONets for the six output functions ($U_x$, $U_y$, $U_z$, $R_x$, $R_y$, and $R_z$) as illustrated in Figure \ref{Multiple_Statergy}\textbf{A.}. Each DeepONet has its own branch and trunk network. The architectures of all DeepONets are identical, with branch networks structured as [\textbf{156}\textit{, 75, 75, 75, 75, 75,} \textbf{75}] and trunk networks as [\textbf{3}\textit{, 75, 75, 75, 75, 75,} \textbf{75}]. The key advantage of this technique over the split branch/trunk strategy is that the weight updates for one output function do not affect the weights of the other outputs, as each output is handled by a separate DeepONet. The total loss is calculated based on the outcomes of each DeepONet and is then propagated back to all DeepONets to update weights independently based on the target output function.

Considering these aspects of the training, we proceed with four different approaches for the combination of loss functions. 1st approach is purely DD, in which the loss function consists of the mean relative \( L_2 \) error (Eq. \ref{Loss_data}), with each mesh point considered for training, resulting in 1882 mesh points. 2nd approach is a combination of DD \& EC, in which the loss function consists of the mean relative \( L_2 \) error with the loss function derived by following the energy conservation principle (Eq. \ref{eq:DDEC}) in which each mesh point is considered for training, resulting in 1882 mesh points. 3rd approach is the same as 2nd approach, however, instead of using all mesh points, only the master nodes are considered for training, resulting in 998 mesh points. After predictions, post-processing based on the FEM model constraints are involved to obtain results for the slave nodes. Finally, the 4th approach is a combination of DD \& SE-S, in which the loss function consists of mean relative \( L_2 \) error with the loss function derived based on the static equilibrium principle using Schur complement (Eq. \ref{eq:SESchur}). In this approach, instead of selecting all master nodes, a few master nodes are selected from the structure (mostly on girders and arches) for training the network. Once the network is trained and predictions are made on these nodes, the outputs at the remaining master nodes are determined using Eq. \ref{Matrix_2}. After obtaining all master node outputs, the predicted outputs for all mesh points are obtained using the relationship between master nodes and slave nodes. 

Table \ref{Details_of_cases} provides an overview of the different approaches used, detailing the DeepONet structure, inputs, outputs, and postprocessing steps. We examine seven cases based on four approaches: Case 1 uses the DD loss function to predict five outputs, excluding $R_y$; Case 2 also uses the DD loss function but predicts all six outputs; Case 3 employs the DD \& EC loss function across the full domain, predicting five outputs; Case 4 applies the DD \& EC loss function only to master nodes and uses postprocessing to extend predictions to the entire domain, also for five outputs; and Cases 5, 6, and 7 use the DD \& SE-S loss function with 201 and 100 nodes for training, predicting five and six outputs, respectively. Predictions in these cases are initially valid only for the selected nodes, with postprocessing used to extend results to the remaining nodes and the full domain. Given that $R_y$ is very small and its effect is minimal (Section \ref{splitbranch}), it is excluded from most networks (Cases 1, 3, 4, and 5). However, Approach 4 (Cases 6 and 7) demonstrates promising results even for $R_y$. Therefore, Case 2 is included to compare predictions for all six variables using the SE-S loss approach with those obtained using the DD loss approach.

For training DeepONet, we select 20,000 epochs with a batch size of 60, utilizing the ADAM optimizer with a learning rate of 0.001. The increased number of epochs is required due to the 20\% training ratio, which necessitates more epochs to effectively learn from the smaller training dataset.

\begin{table}[]
    \centering
    \caption{Summarized approaches, structure, inputs, output, and loss functions}
    \label{Details_of_cases}
    \scriptsize
    \begin{tabular}{|c|c|c|c|c|c|c|c|}
    \hline
    \textbf{Approaches}                                                                                   & \textbf{Case} & \makecell{\textbf{No. of} \\ \textbf{DeepONet}} & \textbf{Structure}                                                                                                                          & \textbf{Input}                                                                                 & \textbf{Output}                                                               & \textbf{Loss}            & \textbf{Postprocess}                                                                      \\ \hline
    \multirow{2}{*}{\textbf{\begin{tabular}[c]{@{}c@{}}1st \\ (DD)\end{tabular}}}               & 1            & 5                 & \multirow{7}{*}{\begin{tabular}[c]{@{}c@{}}Branch: \\ {[}\textbf{156}\textit{,75,75,}\\ \textit{75,75,75,}\textbf{75}{]} \\ Trunk:\\ {[}\textbf{3}\textit{,75,75,75}\\ \textit{,75,75,}\textbf{75}{]}\end{tabular}} & \multirow{3}{*}{\begin{tabular}[c]{@{}c@{}}1882$\times$3 \\ (Mesh)\\ 9000$\times$156 \\ (Load)\end{tabular}} & \begin{tabular}[c]{@{}c@{}}9000$\times$1882$\times$5 \\ (Ux,Uy,Uz\\ ,Rx,Rz)\end{tabular}    & \multirow{2}{*}{Eq.\ref{Loss_data}}   & \multirow{2}{*}{NA}                                                                       \\ \cline{2-3} \cline{6-6}
                                                                                                & 2            & 6                 &                                                                                                                                             &                                                                                                & \begin{tabular}[c]{@{}c@{}}9000$\times$1882$\times$6 \\ (Ux,Uy,Uz,\\ Rx,Ry,Rz)\end{tabular} &                          &                                                                                           \\ \cline{1-3} \cline{6-8} 
    \textbf{\begin{tabular}[c]{@{}c@{}}2nd \\ (DD\\ \& EC)\end{tabular}}                         & 3            & 5                 &                                                                                                                                             &                                                                                                & \begin{tabular}[c]{@{}c@{}}9000$\times$1882$\times$5 \\ (Ux,Uy,Uz,\\ Rx,Rz)\end{tabular}    & Eq.\ref{eq:DDEC}                   & NA                                                                                        \\ \cline{1-3} \cline{5-8} 
    \textbf{\begin{tabular}[c]{@{}c@{}}3rd\\ (DD \\ \& Mast. EC)\end{tabular}}                   & 4            & 5                 &                                                                                                                                             & \begin{tabular}[c]{@{}c@{}}998$\times$3 \\ (Mesh) \\ 9000$\times$156 \\ (Load)\end{tabular}                  & \begin{tabular}[c]{@{}c@{}}9000$\times$998$\times$5 \\ (Ux,Uy,Uz,\\ Rx,Rz)\end{tabular}     & Eq.\ref{eq:DDEC}                  & \begin{tabular}[c]{@{}c@{}}Master to \\ Slave\end{tabular}                                \\ \cline{1-3} \cline{5-8} 
    \multirow{3}{*}{\textbf{\begin{tabular}[c]{@{}c@{}}4th \\ (DD\\  \& SE-S)\end{tabular}}} & 5            & 5                 &                                                                                                                                             & \begin{tabular}[c]{@{}c@{}}201$\times$3 \\ (Mesh) \\ 9000$\times$156 \\ (Load)\end{tabular}                  & \begin{tabular}[c]{@{}c@{}}9000$\times$201$\times$5 \\ (Ux,Uy,Uz,\\ Rx,Rz)\end{tabular}     & \multirow{3}{*}{Eq.\ref{eq:SESchur}} & \multirow{3}{*}{\begin{tabular}[c]{@{}c@{}}Eq. \ref{Matrix_2} and \\ Master to \\ Slave\end{tabular}} \\ \cline{2-3} \cline{5-6}
                                                                                                & 6            & 6                 &                                                                                                                                             & \begin{tabular}[c]{@{}c@{}}201$\times$3 \\ (Mesh) \\ 9000$\times$156 \\ (Load)\end{tabular}                  & \begin{tabular}[c]{@{}c@{}}9000$\times$201$\times$6 \\ (Ux,Uy,Uz,\\ Rx,Ry,Rz)\end{tabular}  &                          &                                                                                           \\ \cline{2-3} \cline{5-6}
                                                                                                & 7            & 6                 &                                                                                                                                             & \begin{tabular}[c]{@{}c@{}}100$\times$3 \\ (Mesh) \\ 9000$\times$156 \\ (Load)\end{tabular}                  & \begin{tabular}[c]{@{}c@{}}9000$\times$100$\times$6 \\ (Ux,Uy,Uz,\\ Rx,Ry,Rz)\end{tabular}  &                          &                                                                                           \\ \hline
    \end{tabular}
\end{table}

\subsubsection{Results}

\begin{figure}[h]
    \centering
    \includegraphics[width=1\textwidth]{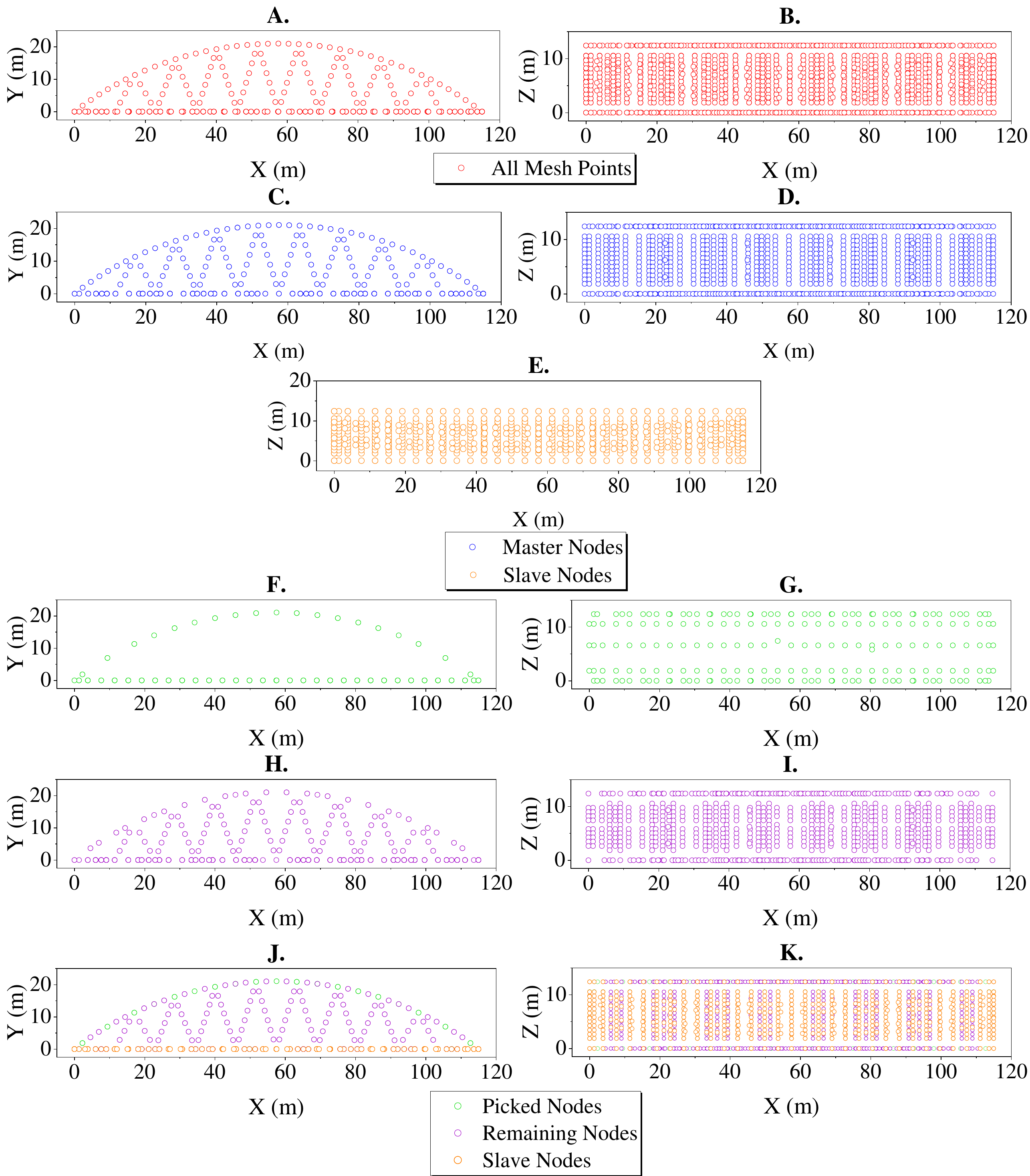}
    \caption{Domain points for all approaches: \textbf{A. \& B.} Side and top view of training nodes for 1st and 2nd approach  \textbf{C. \& D.} Side and top view of training master nodes for 3rd approach \textbf{E.:} Top view of slave nodes (The solution obtained from the post-processing results from \textbf{C. \& D.} nodes) \textbf{F. \& G.} Side and top view of picked master nodes using Schur complement for 4th approach \textbf{H. \& I.} Side and top view of remaining master nodes (The solution obtained from the post-processing results from \textbf{F. \& G.} nodes using Equation \ref{Matrix_2}) and \textbf{J. \& K.} All nodes of FEM model, \textcolor{green}{Green}: Picked nodes for the network training using Schur complement, \textcolor{purple}{Purple}: Remaining master nodes of the domain (The solution obtained from the post-processing of the results on \textcolor{green}{green} nodes using Equation \ref{Matrix_2}), \textcolor{orange}{Orange}: Slave nodes of the domain (The solution obtained from the post-processing all results of master nodes (\textcolor{green}{Green} \& \textcolor{purple}{Purple}) to obtain at slave nodes)}
    \label{All_Approaches}
\end{figure}

\begin{figure}[h]
    \centering
    \includegraphics[width=0.64\textwidth]{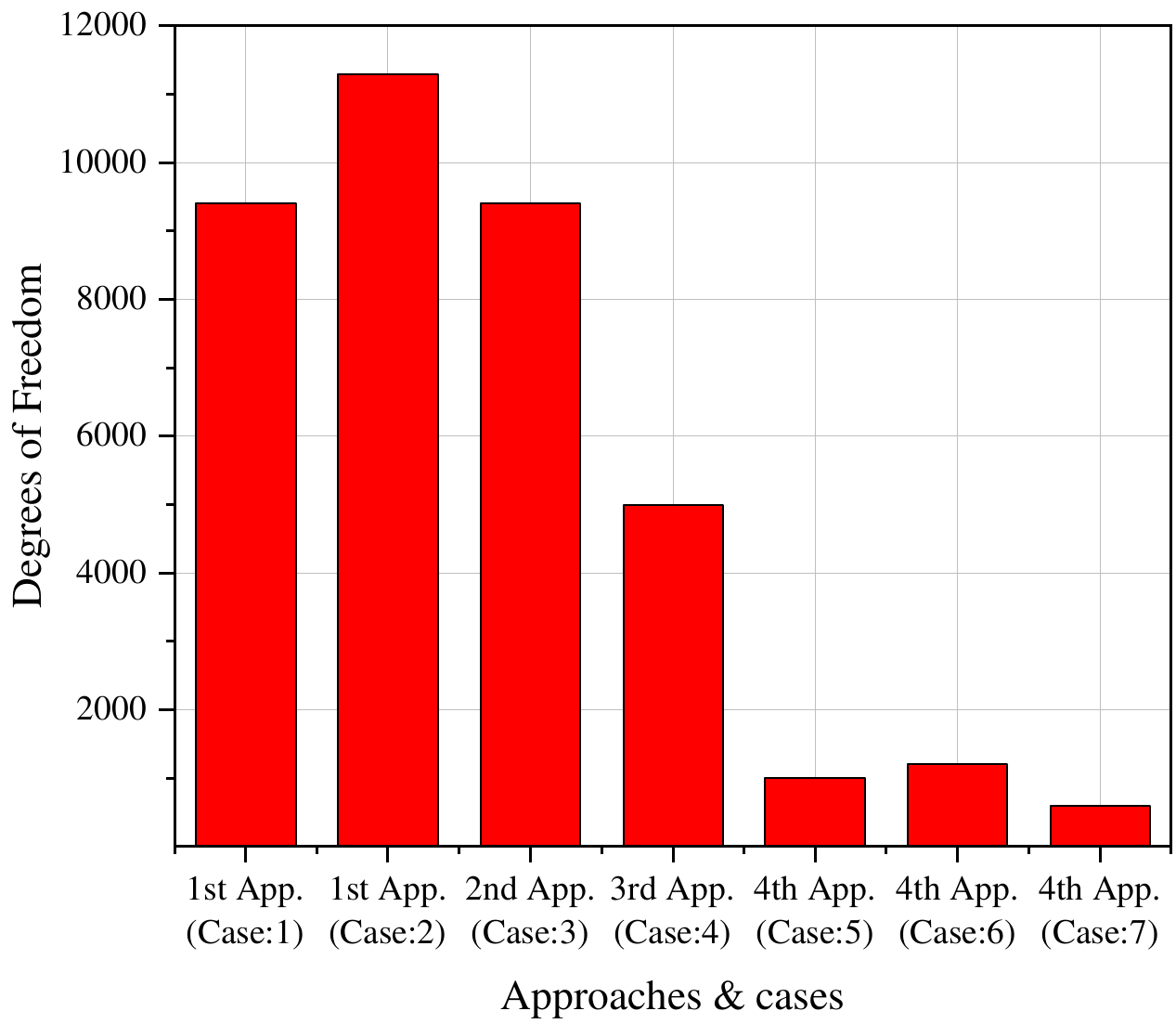}
    \caption{The DeepONet training degrees of freedom associated with each approach and case}
    \label{DOFs}
\end{figure}

Figure \ref{All_Approaches} displays scatter plots of mesh points along the $x$, $y$, and $z$ axes for all considered approaches. For the 1st and 2nd approaches, the entire set of 1,882 mesh points is used for training (Figures \ref{All_Approaches}\textbf{A. \& B.}). In contrast, the 3rd approach (Figures \ref{All_Approaches}\textbf{C. \& D.}) trains only on master nodes and uses post-processing (Master to Slave node constraints) to predict the remaining nodes (Figure \ref{All_Approaches}\textbf{E.}). The 4th approach includes two cases. In the first case, 201 master nodes are selected for training (Figures \ref{All_Approaches}\textbf{F. \& G.}). Predictions for the remaining master nodes (Figures \ref{All_Approaches}\textbf{H. \& I.}) are made using Equation \ref{Matrix_2}, followed by applying Master to Slave node constraints to determine the slave nodes (Figures \ref{All_Approaches}\textbf{J. \& K.}). In the second case, 100 master nodes are chosen for training, with predictions made for the remaining master nodes using Eq. \ref{Matrix_2}, and then Master to Slave node constraints are applied to the slave nodes. This significant reduction from 1,882 to 100 nodes in the training dataset markedly improves the network's accuracy, performance, and computational efficiency. Figure \ref{DOFs} illustrates the training domain DOFs for each approach and case outlined in Table \ref{Details_of_cases}. It is evident from Figure \ref{DOFs} that the SE-S approach (4th approach) substantially reduces the number of DOFs to be learned by the network—from 11,292 DOFs in Case 2 to 600 DOFs in Case 7—resulting in more accurate and efficient learning.

\begin{figure}[h]
    \centering
    \includegraphics[width=1\textwidth]{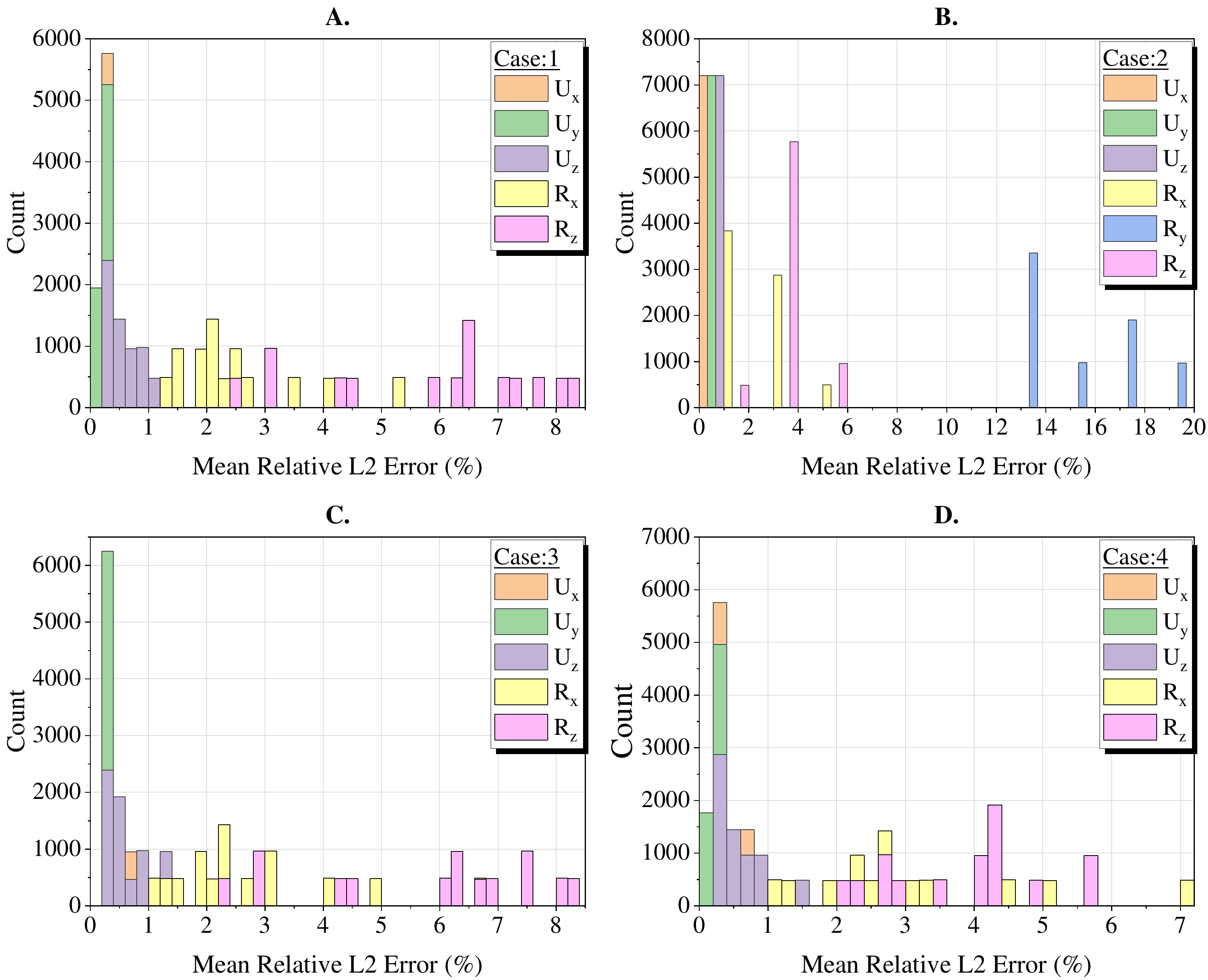}
    \caption{Error Histogram for cases 1 to 4 \textbf{A.} DD (Eq. \ref{Loss_data}) with 5 output variables (1st Approach) \textbf{B.} DD (Eq. \ref{Loss_data}) with 6 output variables (1st Approach) \textbf{C.} DD \& EC (Eq. \ref{eq:DDEC}) with 5 output variables (2nd Approach), \textbf{D.} DD \& Master nodes EC (Eq. \ref{eq:DDEC}) with 5 output variables (3rd Approach)}
    \label{Histogram_KW51_1}
\end{figure}

\begin{figure}[h]
    \centering
    \includegraphics[width=1\textwidth]{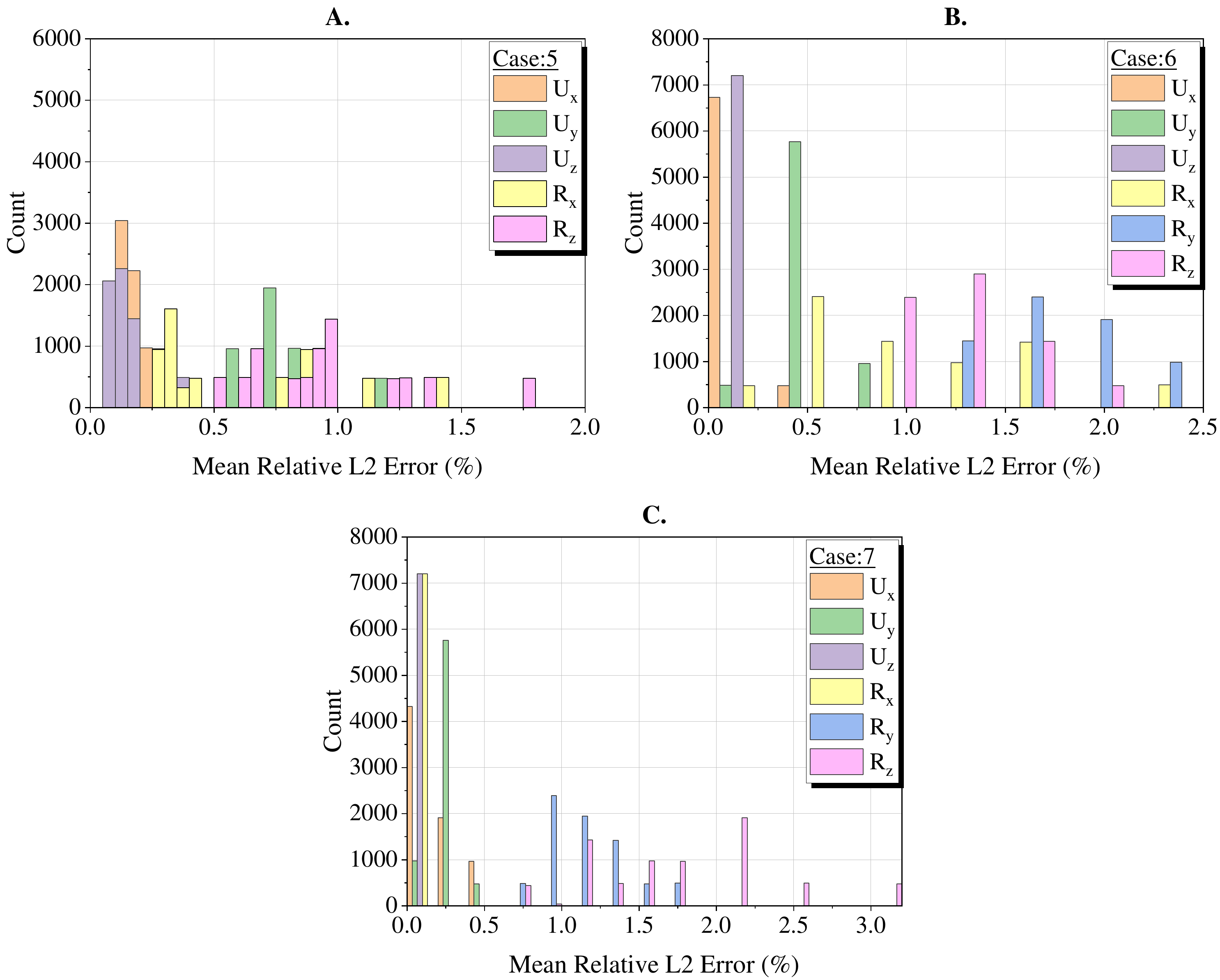}
    \caption{Error Histogram for cases 5 to 7 \textbf{A.} DD \& SE-S (Eq. \ref{eq:SESchur}) with 5 output variables for 201 nodes (4th Approach) \textbf{B.} DD \& SE-S (Eq. \ref{eq:SESchur}) with 6 output variables for 201 nodes (4th Approach) \textbf{C.} DD \& SE-S (Eq. \ref{eq:SESchur}) with 6 output variables for 100 nodes (4th Approach)}
    \label{Histogram_KW51_2}
\end{figure}

\begin{figure}[h]
    \centering
    \includegraphics[width=1.0\textwidth]{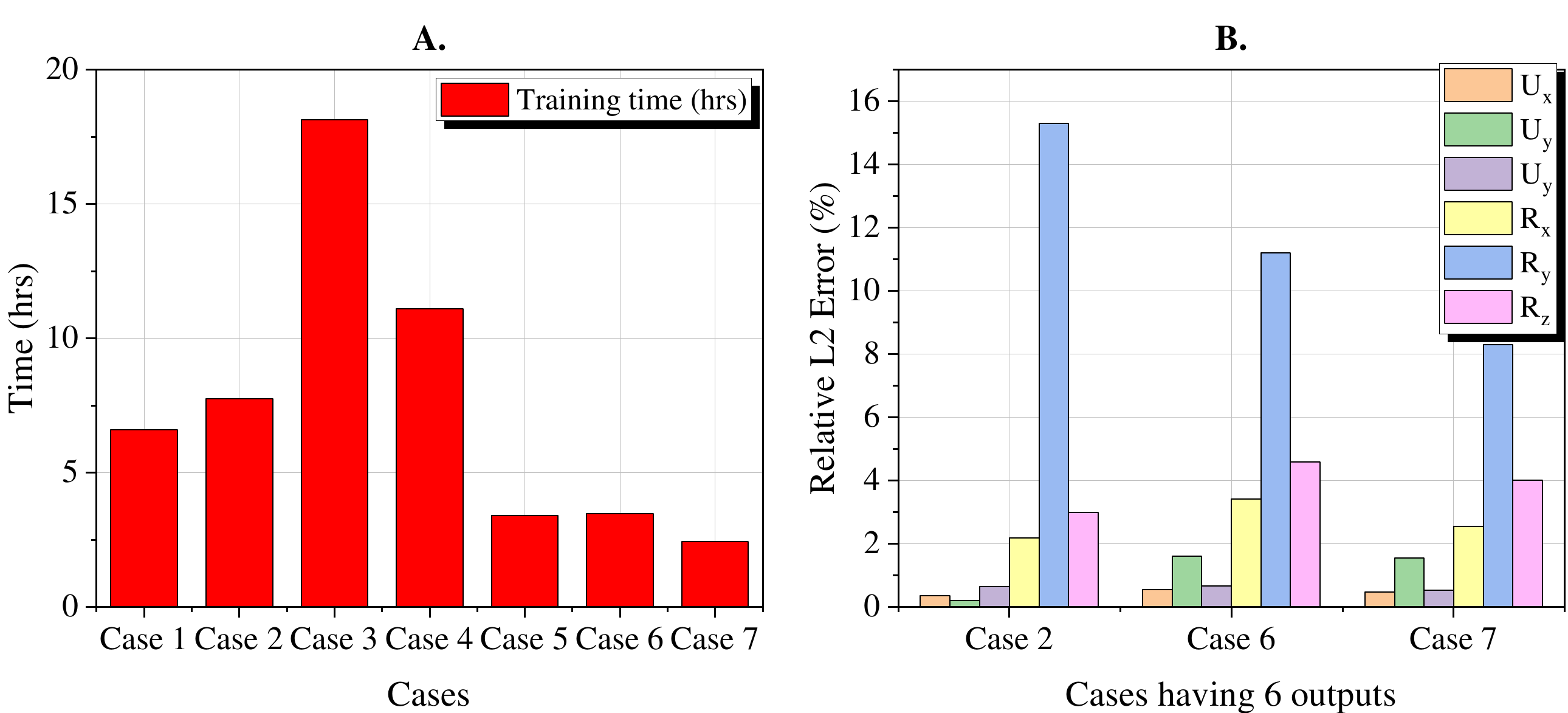}
    \caption{Training time and Post-processed error \textbf{A.} Training time required for all cases, \textbf{B.} Error comparison of Case 2 with post-processed error for Case 6 and Case 7}
    \label{Error_KW51}
\end{figure}

\begin{figure}[h]
    \centering
    \begin{subfigure}[b]{0.46\textwidth}
        \centering
        \includegraphics[width=\textwidth]{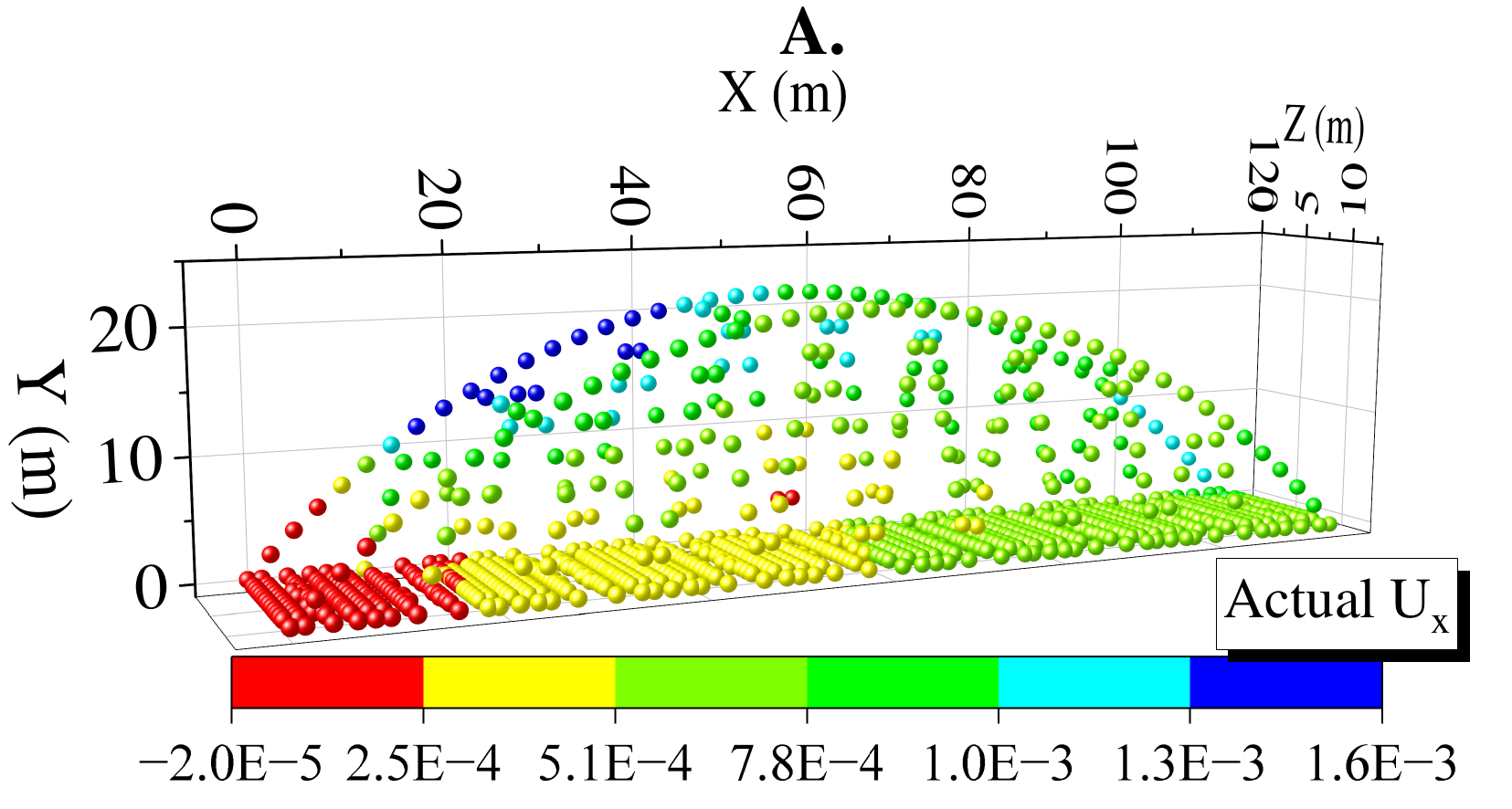}
    \end{subfigure}
    \hfill
    \begin{subfigure}[b]{0.46\textwidth}
        \centering
        \includegraphics[width=\textwidth]{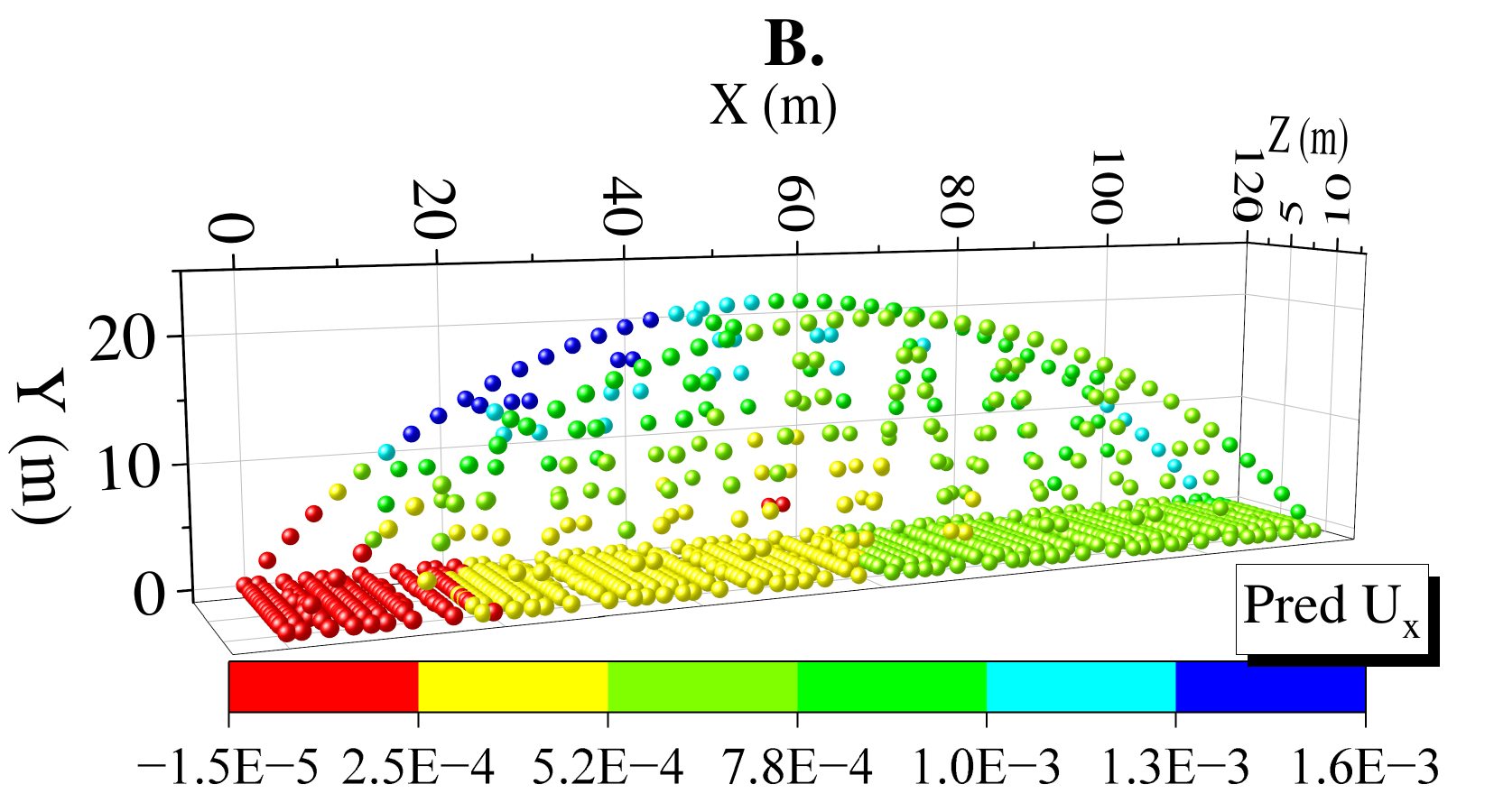}
    \end{subfigure}

    \begin{subfigure}[b]{0.46\textwidth}
        \raggedright
        \includegraphics[width=\textwidth]{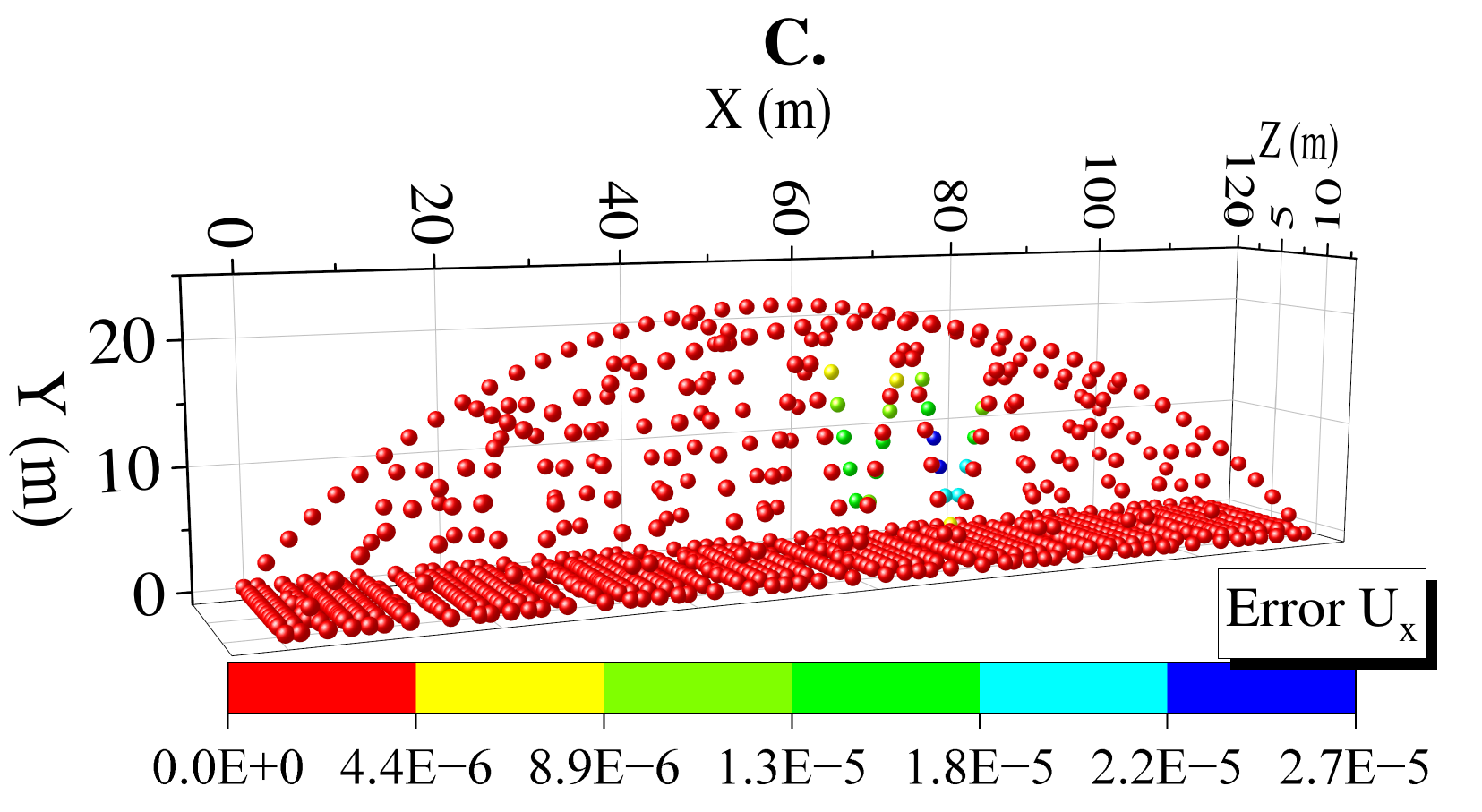}
    \end{subfigure}

    \begin{subfigure}[b]{0.46\textwidth}
        \centering
        \includegraphics[width=\textwidth]{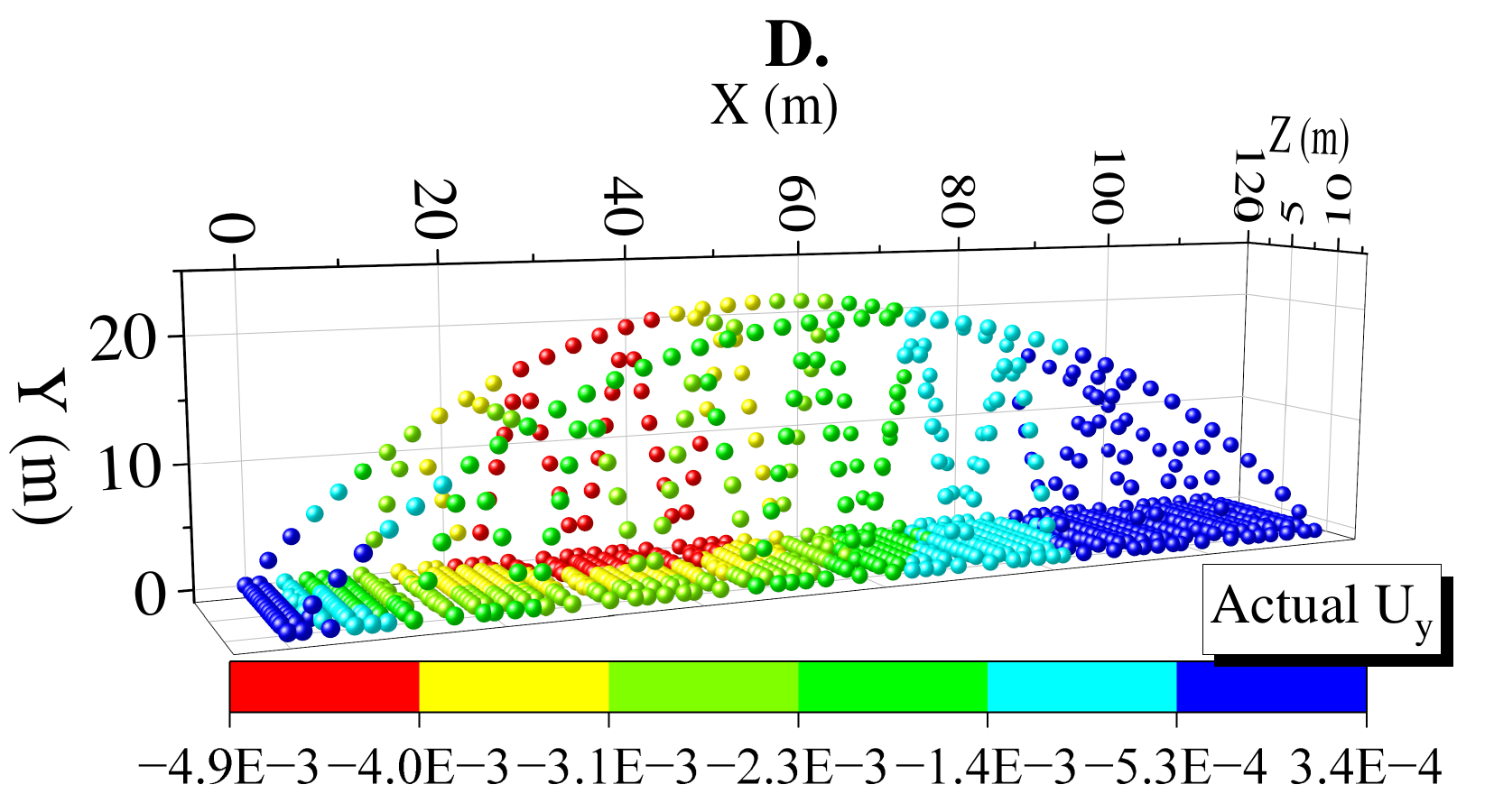}
    \end{subfigure}
    \hfill
    \begin{subfigure}[b]{0.46\textwidth}
        \centering
        \includegraphics[width=\textwidth]{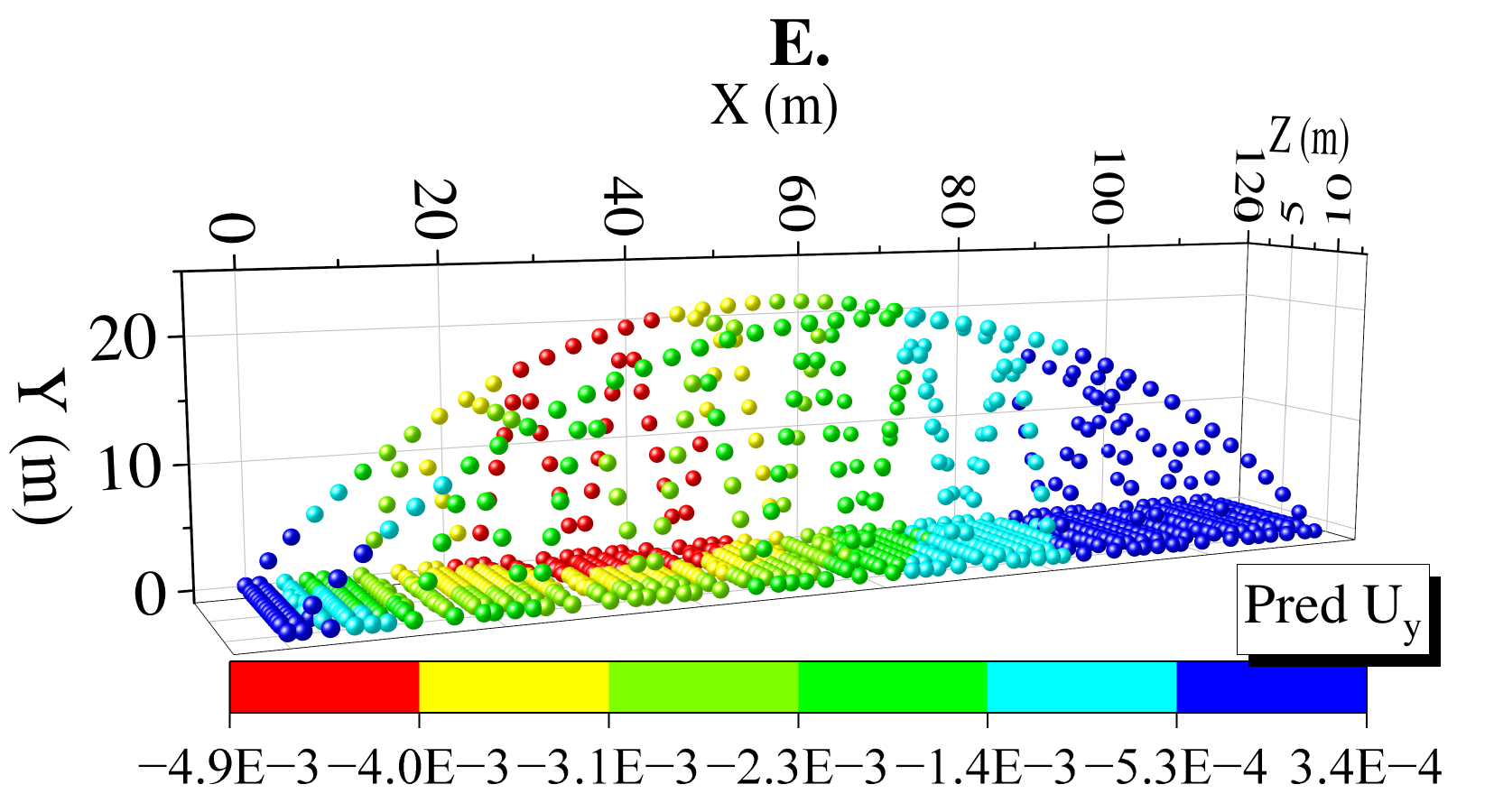}
    \end{subfigure}

    \begin{subfigure}[b]{0.46\textwidth}
        \raggedright
        \includegraphics[width=\textwidth]{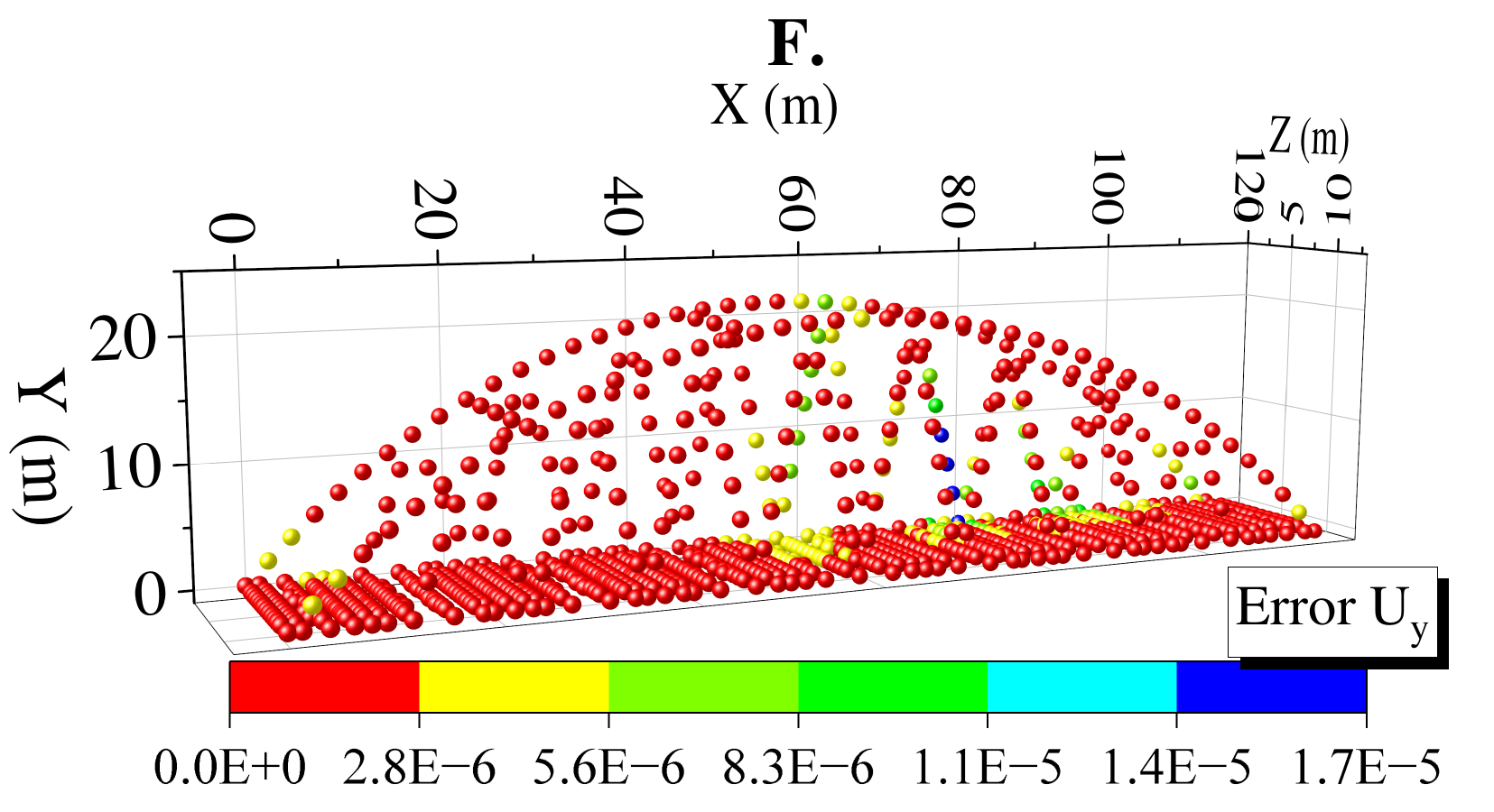}
    \end{subfigure}

    \begin{subfigure}[b]{0.46\textwidth}
        \centering
        \includegraphics[width=\textwidth]{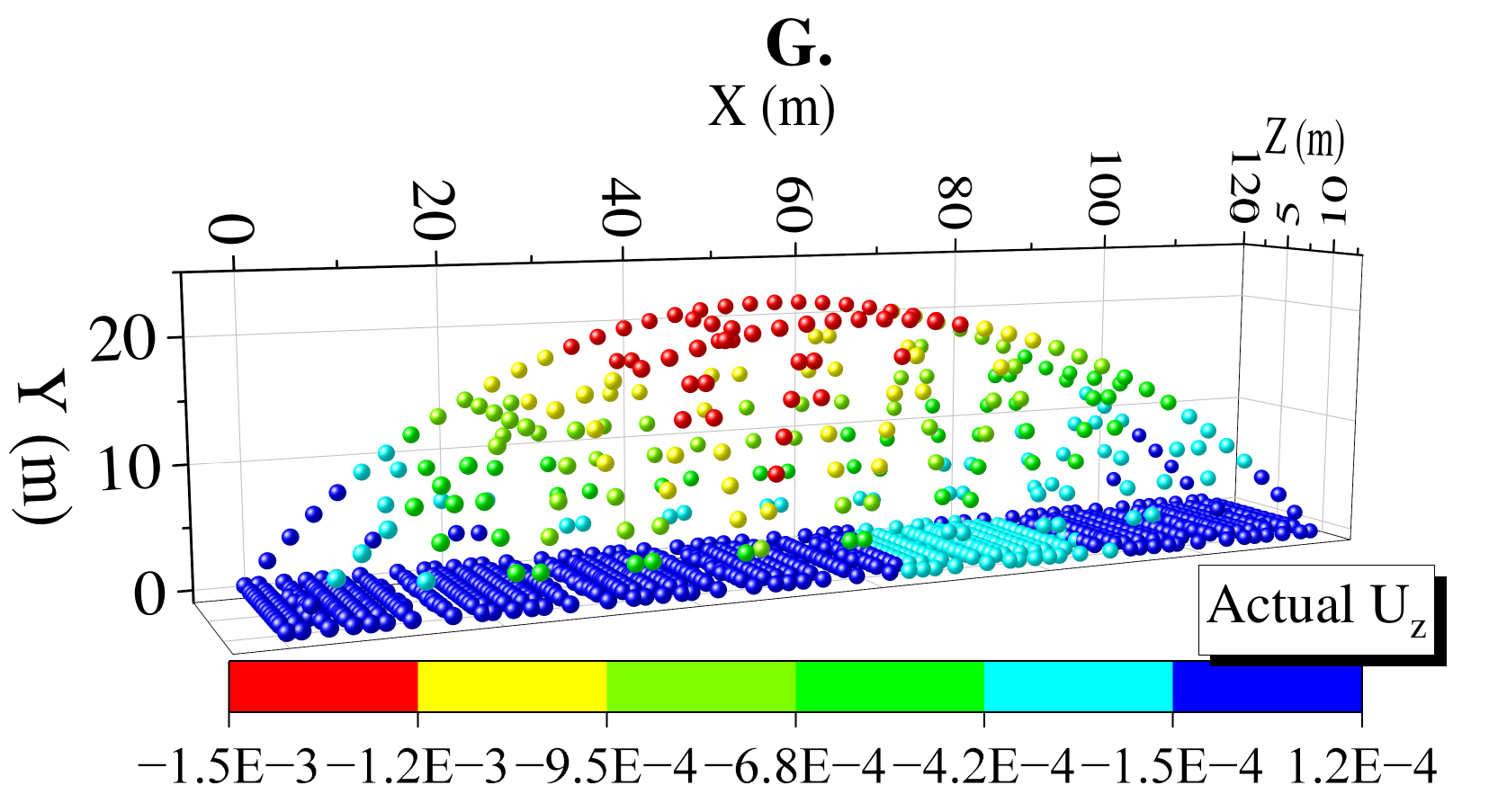}
    \end{subfigure}
    \hfill
    \begin{subfigure}[b]{0.46\textwidth}
        \centering
        \includegraphics[width=\textwidth]{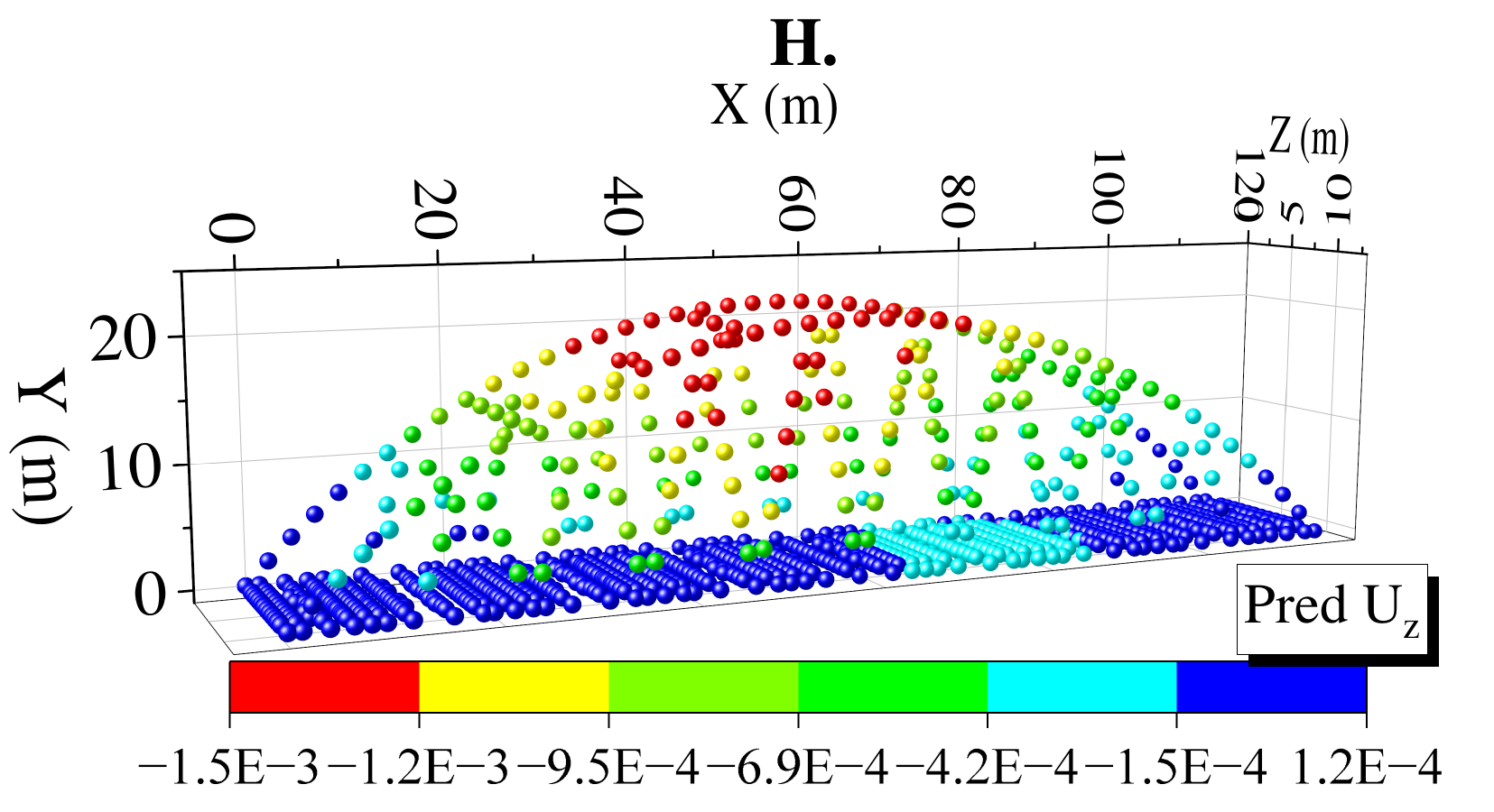}
    \end{subfigure}

    \begin{subfigure}[b]{0.46\textwidth}
        \raggedright
        \includegraphics[width=\textwidth]{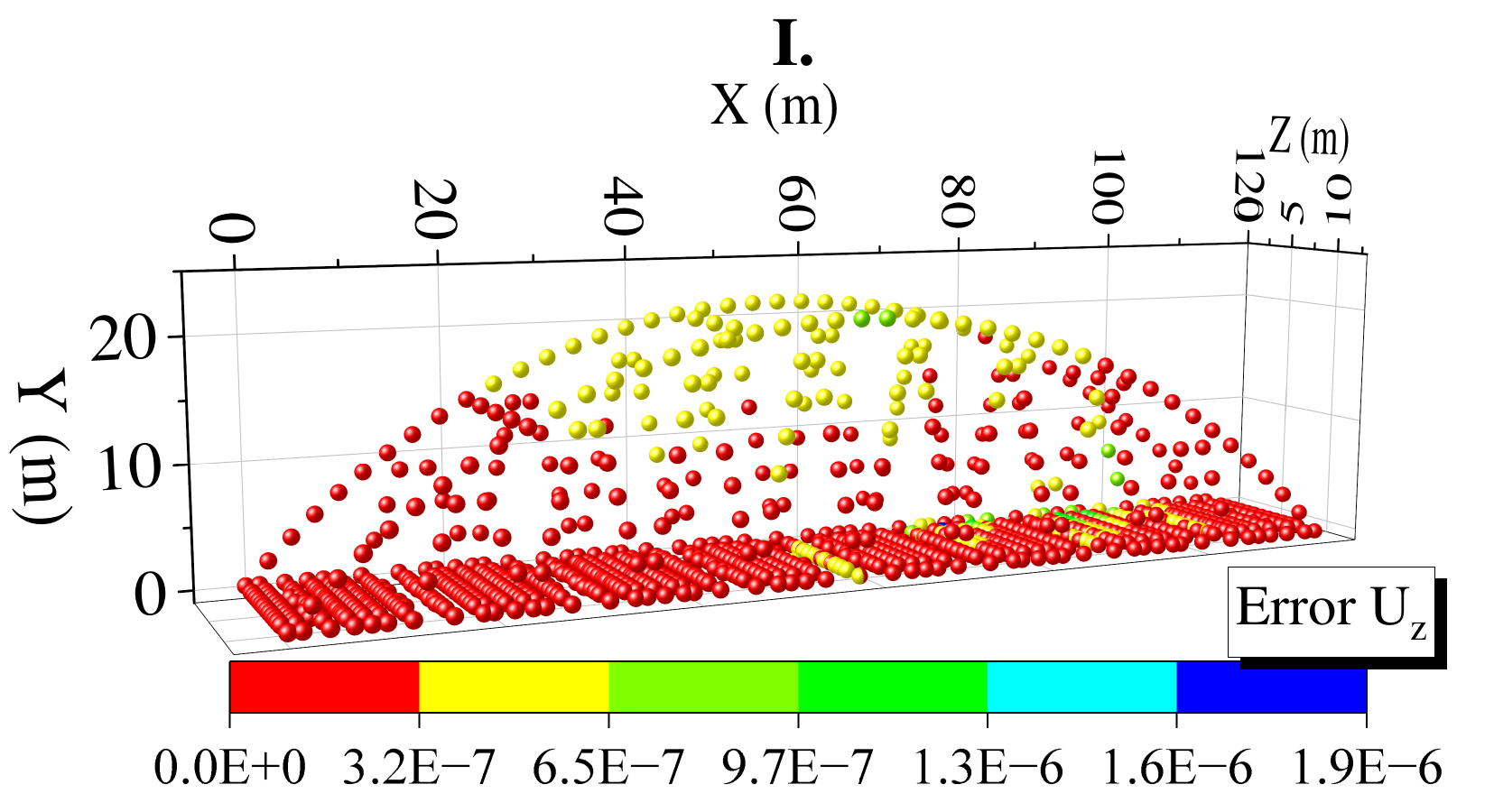}
    \end{subfigure}

    \caption{For random sample \textbf{A.} Actual $U_x$, \textbf{B.} Predicted $U_x$, \textbf{C.} Error for $U_x$, \textbf{D.} Actual $U_y$, \textbf{E.} Predicted $U_y$, \textbf{F.} Error for $U_y$, \textbf{G.} Actual $U_z$, \textbf{H.} Predicted $U_z$, \textbf{I.} Error for $U_z$}
    \label{Disp_KW51}
\end{figure}

\begin{figure}[h]
    \centering
    \begin{subfigure}[b]{0.46\textwidth}
        \centering
        \includegraphics[width=\textwidth]{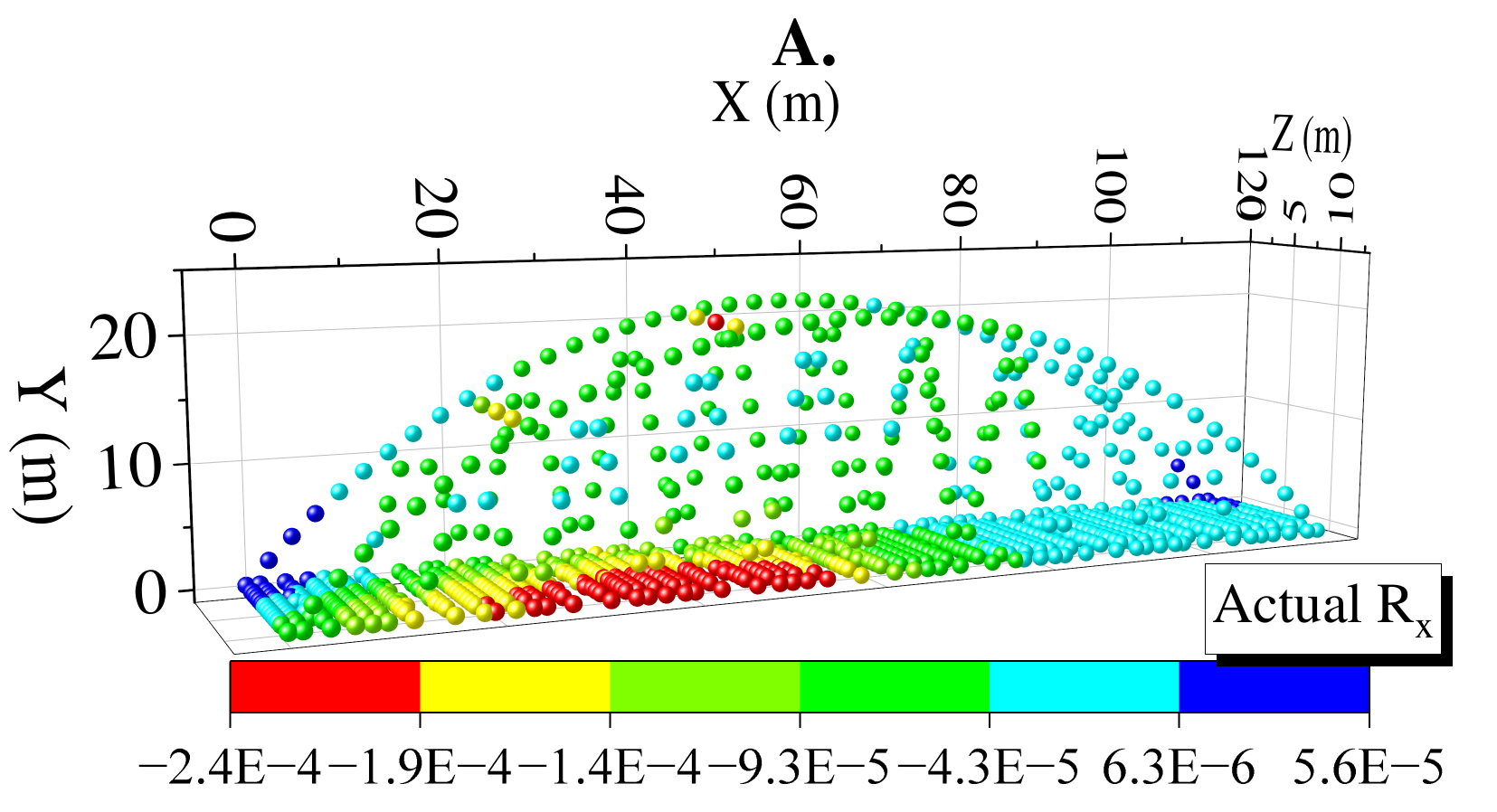}
    \end{subfigure}
    \hfill
    \begin{subfigure}[b]{0.46\textwidth}
        \centering
        \includegraphics[width=\textwidth]{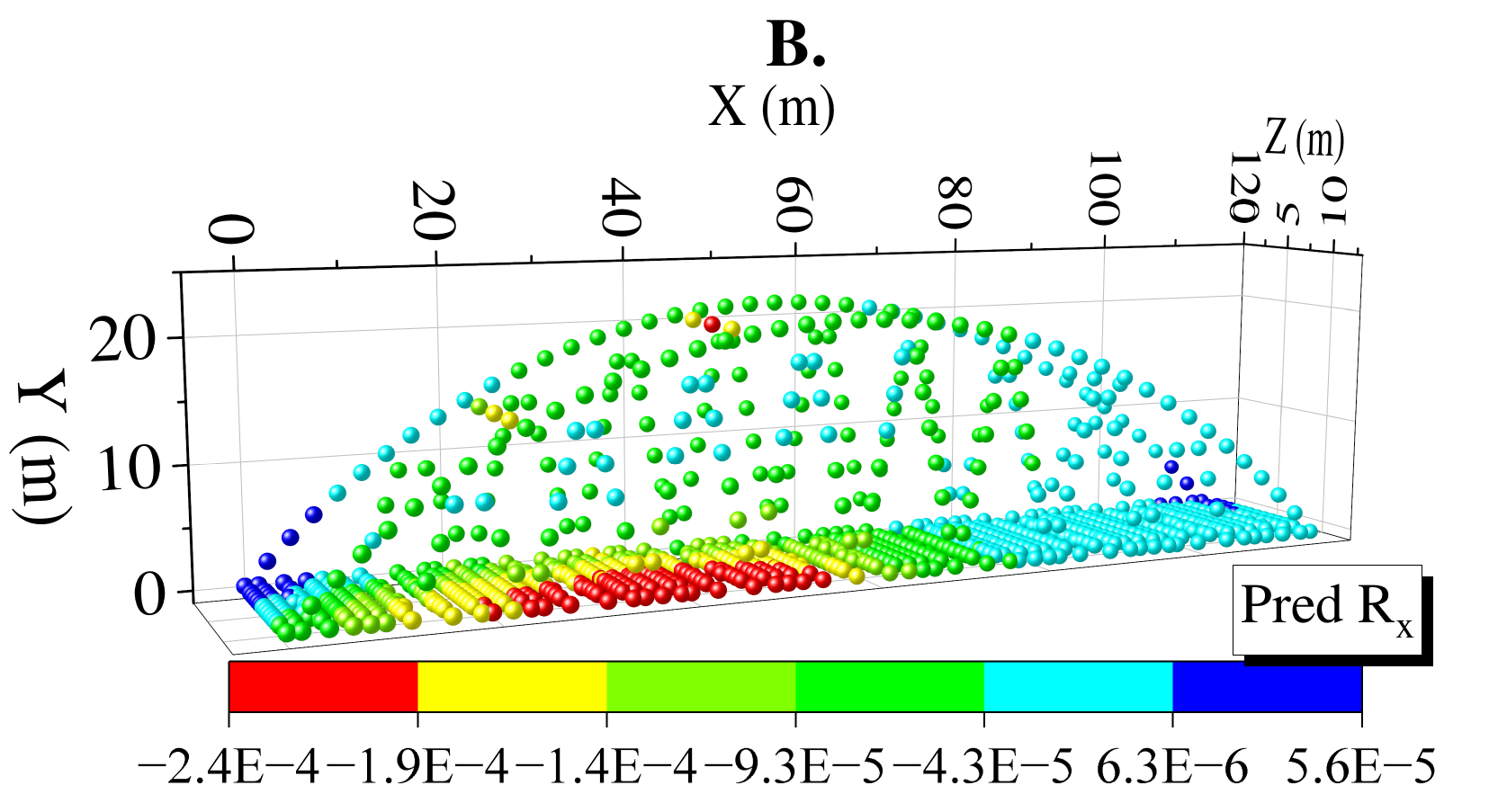}
    \end{subfigure}

    \begin{subfigure}[b]{0.46\textwidth}
        \raggedright
        \includegraphics[width=\textwidth]{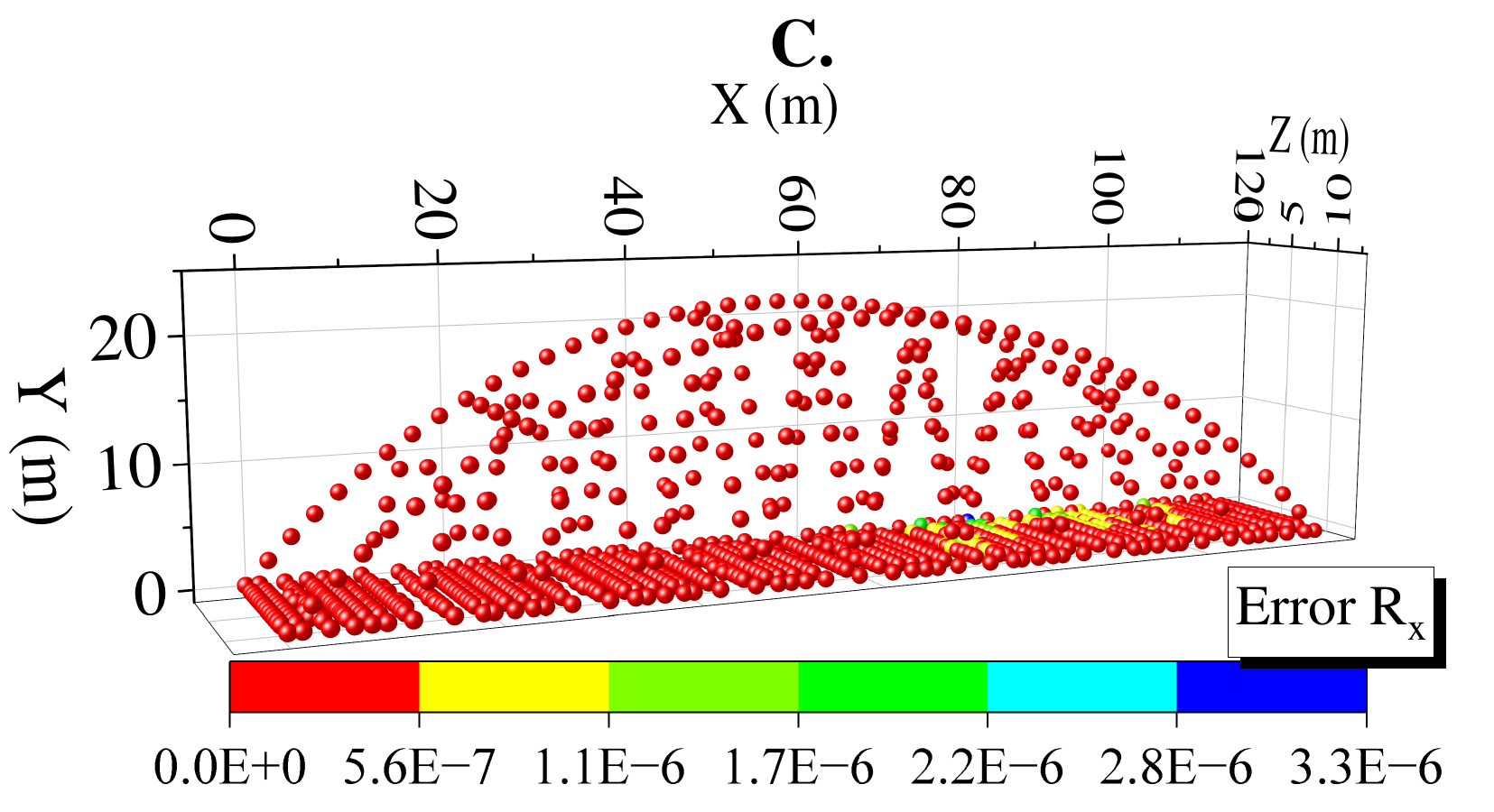}
    \end{subfigure}

    \begin{subfigure}[b]{0.46\textwidth}
        \centering
        \includegraphics[width=\textwidth]{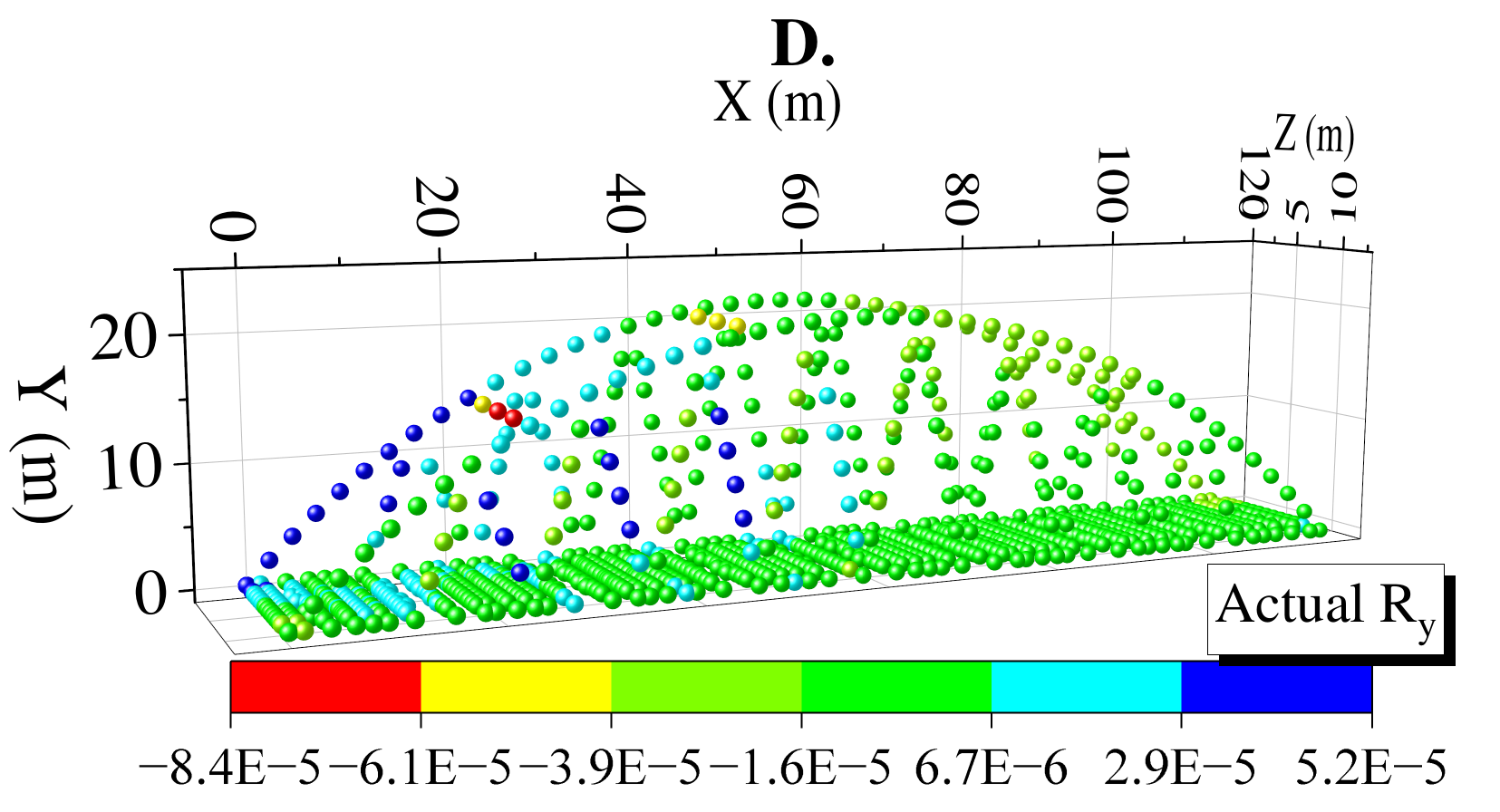}
    \end{subfigure}
    \hfill
    \begin{subfigure}[b]{0.46\textwidth}
        \centering
        \includegraphics[width=\textwidth]{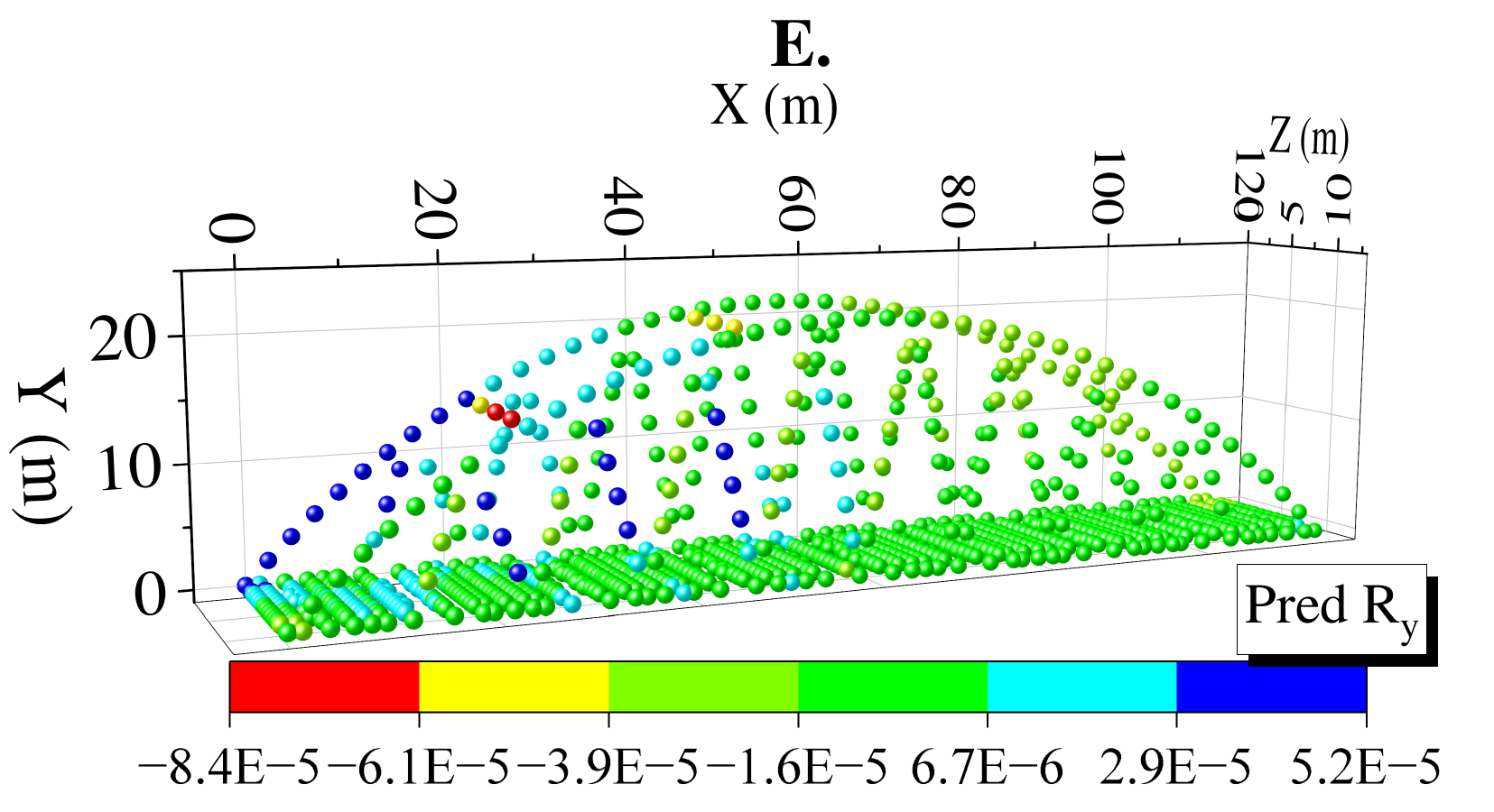}
    \end{subfigure}

    \begin{subfigure}[b]{0.46\textwidth}
        \raggedright
        \includegraphics[width=\textwidth]{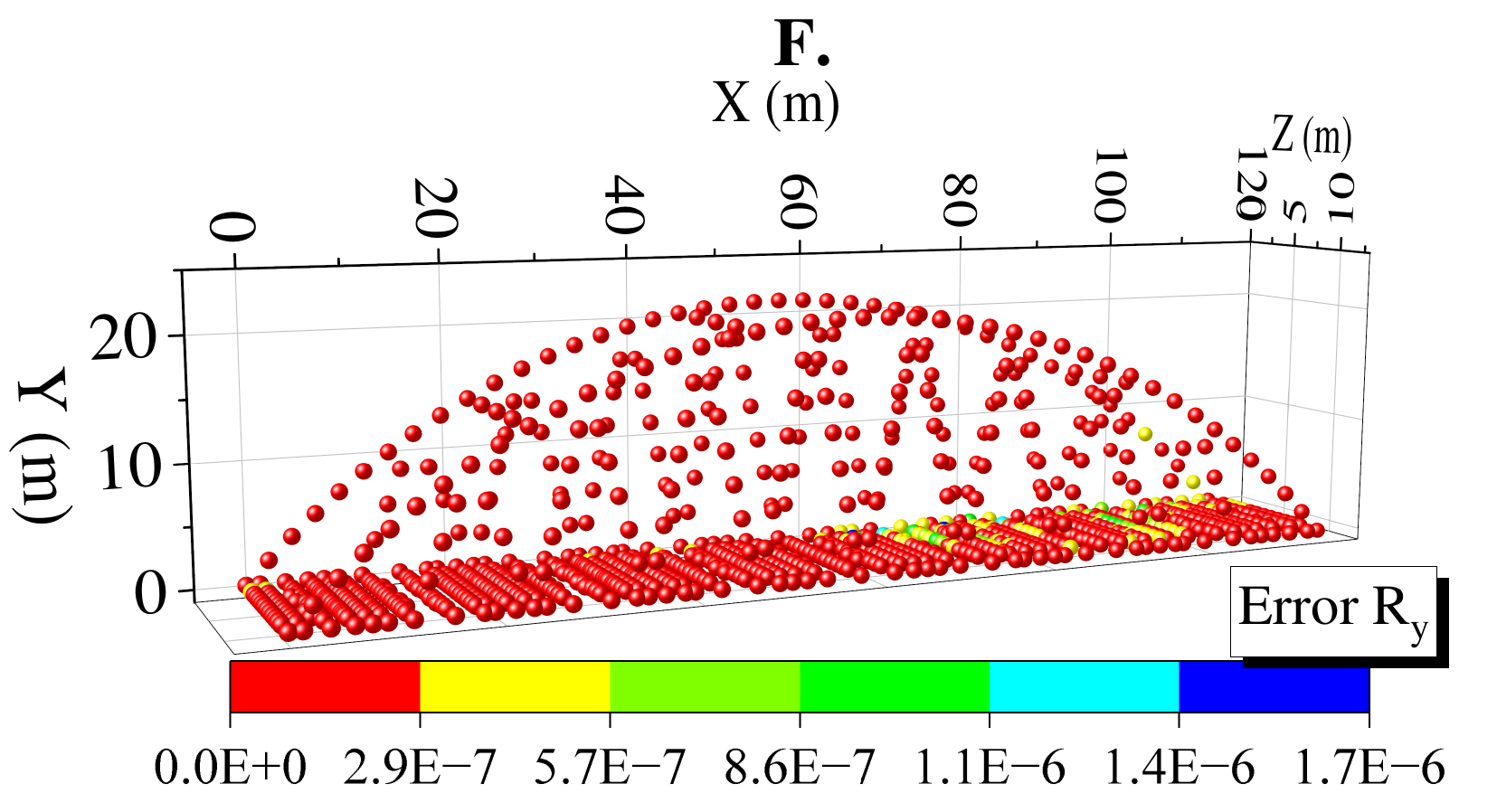}
    \end{subfigure}

    \begin{subfigure}[b]{0.46\textwidth}
        \centering
        \includegraphics[width=\textwidth]{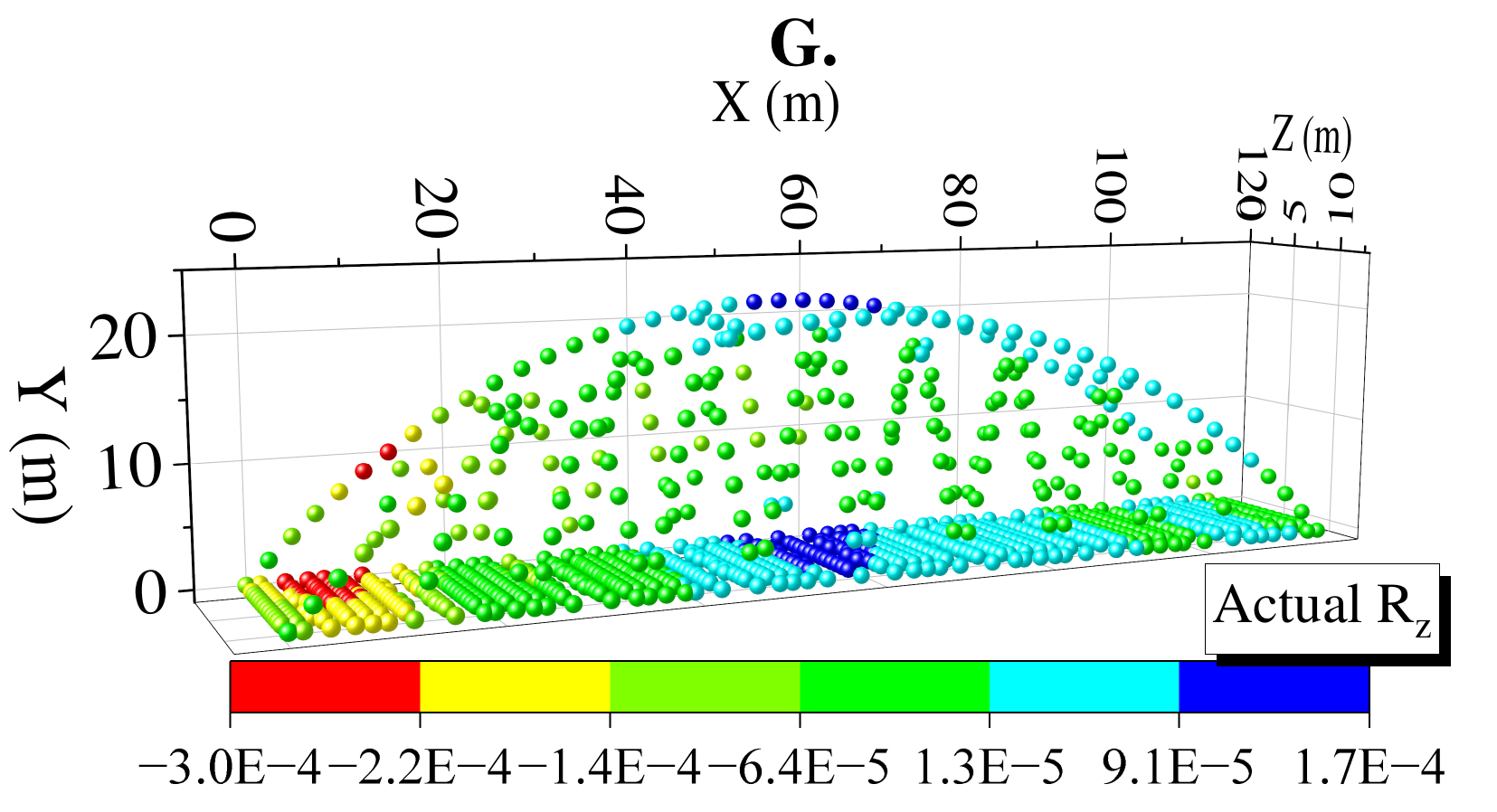}
    \end{subfigure}
    \hfill
    \begin{subfigure}[b]{0.46\textwidth}
        \centering
        \includegraphics[width=\textwidth]{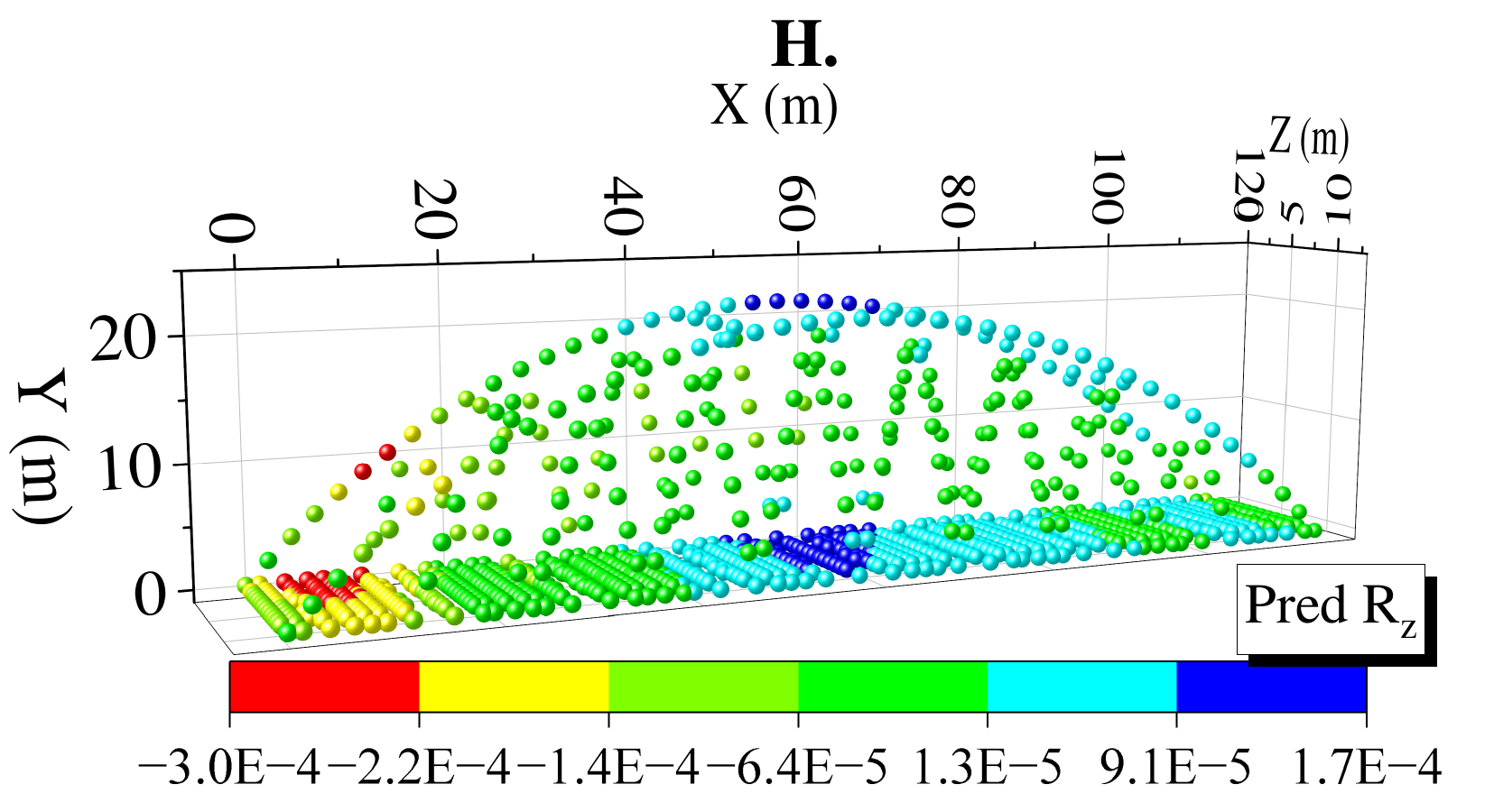}
    \end{subfigure}

    \begin{subfigure}[b]{0.46\textwidth}
        \raggedright
        \includegraphics[width=\textwidth]{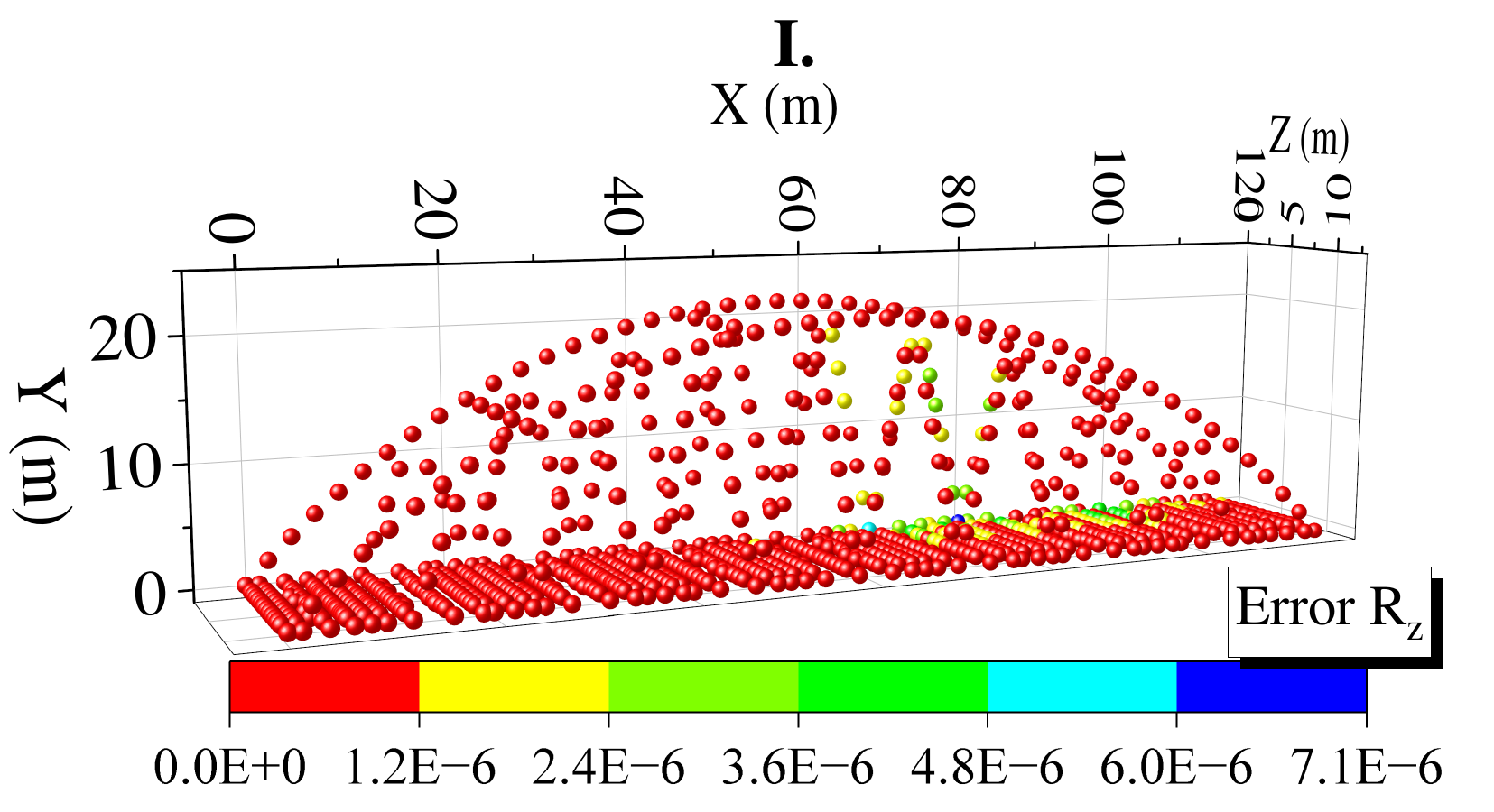}
    \end{subfigure}

    \caption{For random sample \textbf{A.} Actual $R_x$, \textbf{B.} Predicted $R_x$, \textbf{C.} Error for $R_x$, \textbf{D.} Actual $R_y$, \textbf{E.} Predicted $R_y$, \textbf{F.} Error for $R_y$, \textbf{G.} Actual $R_z$, \textbf{H.} Predicted $R_z$, \textbf{I.} Error for $R_z$}
    \label{Rot_KW51}
\end{figure}

Table \ref{Summarized_results} presents the mean, minimum, and maximum errors for each output variable across all cases. Figures \ref{Histogram_KW51_1} and \ref{Histogram_KW51_2} show the error histograms for all cases obtained from DeepONet. Figure \ref{Error_KW51} compares the training time and final mean error for each case, with post-processing results compared to the true values. These results demonstrate that while all approaches provide acceptable outcomes, the 100-node DD \& SE-S approach (Case 7) performs the best by significantly reducing training time compared to the other methods. From Figure \ref{Error_KW51}, it is clear that Case 2 (purely DD) results in an error for $R_y$ of up to 15\%, whereas Case 7 (using DD \& SE-S with post-processing) reduces the $R_y$ error to within 8\%. This notable improvement highlights the effectiveness of the SE-S approach in achieving precise predictions. Additionally, Case 7 has the lowest training time among all cases, further highlighting the efficiency of the SE-S loss function in integrating physics into the training process while minimizing training time. Figures \ref{Disp_KW51} and \ref{Rot_KW51} illustrate the predicted values ($U_x$, $U_y$, $U_z$, $R_x$, $R_y$, $R_z$) for a random sample, demonstrating minimal prediction error. Overall, the DD \& SE-S loss approach offers the best performance in terms of accuracy and training time.

\begin{table}[htbp]
    \centering
    \caption{Summarized mean, minimum, and maximum prediction error for all cases}
    \label{Summarized_results}
    \scriptsize 
    \begin{tabular}{|c|c|cccccc|}
        \hline
        \multirow{2}{*}{\textbf{Approaches}} & \multirow{2}{*}{\textbf{Case}} & \multicolumn{6}{c|}{\textbf{Error (\%)}} \\ \cline{3-8}
                                             &                                 & \multicolumn{1}{c|}{\textbf{Ux}} & \multicolumn{1}{c|}{\textbf{Uy}} & \multicolumn{1}{c|}{\textbf{Uz}} & \multicolumn{1}{c|}{\textbf{Rx}} & \multicolumn{1}{c|}{\textbf{Ry}} & \textbf{Rz} \\ \hline
        \multirow{2}{*}{\textbf{\begin{tabular}[c]{@{}c@{}}1st \\ (DD)\end{tabular}}} & 1 & \multicolumn{1}{c|}{\begin{tabular}[c]{@{}c@{}}0.36 \\ (0.25$\sim$0.64)\end{tabular}} & \multicolumn{1}{c|}{\begin{tabular}[c]{@{}c@{}}0.24 \\ (0.15$\sim$0.38)\end{tabular}} & \multicolumn{1}{c|}{\begin{tabular}[c]{@{}c@{}}0.71 \\ (0.26$\sim$1.90)\end{tabular}} & \multicolumn{1}{c|}{\begin{tabular}[c]{@{}c@{}}2.46 \\ (1.27$\sim$5.28)\end{tabular}} & \multicolumn{1}{c|}{NA} & \begin{tabular}[c]{@{}c@{}}5.84 \\ (2.40$\sim$8.36)\end{tabular} \\ \cline{2-8}
                                             & 2 & \multicolumn{1}{c|}{\begin{tabular}[c]{@{}c@{}}0.36 \\ (0.25$\sim$0.64)\end{tabular}} & \multicolumn{1}{c|}{\begin{tabular}[c]{@{}c@{}}0.22 \\ (0.16$\sim$0.32)\end{tabular}} & \multicolumn{1}{c|}{\begin{tabular}[c]{@{}c@{}}0.64 \\ (0.21$\sim$1.72)\end{tabular}} & \multicolumn{1}{c|}{\begin{tabular}[c]{@{}c@{}}2.20 \\ (0.68$\sim$5.30)\end{tabular}} & \multicolumn{1}{c|}{\begin{tabular}[c]{@{}c@{}}15.3 \\ (13.3$\sim$19.3)\end{tabular}} & \begin{tabular}[c]{@{}c@{}}3.00 \\ (1.73$\sim$4.23)\end{tabular} \\ \hline
        \textbf{\begin{tabular}[c]{@{}c@{}}2nd\\  (DD \\ \& EC)\end{tabular}} & 3 & \multicolumn{1}{c|}{\begin{tabular}[c]{@{}c@{}}0.37 \\ (0.26$\sim$0.66)\end{tabular}} & \multicolumn{1}{c|}{\begin{tabular}[c]{@{}c@{}}0.28 \\ (0.21$\sim$0.41)\end{tabular}} & \multicolumn{1}{c|}{\begin{tabular}[c]{@{}c@{}}0.70 \\ (0.23$\sim$1.86)\end{tabular}} & \multicolumn{1}{c|}{\begin{tabular}[c]{@{}c@{}}2.77 \\ (1.04$\sim$6.60)\end{tabular}} & \multicolumn{1}{c|}{NA} & \begin{tabular}[c]{@{}c@{}}5.94 \\ (2.31$\sim$8.88)\end{tabular} \\ \hline
        \textbf{\begin{tabular}[c]{@{}c@{}}3rd \\ (DD \\ \& Mast. EC)\end{tabular}} & 4 & \multicolumn{1}{c|}{\begin{tabular}[c]{@{}c@{}}0.39 \\ (0.27$\sim$0.70)\end{tabular}} & \multicolumn{1}{c|}{\begin{tabular}[c]{@{}c@{}}0.26 \\ (0.17$\sim$0.41)\end{tabular}} & \multicolumn{1}{c|}{\begin{tabular}[c]{@{}c@{}}0.61 \\ (0.22$\sim$1.50)\end{tabular}} & \multicolumn{1}{c|}{\begin{tabular}[c]{@{}c@{}}2.98 \\ (1.14$\sim$7.03)\end{tabular}} & \multicolumn{1}{c|}{NA} & \begin{tabular}[c]{@{}c@{}}3.86 \\ (2.00$\sim$5.73)\end{tabular} \\ \hline
        \multirow{3}{*}{\textbf{\begin{tabular}[c]{@{}c@{}}4th \\ (DD \\ \& SE-S)\end{tabular}}} & 5 & \multicolumn{1}{c|}{\begin{tabular}[c]{@{}c@{}}0.18 \\ (0.11$\sim$0.39)\end{tabular}} & \multicolumn{1}{c|}{\begin{tabular}[c]{@{}c@{}}0.78 \\ (0.55$\sim$1.16)\end{tabular}} & \multicolumn{1}{c|}{\begin{tabular}[c]{@{}c@{}}0.16 \\ (0.06$\sim$0.36)\end{tabular}} & \multicolumn{1}{c|}{\begin{tabular}[c]{@{}c@{}}0.65 \\ (0.28$\sim$1.42)\end{tabular}} & \multicolumn{1}{c|}{NA} & \begin{tabular}[c]{@{}c@{}}0.98 \\ (0.05$\sim$1.78)\end{tabular} \\ \cline{2-8}
                                             & 6 & \multicolumn{1}{c|}{\begin{tabular}[c]{@{}c@{}}0.16 \\ (0.08$\sim$0.35)\end{tabular}} & \multicolumn{1}{c|}{\begin{tabular}[c]{@{}c@{}}0.51 \\ (0.34$\sim$0.74)\end{tabular}} & \multicolumn{1}{c|}{\begin{tabular}[c]{@{}c@{}}0.12 \\ (0.04$\sim$0.30)\end{tabular}} & \multicolumn{1}{c|}{\begin{tabular}[c]{@{}c@{}}0.98 \\ (0.28$\sim$2.22)\end{tabular}} & \multicolumn{1}{c|}{\begin{tabular}[c]{@{}c@{}}1.75 \\ (1.21$\sim$2.50)\end{tabular}} & \begin{tabular}[c]{@{}c@{}}1.25 \\ (0.86$\sim$2.07)\end{tabular} \\ \cline{2-8}
                                             & 7 & \multicolumn{1}{c|}{\begin{tabular}[c]{@{}c@{}}0.22 \\ (0.10$\sim$0.57)\end{tabular}} & \multicolumn{1}{c|}{\begin{tabular}[c]{@{}c@{}}0.28 \\ (0.13$\sim$0.47)\end{tabular}} & \multicolumn{1}{c|}{\begin{tabular}[c]{@{}c@{}}0.07 \\ (0.02$\sim$0.19)\end{tabular}} & \multicolumn{1}{c|}{\begin{tabular}[c]{@{}c@{}}0.08 \\ (0.05$\sim$0.14)\end{tabular}} & \multicolumn{1}{c|}{\begin{tabular}[c]{@{}c@{}}1.12 \\ (0.75$\sim$1.75)\end{tabular}} & \begin{tabular}[c]{@{}c@{}}1.70 \\ (0.79$\sim$3.10)\end{tabular} \\ \hline
    \end{tabular}
\end{table}

\subsubsection{Discussions}
Based on the results, it is evident that the split branch/trunk strategy does not perform well for complex structures. The primary reason is that, in its network structure, each neuron in the preceding layer connects to every neuron in the subsequent layer. The loss propagates back to all neurons, updating their weights simultaneously for all variables. This makes optimizing the weights challenging because even the smallest change in one neuron's weight affects all outputs, complicating the search for an optimized solution. In contrast, using multiple DeepONets (N-DeepONets) for each variable, linked by a common loss function, changes the weight update strategy. Changes in the weight of one neuron's independent DeepONet do not affect the outputs of other DeepONets. This demonstrates that multiple DeepONets are better suited for handling complex outputs and structures. For the 2D beam problem, the split branch/trunk strategy performs well because the difference between output variables is not significant. Thus, it can be effective for smaller and more compact analyses. However, multiple DeepONets are more appropriate for larger and more complex analyses.

The results also show that the DD loss can provide good predictions, even for large and complex structures. However, relying solely on data can sometimes lead to unpredictable errors, especially when dealing with out-of-distribution data. Incorporating physics into the network training is therefore crucial. In this study, we shift from traditional physics incorporation methods to a novel approach by adopting a pre-calculated stiffness matrix for elastic and static cases. The results indicate that both the energy conservation principle and static equilibrium using the Schur complement effectively incorporate physics into the training. The choice of physics incorporation during training varies by case. For complex structures, the SE-S loss function proves efficient and accurate, while for simpler problems, the EC loss function suffices. The EC loss function ensures that the ML network predicts outputs at all domain points. In contrast, the SE-S loss function limits predictions to a few domain points, with remaining outputs obtained through post-processing.

The best-performing network in this study uses the DD \& SE-S loss function. The Schur complement reduces the model size to be trained using DeepONet. By reducing the model size, we decrease the complexity of the data to be learned and reduce the number of domain points, enabling the network to learn faster and more efficiently than the DD approach.  It is crucial to understand that reducing the number of training DOFs ($U_I$) can sometimes lead to significant errors, which may propagate through the remaining DOFs ($U_N$) during post-processing. For instance, if too few points are chosen for training using DeepONet, the accuracy of the post-processing results could be compromised. Therefore, we must find a balance between the number of points chosen for prediction using DeepONet and the accuracy of post-processing results.

\section{Summary and Conclusions} \label{conclu}
This study introduces an innovative method for real-time prediction of structural static responses using DeepONet. Our approach accurately predicts responses under various load types and magnitudes, eliminating the need for extensive remodeling and analysis typically required for each new scenario in FE modeling. We address the challenge of predicting multiple output functions from a single network through two strategies: the split branch/trunk method and the integration of multiple DeepONets (N-DeepONets) into a unified framework.

To ensure predictions adhere to underlying physics, we introduce innovative physics-informed loss functions. Instead of relying on traditional, computationally expensive methods (PDE-based) to incorporate physics into training, we devise a novel strategy that uses structural stiffness matrices to enforce essential equilibrium and energy conservation principles. This approach results in the creation of two novel physics-informed loss functions. The EC loss function ensures energy conservation for the entire system but introduces computational complexity due to matrix multiplications involving thousands of DOFs, resulting in extremely long training times. To address this, we introduce the Schur complement in the static equilibrium equation, reducing the problem domain for training DeepONet. This results in the SE-S loss function, drastically reducing training time and enhancing prediction accuracy while ensuring that the network training follows the static equilibrium relationship. The development of loss functions that leverage the system's stiffness matrix is, to our knowledge, novel and unprecedented in applying ML to incorporate physics. By utilizing various combinations of these losses, we achieve an error rate of less than 5\% while significantly reducing training time. However, the SE-S approach proves to be the best candidate for complex structures involving thousands of DOFs, offering more efficient and accurate solutions than DD training. Additionally, using multiple DeepONets ($N$-independent DeepONets) for each output variable outperforms the split branch/trunk strategy, offering more stable and accurate predictions.

The current model showcases robust performance by training on just 20\% of the dataset and testing on the remaining 80\%, which consists of cases not seen during training. These test cases span a wide range of real-life loading conditions, emphasizing the model's strong generalization capabilities for static and elastic scenarios. Additionally, the model holds significant potential for extension to more complex structures under nonlinear dynamic loading. This advancement can be achieved by refining the DeepONet architecture and adapting the proposed physics-informed approaches to incorporate time-dependent effects, which will be explored in future work.

Overall, our method achieves over 95\% accuracy, effectively eliminating the need for remodeling FE models under new loading cases and significantly reducing pre-processing and FE analysis time. This makes the approach a strong candidate for predicting structural responses under various types of loading.

\section*{Credit authorship statement}
\textbf{Bilal Ahmed:} Conceptualization, Methodology, Software, Formal Analysis, Writing - Original draft, Data generation, Visualization, Supervision. \textbf{Yuqing Qiu:} Software, Formal Analysis, Writing - Review and Editing. \textbf{Diab Abueidda:} Conceptualization, Methodology, Software, Formal Analysis, Writing - Review and Editing. \textbf{Waleed El-Sekelly:} Supervision, Project administration. \textbf{Borja García de Soto:} Supervision, Project administration, Funding acquisition.  \textbf{Tarek Abdoun:} Supervision, Project administration, Funding acquisition. \textbf{Mostafa Mobasher:} Conceptualization, Methodology, Writing - Review and Editing, Supervision, Project administration, Funding acquisition.

\section*{Acknowledgment}
This work was partially supported by the Sand Hazards and Opportunities for Resilience, Energy, and Sustainability (SHORES) Center, funded by Tamkeen under the NYUAD Research Institute Award CG013. This work was also partially supported by Sandooq Al Watan Applied Research and Development (SWARD), funded by Grant No.: SWARD-F22-018. The authors wish to express their gratitude to the NYUAD Center for Research Computing for their provision of resources, services, and skilled personnel. The authors would like to thank Dr. Kristof Maes (KU Leuven) for his input and advice on modeling the KW51 bridge, and to Mareya Alkhoori (NYUAD) for her assistance with the parametric study in this paper.

\section*{Data availability}
All models and data will be made publicly available when the paper is accepted.

 \bibliographystyle{elsarticle-num} 
 \bibliography{refs}

\begin{thebibliography}{10}
\expandafter\ifx\csname url\endcsname\relax
  \def\url#1{\texttt{#1}}\fi
\expandafter\ifx\csname urlprefix\endcsname\relax\def\urlprefix{URL }\fi
\expandafter\ifx\csname href\endcsname\relax
  \def\href#1#2{#2} \def\path#1{#1}\fi

\bibitem{brenner2002mathematical}
S.~Brenner, L.~Scott, The Mathematical Theory of Finite Element Methods, Texts in Applied Mathematics, Springer New York, 2002.

\bibitem{hughes2012finite}
T.~J. Hughes, The Finite Element Method: Linear Static and Dynamic Finite Element Analysis, Courier Corporation, 2012.

\bibitem{xu2022real}
Y.~Xu, X.~Lu, Y.~Tian, Y.~Huang, Real-time seismic damage prediction and comparison of various ground motion intensity measures based on machine learning, Journal of Earthquake Engineering 26~(8) (2022) 4259--4279.

\bibitem{ye2022real}
Z.~Ye, S.-C. Hsu, H.-H. Wei, Real-time prediction of structural fire responses: A finite element-based machine-learning approach, Automation in Construction 136 (2022) 104165.

\bibitem{guarize2007neural}
R.~Guarize, N.~Matos, L.~Sagrilo, E.~Lima, Neural networks in the dynamic response analysis of slender marine structures, Applied Ocean Research 29~(4) (2007) 191--198.

\bibitem{lagaros2012neural}
N.~D. Lagaros, M.~Papadrakakis, Neural network based prediction schemes of the non-linear seismic response of 3d buildings, Advances in Engineering Software 44~(1) (2012) 92--115.

\bibitem{liao2021automated}
W.~Liao, X.~Lu, Y.~Huang, Z.~Zheng, Y.~Lin, Automated structural design of shear wall residential buildings using generative adversarial networks, Automation in Construction 132 (2021) 103931.

\bibitem{wang2022end}
C.~Wang, L.-h. Song, J.-s. Fan, End-to-end structural analysis in civil engineering based on deep learning, Automation in Construction 138 (2022) 104255.

\bibitem{xu2020prediction}
Z.~Xu, Y.~Wu, M.-z. Qi, M.~Zheng, C.~Xiong, X.~Lu, Prediction of structural type for city-scale seismic damage simulation based on machine learning, Applied Sciences 10~(5) (2020) 1795.

\bibitem{lei2019fault}
J.~Lei, C.~Liu, D.~Jiang, Fault diagnosis of wind turbine based on long short-term memory networks, Renewable energy 133 (2019) 422--432.

\bibitem{mas2017initial}
M.~I. Mas, M.~I. Fanany, T.~Devin, L.~A. Sutawika, An initial exploration of the suitability of long-short-term-memory networks for multiple site fatigue damage prediction on aircraft lap joints, in: 2017 International Conference on Advanced Computer Science and Information Systems (ICACSIS), IEEE, 2017, pp. 415--422.

\bibitem{qiu2019modified}
D.~Qiu, Z.~Liu, Y.~Zhou, J.~Shi, Modified bi-directional lstm neural networks for rolling bearing fault diagnosis, in: ICC 2019-2019 IEEE international conference on communications (ICC), IEEE, 2019, pp. 1--6.

\bibitem{saleem2024machine}
N.~Saleem, S.~Mangalathu, B.~Ahmed, J.-S. Jeon, Machine learning-based peak ground acceleration models for structural risk assessment using spatial data analysis, Earthquake Engineering \& Structural Dynamics 53~(1) (2024) 152--178.

\bibitem{xu2023computer}
Y.~Xu, Y.~Li, X.~Zheng, X.~Zheng, Q.~Zhang, Computer-vision and machine-learning-based seismic damage assessment of reinforced concrete structures, Buildings 13~(5) (2023) 1258.

\bibitem{xu2023vision}
Y.~Xu, W.~Qiao, J.~Zhao, Q.~Zhang, H.~Li, Vision-based multi-level synthetical evaluation of seismic damage for rc structural components: a multi-task learning approach, Earthquake Engineering and Engineering Vibration 22~(1) (2023) 69--85.

\bibitem{xu2019automatic}
Y.~Xu, S.~Wei, Y.~Bao, H.~Li, Automatic seismic damage identification of reinforced concrete columns from images by a region-based deep convolutional neural network, Structural Control and Health Monitoring 26~(3) (2019) e2313.

\bibitem{xu2022typical}
Y.~Xu, W.~Qian, N.~Li, H.~Li, Typical advances of artificial intelligence in civil engineering, Advances in Structural Engineering 25~(16) (2022) 3405--3424.

\bibitem{WANG2020101538}
C.~Wang, X.~Tan, S.~Tor, C.~Lim, Machine learning in additive manufacturing: State-of-the-art and perspectives, Additive Manufacturing 36 (2020) 101538.

\bibitem{he2024sequential}
J.~He, S.~Kushwaha, J.~Park, S.~Koric, D.~Abueidda, I.~Jasiuk, Sequential deep operator networks (s-deeponet) for predicting full-field solutions under time-dependent loads, Engineering Applications of Artificial Intelligence 127 (2024) 107258.

\bibitem{zhang2022dynamic}
W.~Zhang, J.~Xu, T.~Yu, Dynamic behaviors of bio-inspired structures: Design, mechanisms, and models, Engineering Structures 265 (2022) 114490.

\bibitem{ahmed2022seismic}
B.~Ahmed, S.~Mangalathu, J.-S. Jeon, Seismic damage state predictions of reinforced concrete structures using stacked long short-term memory neural networks, Journal of Building Engineering 46 (2022) 103737.

\bibitem{ahmed2023generalized}
B.~Ahmed, S.~Mangalathu, J.-S. Jeon, Generalized stacked lstm for the seismic damage evaluation of ductile reinforced concrete buildings, Earthquake Engineering \& Structural Dynamics 52~(11) (2023) 3477--3503.

\bibitem{ahmed2023unveiling}
B.~Ahmed, S.~Mangalathu, J.-S. Jeon, Unveiling out-of-distribution data for reliable structural damage assessment in earthquake emergency situations, Automation in Construction 156 (2023) 105142.

\bibitem{cha2017deep}
Y.-J. Cha, W.~Choi, O.~B{\"u}y{\"u}k{\"o}zt{\"u}rk, Deep learning-based crack damage detection using convolutional neural networks, Computer-Aided Civil and Infrastructure Engineering 32~(5) (2017) 361--378.

\bibitem{mozaffar2019deep}
M.~Mozaffar, R.~Bostanabad, W.~Chen, K.~Ehmann, J.~Cao, M.~Bessa, Deep learning predicts path-dependent plasticity, Proceedings of the National Academy of Sciences 116~(52) (2019) 26414--26420.

\bibitem{egli2021surrogate}
F.~S. Egli, R.~C. Straube, A.~Mielke, T.~Ricken, Surrogate modeling of a nonlinear, biphasic model of articular cartilage with artificial neural networks, PAMM 21~(1) (2021) e202100188.

\bibitem{abueidda2021deep}
D.~W. Abueidda, S.~Koric, N.~A. Sobh, H.~Sehitoglu, Deep learning for plasticity and thermo-viscoplasticity, International Journal of Plasticity 136 (2021) 102852.

\bibitem{raissi2019physics}
M.~Raissi, P.~Perdikaris, G.~E. Karniadakis, Physics-informed neural networks: A deep learning framework for solving forward and inverse problems involving nonlinear partial differential equations, Journal of Computational physics 378 (2019) 686--707.

\bibitem{abueidda2021meshless}
D.~W. Abueidda, Q.~Lu, S.~Koric, Meshless physics-informed deep learning method for three-dimensional solid mechanics, International Journal for Numerical Methods in Engineering 122~(23) (2021) 7182--7201.

\bibitem{pantidis2023integrated}
P.~Pantidis, M.~E. Mobasher, Integrated finite element neural network ({I-FENN}) for non-local continuum damage mechanics, Computer Methods in Applied Mechanics and Engineering 404 (2023) 115766.

\bibitem{abueidda2024fenn}
D.~W. Abueidda, M.~E. Mobasher, {I-FENN} for thermoelasticity based on physics-informed temporal convolutional network ({PI-TCN}), Computational Mechanics (2024).

\bibitem{niaki2021physics}
S.~A. Niaki, E.~Haghighat, T.~Campbell, A.~Poursartip, R.~Vaziri, Physics-informed neural network for modelling the thermochemical curing process of composite-tool systems during manufacture, Computer Methods in Applied Mechanics and Engineering 384 (2021) 113959.

\bibitem{henkes2022physics}
A.~Henkes, H.~Wessels, R.~Mahnken, Physics informed neural networks for continuum micromechanics, Computer Methods in Applied Mechanics and Engineering 393 (2022) 114790.

\bibitem{rao2021physics}
C.~Rao, H.~Sun, Y.~Liu, Physics-informed deep learning for computational elastodynamics without labeled data, Journal of Engineering Mechanics 147~(8) (2021) 04021043.

\bibitem{yu2018deep}
B.~Yu, et~al., The deep ritz method: a deep learning-based numerical algorithm for solving variational problems, Communications in Mathematics and Statistics 6~(1) (2018) 1--12.

\bibitem{liao2019deep}
Y.~Liao, P.~Ming, Deep nitsche method: Deep ritz method with essential boundary conditions, arXiv preprint arXiv:1912.01309 (2019).

\bibitem{samaniego2020energy}
E.~Samaniego, C.~Anitescu, S.~Goswami, V.~M. Nguyen-Thanh, H.~Guo, K.~Hamdia, X.~Zhuang, T.~Rabczuk, An energy approach to the solution of partial differential equations in computational mechanics via machine learning: Concepts, implementation and applications, Computer Methods in Applied Mechanics and Engineering 362 (2020) 112790.

\bibitem{nguyen2020deep}
V.~M. Nguyen-Thanh, X.~Zhuang, T.~Rabczuk, A deep energy method for finite deformation hyperelasticity, European Journal of Mechanics-A/Solids 80 (2020) 103874.

\bibitem{abueidda2022deep}
D.~W. Abueidda, S.~Koric, R.~A. Al-Rub, C.~M. Parrott, K.~A. James, N.~A. Sobh, A deep learning energy method for hyperelasticity and viscoelasticity, European Journal of Mechanics-A/Solids 95 (2022) 104639.

\bibitem{lu2021learning}
L.~Lu, P.~Jin, G.~Pang, Z.~Zhang, G.~E. Karniadakis, Learning nonlinear operators via deeponet based on the universal approximation theorem of operators, Nature machine intelligence 3~(3) (2021) 218--229.

\bibitem{wang2021learning}
S.~Wang, H.~Wang, P.~Perdikaris, Learning the solution operator of parametric partial differential equations with physics-informed deeponets, Science advances 7~(40) (2021) eabi8605.

\bibitem{koric2023data}
S.~Koric, D.~W. Abueidda, Data-driven and physics-informed deep learning operators for solution of heat conduction equation with parametric heat source, International Journal of Heat and Mass Transfer 203 (2023) 123809.

\bibitem{yin2022interfacing}
M.~Yin, E.~Zhang, Y.~Yu, G.~E. Karniadakis, Interfacing finite elements with deep neural operators for fast multiscale modeling of mechanics problems, Computer methods in applied mechanics and engineering 402 (2022) 115027.

\bibitem{goswami2022physics}
S.~Goswami, M.~Yin, Y.~Yu, G.~E. Karniadakis, A physics-informed variational deeponet for predicting crack path in quasi-brittle materials, Computer Methods in Applied Mechanics and Engineering 391 (2022) 114587.

\bibitem{he2023novel}
J.~He, S.~Koric, S.~Kushwaha, J.~Park, D.~Abueidda, I.~Jasiuk, Novel deeponet architecture to predict stresses in elastoplastic structures with variable complex geometries and loads, Computer Methods in Applied Mechanics and Engineering 415 (2023) 116277.

\bibitem{wang2023long}
S.~Wang, P.~Perdikaris, Long-time integration of parametric evolution equations with physics-informed deeponets, Journal of Computational Physics 475 (2023) 111855.

\bibitem{goswami2023physics}
S.~Goswami, A.~Bora, Y.~Yu, G.~E. Karniadakis, Physics-informed deep neural operator networks, in: Machine Learning in Modeling and Simulation: Methods and Applications, Springer, 2023, pp. 219--254.

\bibitem{lu2022comprehensive}
L.~Lu, X.~Meng, S.~Cai, Z.~Mao, S.~Goswami, Z.~Zhang, G.~E. Karniadakis, A comprehensive and fair comparison of two neural operators (with practical extensions) based on fair data, Computer Methods in Applied Mechanics and Engineering 393 (2022) 114778.

\bibitem{chen1995universal}
T.~Chen, H.~Chen, Universal approximation to nonlinear operators by neural networks with arbitrary activation functions and its application to dynamical systems, IEEE transactions on neural networks 6~(4) (1995) 911--917.

\bibitem{back2002universal}
A.~D. Back, T.~Chen, Universal approximation of multiple nonlinear operators by neural networks, Neural Computation 14~(11) (2002) 2561--2566.

\bibitem{CARLSON1986257}
D.~Carlson, What are schur complements, anyway?, Linear Algebra and its Applications 74 (1986) 257--275.

\bibitem{langer2005coupled}
U.~Langer, O.~Steinbach, Coupled boundary and finite element tearing and interconnecting methods, in: Domain decomposition methods in science and engineering, Springer, 2005, pp. 83--97.

\bibitem{pechstein2012finite}
C.~Pechstein, Finite and boundary element tearing and interconnecting solvers for multiscale problems, Vol.~90, Springer Science \& Business Media, 2012.

\bibitem{mobasher2016adaptive}
M.~E. Mobasher, H.~Waisman, Adaptive modeling of damage growth using a coupled fem/bem approach, International Journal for Numerical Methods in Engineering 105~(8) (2016) 599--619.

\bibitem{hackbusch2005direct}
W.~Hackbusch, B.~N. Khoromskij, R.~Kriemann, Direct schur complement method by domain decomposition based on h-matrix approximation, Computing and visualization in science 8 (2005) 179--188.

\bibitem{lu2021deepxde}
L.~Lu, X.~Meng, Z.~Mao, G.~E. Karniadakis, {DeepXDE}: A deep learning library for solving differential equations, SIAM Review 63~(1) (2021) 208--228.

\bibitem{pantidis2023error}
P.~Pantidis, H.~Eldababy, C.~M. Tagle, M.~E. Mobasher, Error convergence and engineering-guided hyperparameter search of pinns: towards optimized i-fenn performance, Computer Methods in Applied Mechanics and Engineering 414 (2023) 116160.

\bibitem{KW51-Monitoring-Frequency}
K.~Maes, G.~Lombaert, Monitoring railway bridge kw51 before, during, and after retrofitting, Journal of Bridge Engineering 26~(3) (2021) 04721001.

\bibitem{Structurae}
{S}tructurae - {I}nternational {D}atabase and {G}allery of {S}tructures --- structurae.net, \url{https://structurae.net/en}, [Accessed 22-06-2024].

\bibitem{KW51-MSSP-stresses}
K.~Maes, G.~Lombaert, Validation of virtual sensing for the reconstruction of stresses in a railway bridge using field data of the kw51 bridge, Mechanical Systems and Signal Processing 190 (2023) 110142.

\bibitem{KW51-NaturalFrequency-MSSP}
D.~Anastasopoulos, G.~{De Roeck}, E.~P. Reynders, One-year operational modal analysis of a steel bridge from high-resolution macrostrain monitoring: Influence of temperature vs. retrofitting, Mechanical Systems and Signal Processing 161 (2021) 107951.

\end{thebibliography}

\end{document}